\documentclass[runningheads]{llncs}
\usepackage{amssymb}
\usepackage{float}
\usepackage{graphicx}
\usepackage{subfigure}
\usepackage{algorithm}
\usepackage{algorithmicx}
\usepackage[noend]{algpseudocode}
\usepackage{amsmath}
\usepackage{CJK}
\usepackage{verbatim}
\usepackage{url}
\usepackage{todonotes}
\begin{document}
\title{Analysis of Hyper-Parameters for Small Games:\\ Iterations or Epochs in Self-Play?}
\titlerunning{Analysis of Hyper-Parameters for Small Games}
% If the paper title is too long for the running head, you can set
% an abbreviated paper title here
%
\author{Hui Wang\and
Michael Emmerich\and
Mike Preuss\and Aske Plaat}
\authorrunning{H. Wang et al.}
% First names are abbreviated in the running head.
% If there are more than two authors, 'et al.' is used.
%
\institute{Leiden Institute of Advanced Computer Science, Leiden University,\\ Leiden, the Netherlands\\
\email{h.wang.13@liacs.leidenuniv.nl}\\
\url{http://www.cs.leiden.edu}
}

\maketitle              % typeset the header of the contribution
\begin{abstract}
The landmark achievements of AlphaGo Zero have created great research interest into self-play in reinforcement learning. 
%Since AlphaGo and AlphaGo Zero have achieved groundbreaking successes in the game of Go, the programs have been  generalized to solve other tasks. Subsequently, AlphaZero was developed to play Go, Chess and Shogi. 
In self-play, Monte Carlo Tree Search is used to train a deep neural network, that is then used in tree searches. Training itself is governed by many hyper-parameters. % The  neural network is two-headed: it has a policy-head and a value-head, and during training, the optimizer minimizes the \textem {sum} of the  policy loss and the value loss. 
% In the literature, the algorithms are explained  well. However, AlphaZero contains many parameters, and for neither AlphaGo, AlphaGo Zero nor AlphaZero, 
There has been surprisingly little research on design choices for hyper-parameter values and loss-functions, presumably because of the prohibitive computational cost to explore the parameter space. 
%insufficient discussion about how to set parameter values in these algorithms. In addition, it is also not clear if and under which circumstances other formulations of the loss function are better. Therefore, 
In this paper, we investigate 12 hyper-parameters in an AlphaZero-like self-play algorithm and evaluate how these parameters contribute to training. We use small games, to achieve meaningful exploration with moderate computational effort.
The experimental results show that training is  highly sensitive to hyper-parameter choices. %, proving the importance of such a parameter sweep.  We categorize these 12 parameters into time-sensitive parameters and time-friendly parameters. Moreover, t
Through multi-objective analysis we identify %, this paper provides an insightful basis for further hyper-parameter optimization. In terms of time cost and playing strength, we select 
4 important hyper-parameters
%~(iteration, MCTS simulation, episode and epoch) 
to further assess. To start, we find surprising results where too much training can sometimes lead to lower performance. Our main result is that the number of self-play iterations subsumes MCTS-search simulations, game-episodes, and training epochs. The intuition is that  these three increase together as self-play iterations increase, and that increasing them individually is sub-optimal. A  consequence of our experiments is a direct recommendation for setting hyper-parameter values in self-play: the overarching outer-loop of self-play iterations should be maximized, in favor of the three  inner-loop hyper-parameters, which should be set at lower values. A secondary result of our experiments concerns the choice of optimization goals, for which we also provide recommendations. 
\end{abstract}
\section{Introduction}\label{sec:introduction}
% The very first letter is a 2 line initial drop letter followed
% by the rest of the first word in caps.
% 
% form to use if the first word consists of a single letter:
% \IEEEPARstart{A}{demo} file is ....
% 
% form to use if you need the single drop letter followed by
% normal text (unknown if ever used by the IEEE):
% \IEEEPARstart{A}{}demo file is ....
% 
% Some journals put the first two words in caps:
% \IEEEPARstart{T}{his demo} file is ....
% 
% Here we have the typical use of a "T" for an initial drop letter
% and "HIS" in caps to complete the first word.
The AlphaGo series of papers~\cite{Silver2016,Silver2017a,Silver2018} have sparked much interest of researchers and the general public alike into deep reinforcement learning. %AlphaGo~\cite{Silver2016} achieves superhuman performance in playing Go by applying tree search to evaluate  positions and selecting moves from the trained neural networks. However, these networks are trained from human experts' data by means of supervised learning.
%AlphaGo Zero~\cite{Silver2017a},  the successor of AlphaGo, masters the game of Go  without human knowledge. It generates game playing data purely by an elegant form of self-play, training a single unified neural network with a policy head and a value head, in a Monte Carlo Tree Search~(MCTS) searcher. AlphaZero~\cite{Silver2018} is a generalized version of AlphaGo Zero and claims a general framework of playing different games~(such as Go, Chess and Shogi) without human knowledge. The AlphaZero framework demonstrates the strong adaptability of such  deep reinforcement learning algorithms which combine self-play, neural networks and tree search to solve game playing problems. Inspired by the AlphaGo series of algorithms, many analysis reviews, applications and optimization methods~\cite{Granter2017,Wang2016,Fu2016} have been published, making deep reinforcement learning a more and more active and practical research field.
Despite the success of AlphaGo and related methods in Go and other application areas~\cite{Tao2016,Zhang2016}, there are unexplored and unsolved puzzles in the design and parameterization of the algorithms. Different hyper-parameter settings can lead to very different results. %A proper parameter setting should be found to guarantee the expected capability of the algorithm. 
However, hyper-parameter design-space sweeps are computationally very expensive, and in the original publications, we can only find limited information of how to set the values of some important parameters and why. Also, there are few works on how to set the hyper-parameters for these algorithms, and more insight into the  hyper-parameter interactions is  necessary. In our work, we study the most general framework algorithm in the aforementioned AlphaGo series by using a lightweight re-implementation of AlphaZero: AlphaZeroGeneral~\cite{Surag2018}. 

In order to optimize hyper-parameters, it is  important to understand their function and interactions in an algorithm. A single iteration in the AlphaZeroGeneral framework consists of three stages: self-play, neural network training and arena comparison. In these stages, we explore 12 hyper-parameters~(see section~\ref{a0gintroduction}) in AlphaZeroGeneral. Furthermore, we observe 2 objectives~(see section~\ref{a0gtargets}): training loss and time cost in each single run. A sweep of the hyper-parameter space is computationally demanding. In order to provide a meaningful analysis we use small board sizes of typical combinatorial games. 
This sweep provides an overview of the hyper-parameter contributions and provides a basis for further analysis. Based on these results, we choose 4 interesting parameters to further evaluate in depth.

%This work provides a number of experiments in order to better understand the parameter interactions. We use small games in order to be able to analyze a sufficiently large part of the design space within a reasonable computation budget. 
%This is not ideal in terms of experimental setup but we need to compromise with the enormous effort the combinatorial explosion of parameter combinations would entail.

As performance measure, we use the Elo rating that can be computed during training time of the self-play system, as a running relative Elo, and computed separately, in a dedicated tournament between different trained players.

Our contributions can be summarized as follows:
\begin{enumerate}
\item %We sweep 12 hyper-parameters in AlphaZeroGeneral and analyse loss and time cost for 6$\times$6 Othello, 
%see Fig.~\ref{fig:figalldefault}--\ref{fig:subfigparasweeploss} and Table.~\ref{timecosttab},
%and select the 4 most promising parameters for further optimization.
%, see Table.~\ref{losstimetab}.
 We find (1) that in general higher values of all hyper-parameters lead to higher playing strength, but (2) that  within a limited budget, a higher number of outer iterations is more promising than higher numbers of inner iterations: these are subsumed by outer iterations.

\item We evaluate 4 alternative loss functions for 3 games and 2 board sizes, and find that 
the best setting depends on the game and is usually not the sum of policy and value loss. However, the sum may be a good default compromise if no further information about the game is present.
%only optimizing value loss achieves higher Elo ratings in game Othello and Connect Four. However, minimizing only policy value achieves the highest Elo for 6$\times$6 Gobang game.
\end{enumerate}

The paper is structured as follows. We first give an overview of the most relevant literature, before  describing the 
considered test games in Sect.\,\ref{sec:games}. Then we describe the AlphaZero-like self-play algorithm in Sect.\,\ref{sec:alphazero}. After setting up experiments, we present the results in Sect.\,\ref{sec:experimentresults}. Finally, we conclude our paper and discuss the promising future work.

\section{Related work}\label{sec:relatedwork}

Hyper-parameter tuning by optimization is very important for many practical algorithms. In reinforcement learning, for instance, the $\epsilon$-greedy strategy of classical Q-learning is used to balance exploration and exploitation. Different $\epsilon$ values lead to different learning performance~\cite{Wang2018}. Another well known example of hyper-parameter tuning is the parameter $C_p$ in Monte Carlo Tree Search (MCTS)~\cite{Browne2012}. There are many works on tuning $C_p$ for different kinds of tasks. These provide insight on setting its value for MCTS in order to balance exploration and exploitation~\cite{Ruijl2014}. In deep reinforcement learning, the effect of the many neural network parameters are a black-box that precludes understanding, although the strong decision accuracy of deep learning is undeniable~\cite{Schmidhuber2015}, % of training, 
as the results in Go (and many other applications) have shown~\cite{Clark2015}. 
% leave out Mnih, not relevant I would say
%Since Mnih et al. reported their work on human-level control through deep reinforcement learning~\cite{Mnih2015} in 2015, the performance of deep Q-network~(DQN) shows us an amazing impression on playing Atari 2600 Games. Thereafter, the applications based on DQN have shown surprising ability of learning to cope with game playing problems. For example, 
After AlphaGo~\cite{Silver2016}, the role of self-play became more and more important. Earlier works on self-play in reinforcement learning are~\cite{Heinz2000,Wiering2010,Van2013}. An overview is provided in~\cite{Plaat2020}. % Aske Plaat, Learning to Play: Reinforcement Learning and Games, Leiden, 2020, forthcoming.

%\subsubsection{Loss functions}

On loss-functions and hyper-parameters for AlphaZero-like systems there are a few studies:
\cite{mandai2018alternative} studied policy and value network optimization as a multi-task learning problem~\cite{caruana1997multitask}. Matsuzaki compares MCTS with evaluation functions of different quality, and finds different results in Othello~\cite{MatsuzakiK2018} than AlphaGo's PUCT. Moreover,~\cite{Matsuzaki18} showed that the value function has more importance than the policy function in the PUCT algorithm for Othello. In our study, we extend this work and look more deeply into the relationship between value and policy functions in games.

%In addition, 
%The neural network in AlphaZero is represented as $f_\theta=(\textbf{p},v)$ (a unified deep network with a policy head and a value head). Policy $\textbf{p}$ is a probability distribution over the best action (used for choosing the best move). The loss function computes a difference between training examples and the predicted policy. A lower policy loss~($l_\textbf{p}$) indicates a more accurate selection of the best move. The value function $v$ is the prediction of the final outcome. A lower value loss~($l_v$) indicates a more accurate prediction of the final outcome. The use of a double-headed network in Alpha(Go) Zero is innovative, and we know of no in-depth study of how the two losses ($l_\textbf{p}$ and $l_v$) contribute to the playing strength of the final player. In Alpha(Go) Zero, the unweighted sum of the two losses is used. Other studies based on the AlphaGo series of algorithms also use the simple summation of the two losses.   However, Matsuzaki et al.~\cite{MatsuzakiK2018} remind us to carefully study alternative evaluation functions.  In order to increase our understanding of the inner workings of the minimization of the double-headed network we study different combinations of policy and value loss.
%Therefore, in this work, we investigate:

%\noindent a) what will happen if we only minimize a single target?

%\noindent b) whether a product combination is a good alternative to summation?

Our experiments are also performed using 
%We perform our experiments using a light-weight AlphaZero implementation named 
AlphaZeroGeneral~\cite{Surag2018} on several smaller games, namely 5$\times$5 and 6$\times$6 Othello~\cite{Iwata1994}, 5$\times$5 and 6$\times$6 Connect~Four~\cite{Allis1988} and 5$\times$5 and 6$\times$6 Gobang~\cite{Reisch1980}.
The smaller size of these games allows us to do more experiments, and they also provide us largely uncharted territory where we hope to find effects that cannot be seen in Go or Chess.\footnote{This part of the work is published at the IEEE SSCI 2019 conference~\cite{Wang2019}, and is included here for completeness.}

\section{Test Games}
 \label{sec:games}

\begin{figure}[!tbh]
\centering
%\hspace*{-2em}
\subfigure[6$\times$6 Othello]{\label{fig:subfiggames:66othello}
\includegraphics[width=0.45\textwidth]{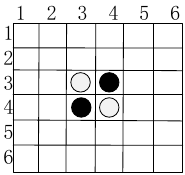}}
\hspace*{0.5em}
\subfigure[5$\times$5 Othello]{\label{fig:subfiggames:55othello}
\includegraphics[width=0.45\textwidth]{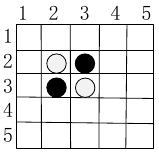}}
%\hspace*{0.5em}
\subfigure[5$\times$5 Connect Four]{\label{fig:subfiggames:55connect4}
\includegraphics[width=0.45\textwidth]{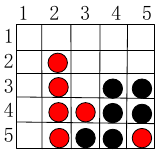}}
\hspace*{0.5em}
\subfigure[5$\times$5 Gobang]{\label{fig:subfiggames:55Gobang}
\includegraphics[width=0.45\textwidth]{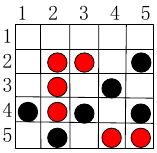}}
\caption{Starting positions for Othello, examples for Connect Four and Gobang}
%Our test games on $5\times5$ boards}
\label{fig:subfiggames} %% label for entire figure
\end{figure}

 In our hyper-parameter sweep experiments, we use Othello with  a 6$\times$6 board size, see Fig.~\ref{fig:subfiggames:66othello}. In the alternative loss function experiments, we use the games Othello, Connect Four and Gobang, each with 5$\times$5 and 6$\times$6 board sizes. Othello is a two-player game. Players take turns placing their own color pieces. Any opponent's color pieces that are in a straight line and bounded by the piece just placed and another piece of the current player's are flipped to the current player's color. While the last legal position is filled, the player who has most pieces wins the game. Fig.~\ref{fig:subfiggames}(a/b) show the start configurations for Othello. Connect Four is a two-player connection game. Players take turns dropping their own pieces from the top into a vertically suspended grid. The pieces fall straight down and occupy the lowest  position within the column. The player who first forms a horizontal, vertical, or diagonal line of four  pieces wins the game. Fig.~\ref{fig:subfiggames:55connect4} is a game termination example for 5$\times$5 Connect Four where the red player wins the game. Gobang is another connection games that  traditionally is played with Go pieces (black and white stones) on a Go board. %We employ  5$\times$5 and 6$\times$6 boards. 
 Players alternate turns, placing a stone of their color on an empty position. The winner is the first player to form an unbroken chain of 4 stones horizontally, vertically, or diagonally. Fig.~\ref{fig:subfiggames:55Gobang} is a game termination example for 5$\times$5 Gobang where the black player wins the game.

There is a wealth of research on finding playing strategies for these three games by means of different methods. For example, Buro created Logistello~\cite{Buro97} to play Othello. Chong et al.\ described the evolution of neural networks for learning to play Othello~\cite{ChongTW05}. Thill et al.\ applied temporal difference learning to play Connect Four~\cite{ThillBKK14}. Zhang et al.\ designed evaluation functions for Gobang~\cite{ZhangWu2012}. Moreover, Banerjee et al.\ tested knowledge transfer in General Game Playing on small games including 4$\times$4 Othello~\cite{BanerjeeS07}. Wang et al.\ assessed the potential of classical Q-learning based on small games including 4$\times$4 Connect Four~\cite{Wang18}. The board size gives us a handle to reduce or increase the overall difficulty of these games.
%Obviously, these three games are commonly tested with flexible board size in game playing. 
In our experiments we use AlphaZero-like zero learning, where a reinforcement learning system learns from tabula rasa, by playing games against itself using a combination of deep reinforcement learning and MCTS.

\section{AlphaZero-like Self-play}
\label{sec:alphazero}

\subsection{The Base Algorithm}\label{a0gintroduction}
Following the works by Silver et al.~\cite{Silver2017a,Silver2018} the fundamental structure of AlphaZero-like Self-play is an iteration over three different stages (see Algorithm~\ref{alg:a0g}). 
%\allowdisplaybreaks
\begin{algorithm*}[t]
%\footnotesize
\caption{AlphaZero-like Self-play Algorithm}
\label{alg:a0g}
\begin{algorithmic}[1]
\Function{AlphaZeroGeneral}{}
\State Initialize $f_\theta$ with random weights; Initialize retrain buffer $D$ with capacity $N$ 
\For{iteration=1, $\dots$, $I$}
\For{episode=1,$\dots$, $E$}\Comment{stage 1}
\For{t=1, $\dots$, $T\prime$, $\dots$, $T$}
\State Get an enhanced best move prediction $\pi_{t}$ by performing MCTS based on $f_\theta(s_t)$ 
\State Before step $T\prime$, select random action $a_t$ based on probability $\pi_t$, else select action $a_t=\arg\max_a(\pi_t)$
\State Store example ($s_t$, $\pi_t$, $z_t$) in $D$
\State Set $s_t$=excuteAction($s_t$, $a_t$)
\EndFor
\State Label reward $z_t$ ~($t\in[1,T]$) as $z_T$ in examples
\EndFor
\State Randomly sample  minibatch of examples~($s_j$, $\pi_j$, $z_j$) from $D$ \Comment{stage 2}
\State $f_\theta\prime\leftarrow$ Train $f_\theta$ by performing optimizer to minimize Equation ~\ref{totalloss} based on sampled examples 

\State Set $f_\theta=f_\theta\prime$ if $f_\theta\prime$ is better than $f_\theta$ \Comment{stage 3}

\EndFor
\State \Return $f_\theta$;
\EndFunction
\end{algorithmic}
\end{algorithm*}

The first stage is a \textbf{self-play} tournament. The computer player performs several games against itself in order to generate data for further training. In each step of a game~(episode), the player runs MCTS to obtain, for each move,  an enhanced policy $\pi$ based on the probability $\textbf{p}$ provided by the neural network $f_\theta$. We now introduce the hyper-parameters, and their abbreviation that we use in this paper. In MCTS, hyper-parameter $C_p$ is used to balance exploration and exploitation of game tree search, and we  abbreviate it to {\em c}. Hyper-parameter \emph{m} is the number of times to run down from the root for building the game tree, where the parameterized network $f_\theta$ provides the value~($v$) of the states for MCTS. For actual (self-)play, from \emph{T'} steps on, the player always chooses the best move according to $\pi$. Before that, the player always chooses a random move based on the probability distribution of $\pi$. After finishing the games, the new examples are normalized as a form of ($s_t, \pi_t, z_t$) and stored in \emph{D}. 

The second stage consists of \textbf{neural network training},  using data from the self-play tournament. Training lasts for several epochs. In each epoch~(\emph{ep}), training examples are divided into several small batches~\cite{Ioffe2015} according to the specific batch size~(\emph{bs}). The neural network is trained to minimize~\cite{Kingma2014} the value of the \emph{loss function} which (see Equation~\ref{totalloss}) sums up the mean-squared error between predicted outcome and real outcome and the cross-entropy losses between $\textbf{p}$ and $\pi$ with a learning rate~(\emph{lr}) and dropout~(\emph{d}). Dropout is used as probability to randomly ignore some nodes of the hidden layer in order to avoid overfitting~\cite{Srivastava2014}.

The last stage is \textbf{arena comparison}, in which  the newly trained neural network model~($f_\theta\prime$) is run against the previous neural network model~($f_\theta$). The better model is adopted for the next iteration. In order to achieve this, $f_{\theta\prime}$ and $f_\theta$ play against each other for $n$ games. If $f_\theta\prime$ wins more than a fraction of $u$ games, it is replacing the previous best $f_\theta$. Otherwise, $f_{\theta\prime}$ is rejected and $f_\theta$ is kept as current best model. Compared with AlphaGo Zero, AlphaZero does not entail the arena comparison stage anymore. However, we keep this stage for making sure that we can safely recognize improvements. 

\subsection{Loss Function}\label{a0gtargets}
The \textbf{training loss function} consists of $l_\textbf{p}$ and $l_v$. The neural network $f_{\theta}$ is parameterized by $\theta$. $f_{\theta}$ takes the game board state $s$ as input, and provides the value $v_{\theta}\in [-1,1]$ of $s$ and a policy probability distribution vector $\textbf{p}$ over all legal actions as outputs. $\textbf{p}_{\theta}$ is the policy provided by $f_{\theta}$ to guide MCTS for playing games. After performing MCTS, we obtain an improvement estimate for policy $\pi$. Training aims at making $\textbf{p}$ more similar to $\pi$. This can be achieved by minimizing the cross entropy of both distributions. Therefore,  $l_\textbf{p}$ is defined as $-\pi^\top\log\textbf{p}$. The other aim  is to minimize the difference between the output value~($v_{\theta}(s_t)$) of the state $s$ according to $f_{\theta}$ and the real outcome~($z_t\in\{-1,1\}$) of the game.  Therefore, $l_v$ is defined as the mean squared error $(v-z)^2$. Summarizing, the total loss function of AlphaZero is defined in Equation~\ref{totalloss}.
\begin{equation}\label{totalloss}
l_+=-\pi^\top\log\textbf{p}+(v-z)^2
\end{equation}

Note that in AlphaZero's loss function, there is an extra regularization term to guarantee the training stability of the neural network. In order to pay more attention to two evaluation function components, instead, we apply standard measures to avoid overfitting such as the \textbf{drop out} mechanism. 

\subsection{Bayesian Elo System}\label{elointroduction}
The \textbf{Elo rating function} has been developed as a method for calculating the relative skill levels of players in games. Usually, in zero-sum games, there are two players, A and B. If their Elo ratings are $R_A$ and $R_B$, respectively, then the expectation that player A wins the next game is $E_A=\frac{1}{1+10^{(R_B-R_A)/400}}$. If the real outcome of the next game is $S_A$, then the updated Elo of player A can be calculated by $R_A=R_A+K (S_A-E_A)$, where K is the factor of the maximum possible adjustment per game. In practice, K should be bigger for weaker players but smaller for stronger players. Following~\cite{Silver2018}, in our design, we adopt the Bayesian Elo system~\cite{Coulom08} to show the improvement curve of the learning player during self-play. We furthermore also employ this method to assess the playing strength of the final models. 

\subsection{Time Cost Function}\label{timecostequation}
Because of the high computational cost of self-play reinforcement learning, the running time of self-play is of great importance. We have created a  \textbf{time cost function} to predict the running time, based on the algorithmic structure in Algorithm~\ref{alg:a0g}. According to Algorithm~\ref{alg:a0g}, the whole training process consists of several iterations with three steps as  introduced in section~\ref{a0gintroduction}. Please refer to the algorithm and to equation~\ref{timecostfunction}. In $i$th iteration~($1\leq i\leq I$), if we assume that in $j$th episode~($1\leq j\leq E$), for $k$th game step~(the size of $k$ mainly depends on the game complexity), the time cost of $l$th MCTS~($1\leq l\leq m$) simulation is $t_{jkl}^{(i)}$, and assume that for $p$th epoch~($1\leq p \leq ep$), the time cost of pulling $q$th batch~($1\leq q \leq trainingExampleList.size/bs$)\footnote{the size of \emph{trainingExampleList} is also relative to the game complexity} through the neural network is $t_{pq}^{(i)}$, and assume that in $w$th arena comparison~($1\leq w \leq n$), for $x$th game step, the time cost of $y$th  MCTS simulation~($1\leq y \leq m$) is $t_{xyw}^{(i)}$. The time cost of the whole training process is summarized in equation~\ref{timecostfunction}.
\begin{equation}\label{timecostfunction}
\sum_{i}(
\stackrel{self-play}{\overbrace{\sum_{j}\sum_{k}\sum_{l}t_{jkl}^{(i)}}}\ +\stackrel{training}{\overbrace{\sum_{p}\sum_{q}t_{pq}^{(i)}}}+\ \stackrel{arena\ comparison}{\overbrace{\sum_{x}\sum_{y}\sum_{w}t_{xyw}^{(i)}}})
\end{equation}

Please refer to Table~\ref{setparasweepexperiments} for an overview of the hyper-parameters. From Algorithm~\ref{alg:a0g} and equation~\ref{timecostfunction}, we can  see that the hyper-parameters, such as \emph{I}, \emph{E}, \emph{m}, \emph{ep}, \emph{bs}, \emph{rs}, \emph{n} etc., influence training time. In addition, $t_{jkl}^{(i)}$ and $t_{xyw}^{(i)}$ are   simulation time costs that rely on hardware capacity and game complexity. $t_{uv}^{(i)}$ also relies on the structure of the neural network. In our experiments, all neural network models share the same structure, which consists of 4 convolutional layers and 2 fully connected layers.

\section{Experimental Setup}\label{secsetup}
We   sweep the 12 hyper-parameters by configuring 3 different values~(minimum value, default value and maximum value) to find the most promising parameter values.
In each single run of training, we play 6$\times$6 Othello~\cite{Iwata1994} and change the value of one hyper-parameter, keeping the other hyper-parameters at default values (see Table~\ref{defaulttab}).

Our experiments are run on a machine with  128GB RAM, 3TB local storage, 20-core Intel Xeon E5-2650v3 CPUs~(2.30GHz, 40 threads), 2 NVIDIA Titanium GPUs~(each with 12GB memory) and 6 NVIDIA GTX 980 Ti GPUs~(each with 6GB memory). In order to keep using the same GPUs, we deploy each run of experiments on the NVIDIA GTX 980 Ti GPU. Each run of experiments takes 2 to 3 days.

\subsection{Hyper-Parameter Sweep}\label{setparasweepexperiments}
In order to train a player to play 6$\times$6 Othello based on Algorithm~\ref{alg:a0g}, we employ the parameter values in Tab.~\ref{defaulttab}. Each experiment only observes one hyper-parameter, keeping the other hyper-parameters at default values.
\begin{table}[H]
\centering\hspace*{-2.3em}
%\linebreak
\caption{Default Hyper-Parameter Setting}\label{defaulttab}
\begin{tabular}{|l|l|l|l|l|}
\hline
-&Description& Minimum&Default&Maximum\\
\hline
\emph{I}	&number of iteration	&50	&100	&150\\
\hline
\emph{E}    &number of episode	&10	&50&100\\
\hline
\emph{T'}	&step threshold	&10		&15&20\\
\hline
\emph{m}	&MCTS simulation times	&25	&100&	200\\
\hline
\emph{c}    &weight in UCT	&0.5	&1.0&	2.0\\
\hline
\emph{rs}	& number of retrain iteration	&1	&20&40\\
\hline
\emph{ep}	& number of epoch	&5	&10	&15\\
\hline
\emph{bs}	& batch size	&32	&64&96\\
\hline
\emph{lr}	& learning rate	&0.001&	0.005&0.01\\
\hline
\emph{d}    & dropout probability &	0.2&0.3&0.4\\
%\hline
%channel&	32&	256&512\\
\hline
\emph{n}   & number of comparison games &20	&40	&100\\
\hline
\emph{u}	& update threshold	&0.5	&0.6 &	0.7\\
\hline
\end{tabular}
\end{table}

\subsection{Hyper-Parameters Correlation Evaluation}\label{setcorrelationevaluationexperiments}
Based on the above experiments, we further explore the correlation of interesting hyper-parameters~(i.e. \emph{I}, \emph{E}, \emph{m} and \emph{ep}) in terms of their best final player's playing strength and overall training time. We set values for these 4 hyper-parameters as Table~\ref{correlationparatab}, and other parameters values are set to the default values in Tab.~\ref{defaulttab}. In addition, for (and only for) this part of experiments, the stage 3 of Algorithm~\ref{alg:a0g} is cut off. Instead, for every iteration, the trained model $f_\theta\prime$ is accepted as the current best model $f_\theta$ automatically, which is also adopted by AlphaZero and saves a lot of time.

\begin{table}[H]
\centering\hspace*{-2.3em}
%\linebreak
\caption{Correlation Evaluation Hyper-Parameter Setting}\label{correlationparatab}
\begin{tabular}{|l|l|l|l|l|}
\hline
-& Description & Minimum	& Middle& Maximum\\
\hline
\emph{I}	&number of iteration	&25	&50	&75\\
\hline
\emph{E}    &number of episode	&10	&20&30\\
\hline
\emph{m}	&MCTS simulation times	&25	&50&	75\\
\hline
\emph{ep}	&number of epoch	&5&10&15\\
\hline
\end{tabular}
\end{table}

Note that due to computation resource limitations, for hyper-parameter sweep experiments on 6$\times$6 Othello, we only perform single run experiments. This may cause noise, but still provides valuable insights on the importance of hyper-parameters under the AlphaZero-like self-play framework.
 
\subsection{Alternative Loss Function Evaluation}\label{setlossfuctionexperiments}
As we  want to assess the effect of  different loss functions, we employ a weighted sum loss function based on (\ref{weightedtotalloss}):

\begin{equation}\label{weightedtotalloss}
l_\lambda=\lambda(-\pi^\top\log\textbf{p})+(1-\lambda)(v-z)^2
\end{equation}

where $\lambda$ is a weight parameter. This provides some flexibility to gradually change the nature of the function. In our experiments, we first set $\lambda$=0 and $\lambda$=1 in order to assess $l_\textbf{p}$ or $l_v$ independently. Then we use  Equation~\ref{totalloss} as training loss function. Furthermore, we note from the theory of multi-attribute utility functions in multi-criteria optimization~\cite{MichaelA2006} that a sum tends to prefer extreme solutions, whereas a product prefers a more balanced solution. We employ a product combination loss function as follows:
\begin{equation}\label{producttotalloss}
l_\times=-\pi^\top\log\textbf{p}\times(v-z)^2
\end{equation}

For all loss function experiments, each setting is run 8 times to get statistically significant results (we show error bars) using the hyper-parameters of Table~\ref{defaulttab} with their default values. However, in order to allow longer training, we enhance the iteration number to 200 in the smaller games~(5$\times$5 Othello, 5$\times$5 Connect Four and 5$\times$5 Gobang).

The  loss function in question is used to guide each training process, with the expectation that smaller loss means a stronger model. However, in practice, we have found that this is not always the case and another measure is needed to check.
Following Deep Mind's work, we use Bayesian Elo ratings~\cite{Coulom08} to describe the playing strength of the model in every iteration. In addition, for each game, we use all best players trained from the four different  targets ($l_\textbf{p}$, $l_v$, $l_+$, $l_\times$) and \textbf{8 repetitions}\footnote{In alternative loss function evaluation experiments, multiple runs for each setting are employed to avoid bias} plus a random player to play 20 round-robin rounds. We calculate the Elo ratings of the 33 players as the real playing strength of a player, rather than the one measured while training.

\section{Experimental Results}\label{sec:experimentresults}

\begin{figure}[H]
\centering
\hspace*{-2.3em}
\includegraphics[width=0.9\textwidth]{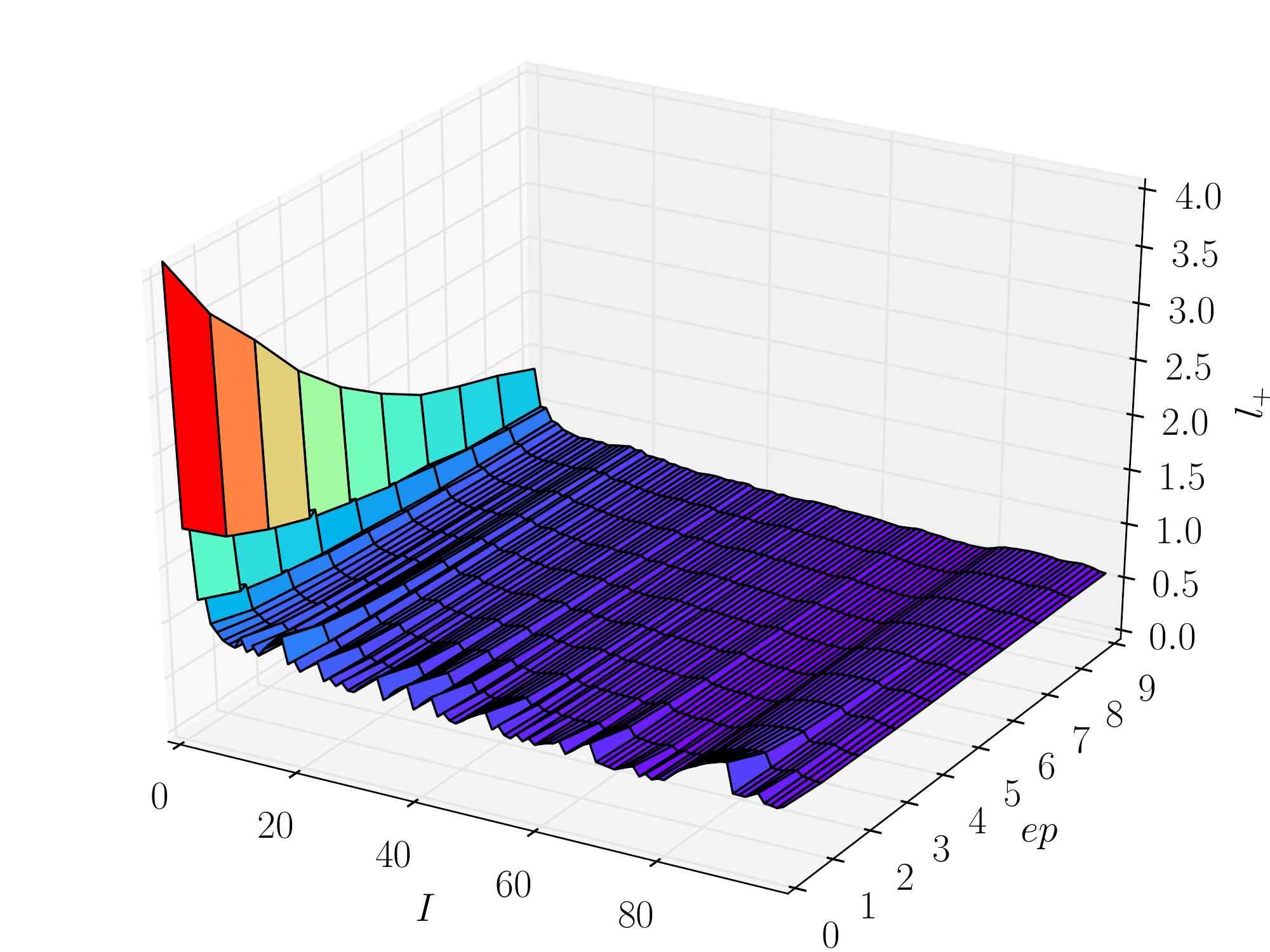}
\caption{Single run training loss over iterations I and epochs ep}
\label{fig:figalldefault} %% label for entire figure
\end{figure}

In order to better understand the training process, first, we depict training loss evolution for default settings in Fig.~\ref{fig:figalldefault}.
%In Fig.~\ref{fig:figalldefault},
We plot the training loss of each epoch in every iteration and see that (1) in each iteration, loss decreases along with increasing epochs, and that (2) loss also decreases with increasing iterations up  to a relatively stable level.

\subsection{Hyper-Parameter Sweep Results}\label{parasweepresults}

\begin{figure}[tbh!]
\centering
\hspace*{-2em}
\subfigure[$l_+$ vs $I$]{\label{fig:subfigparasweep:iteration} %% label for first subfigure
\includegraphics[width=0.35\textwidth]{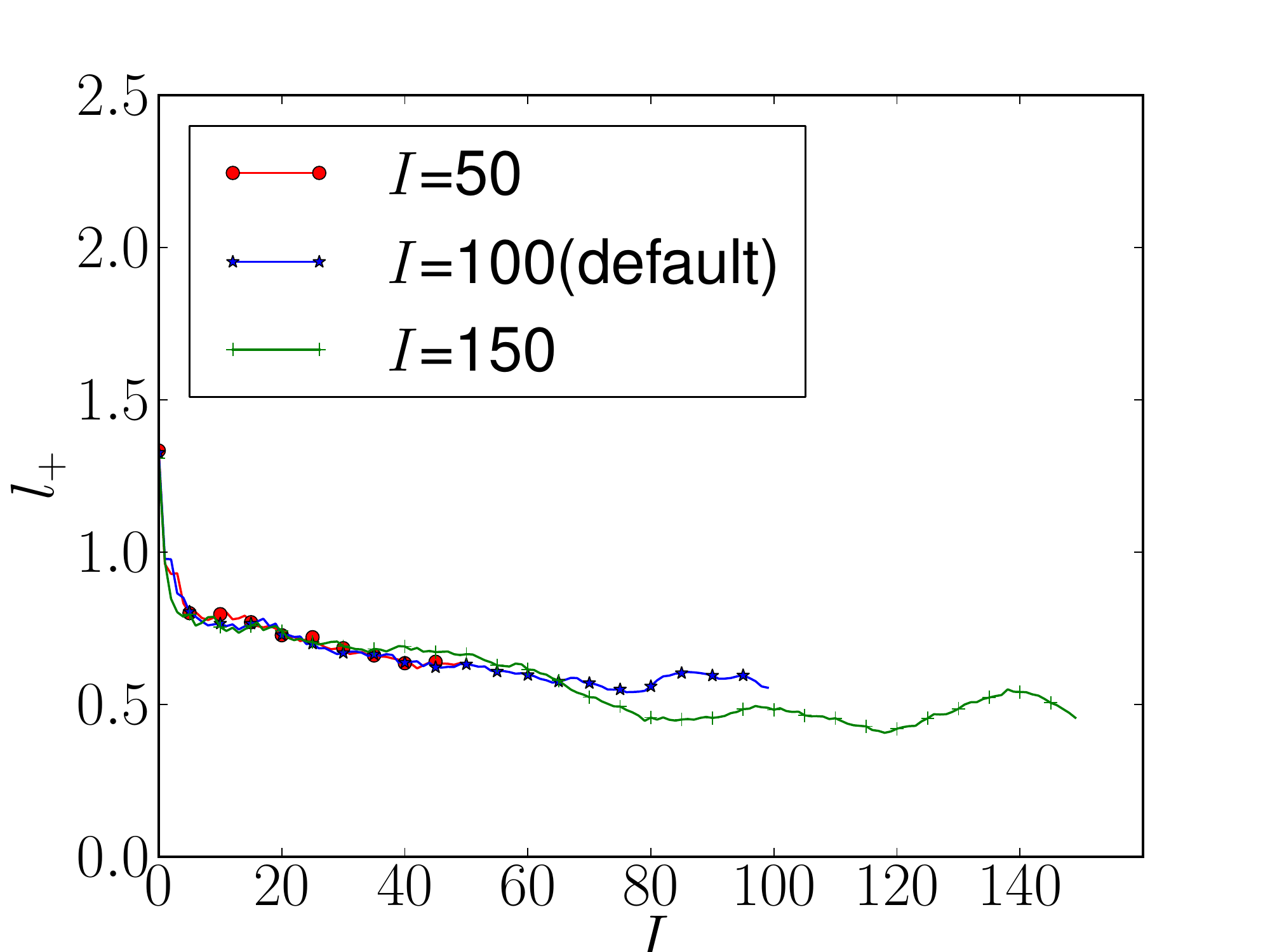}}
\hspace*{-1.5em}
\subfigure[$l_+$ vs $E$]{\label{fig:subfigparasweep:episode} %% label for first subfigure
\includegraphics[width=0.35\textwidth]{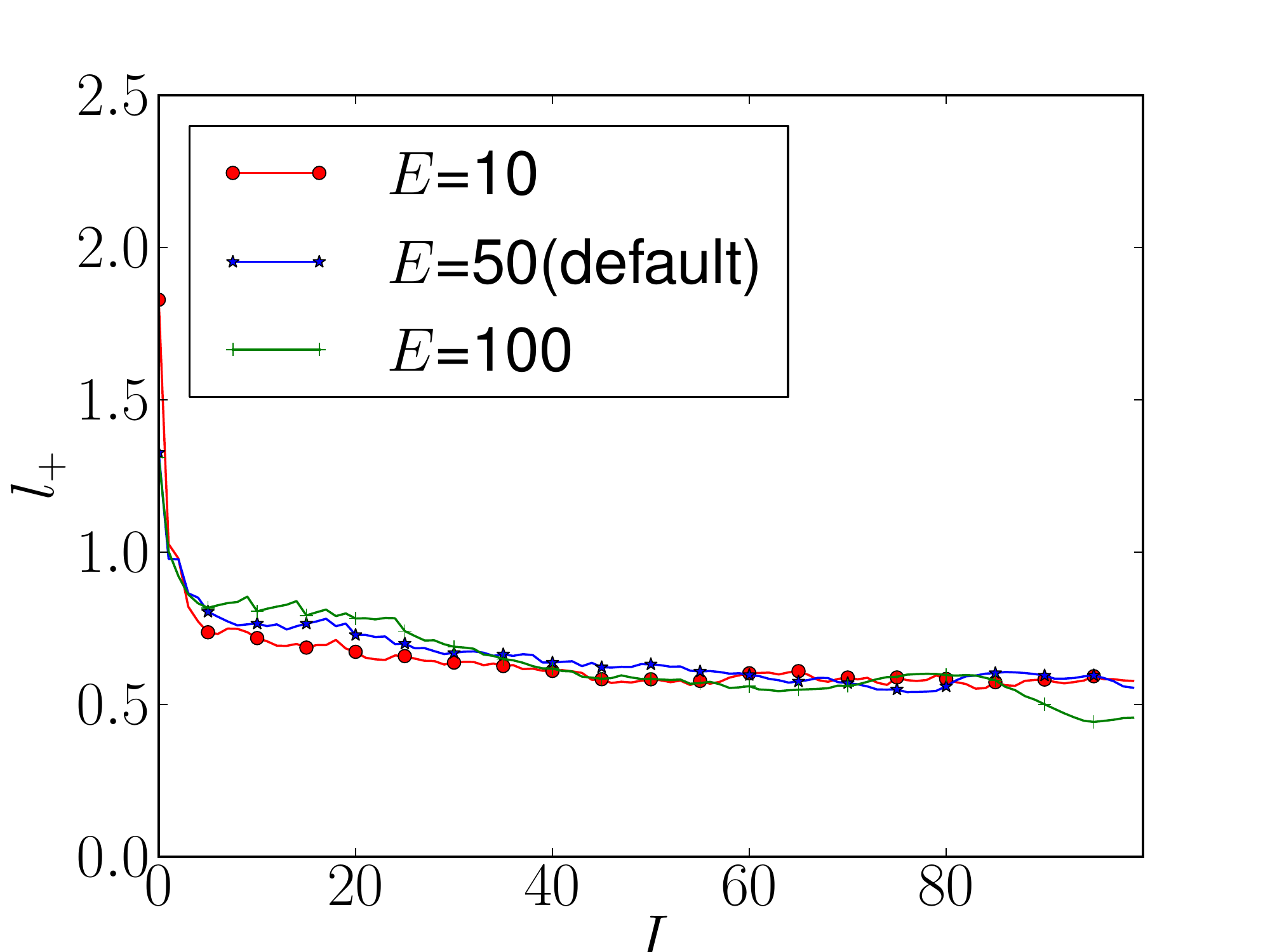}}
\hspace*{-1.5em}
\subfigure[$l_+$ vs $T'$]{\label{fig:subfigparasweep:tempThreshold} %% label for first subfigure
\includegraphics[width=0.35\textwidth]{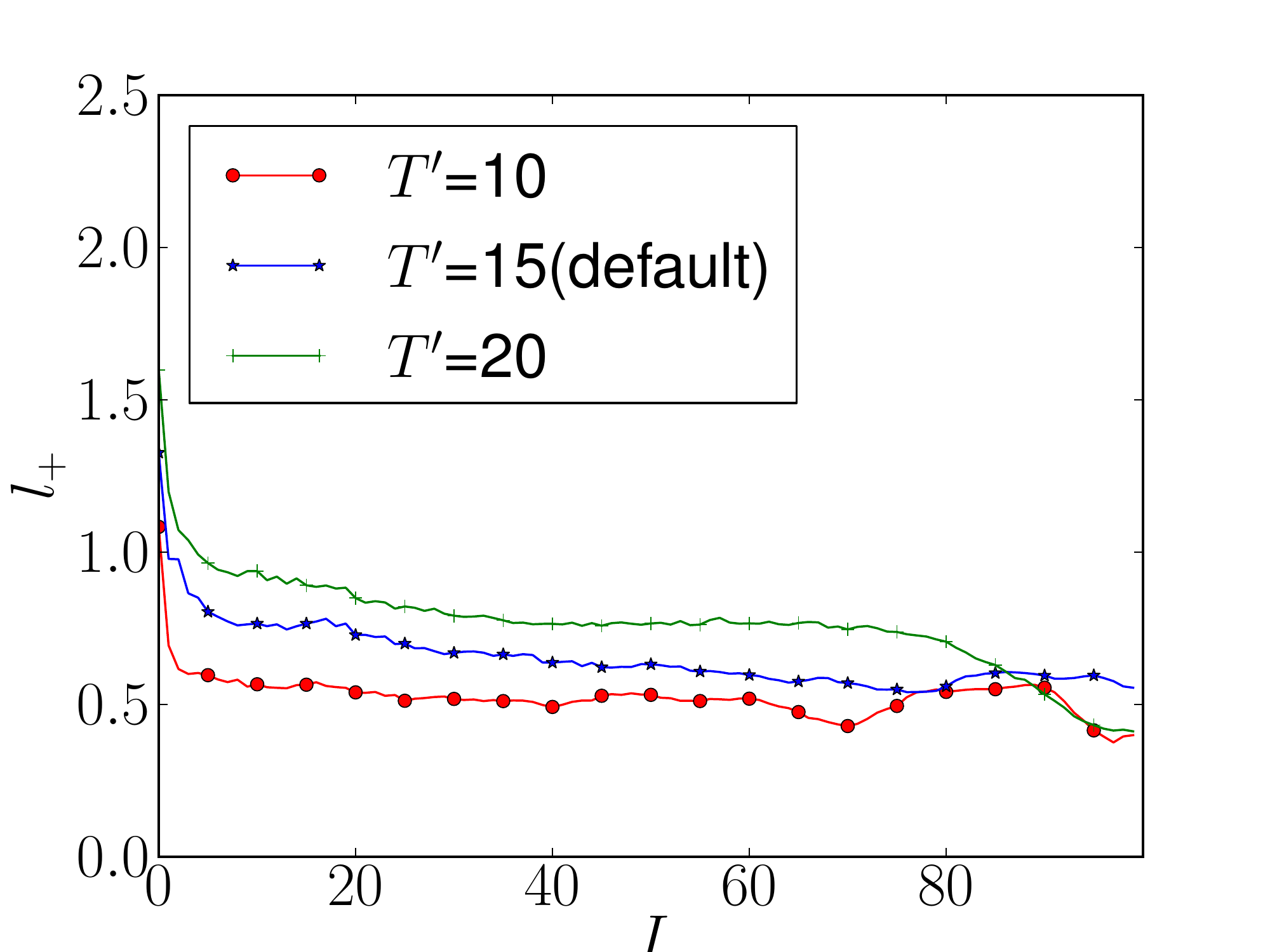}}
\hspace*{-2em}
\subfigure[$l_+$ vs $m$]{\label{fig:subfigparasweep:mctssimulation} %% label for first subfigure
\includegraphics[width=0.35\textwidth]{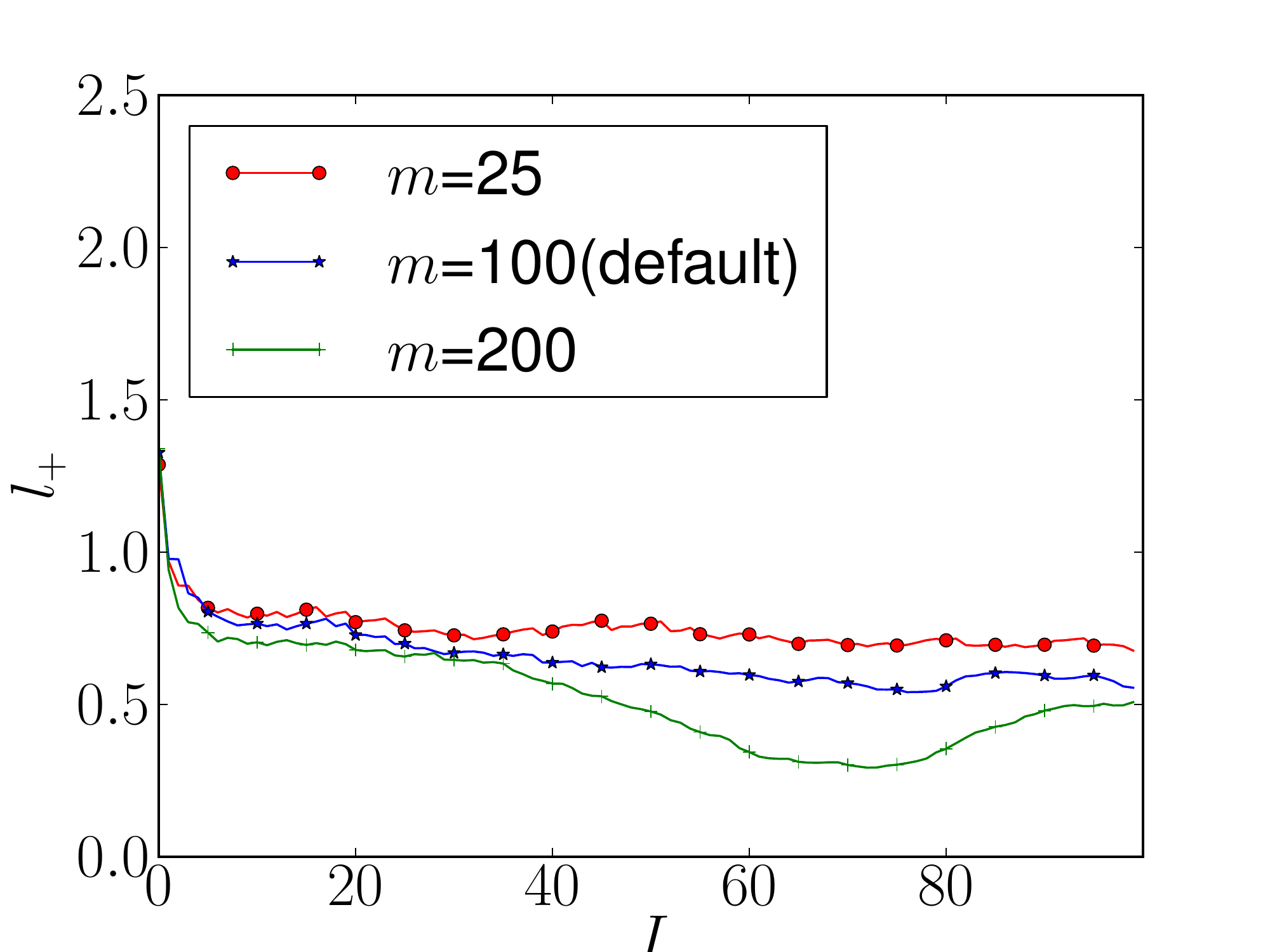}}
\hspace*{-1.5em}
\subfigure[$l_+$ vs $c$]{\label{fig:subfigparasweep:Cpuct} %% label for first subfigure
\includegraphics[width=0.35\textwidth]{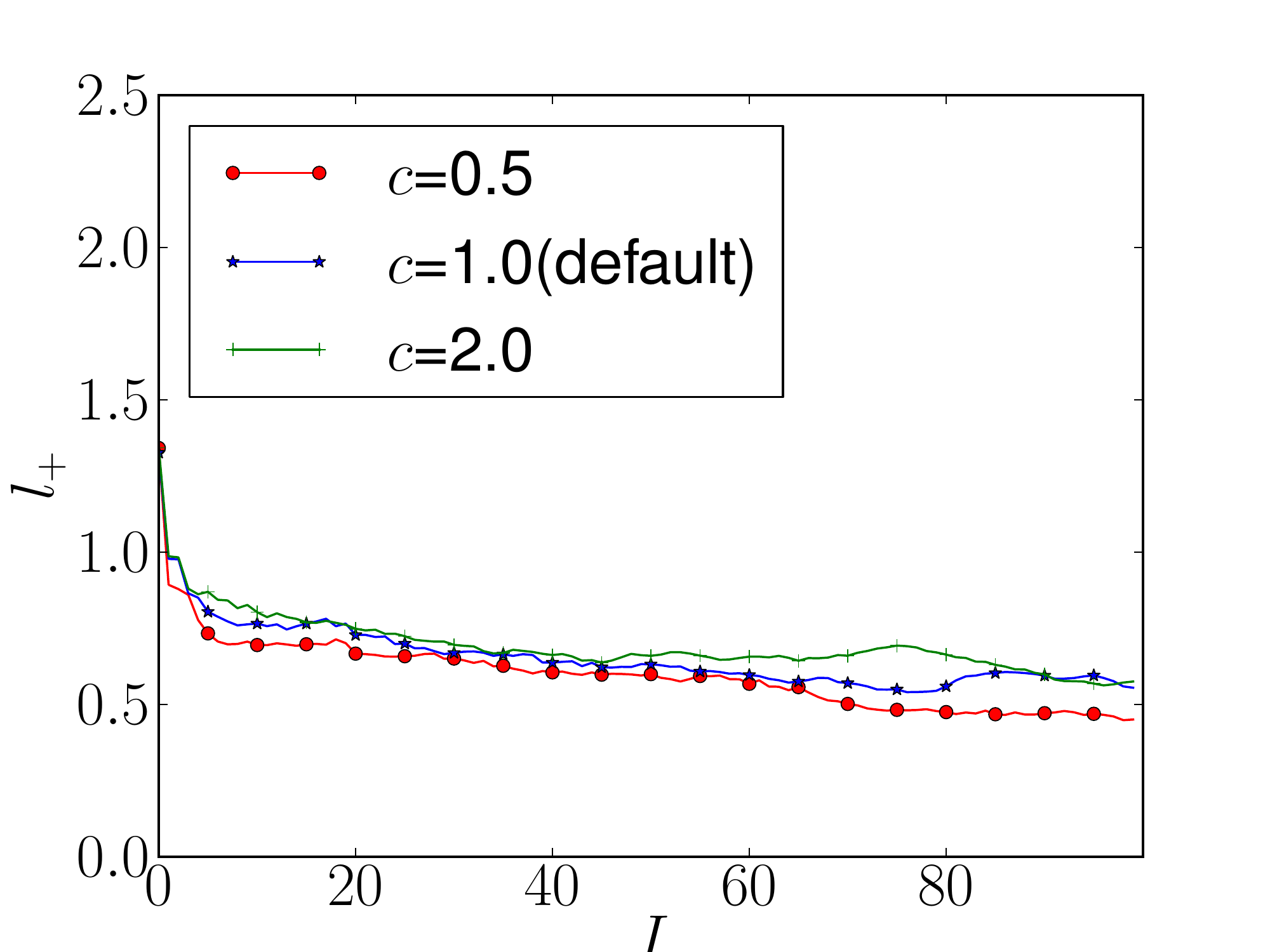}}
\hspace*{-1.5em}
\subfigure[$l_+$ vs $rs$]{\label{fig:subfigparasweep:retrainlength} %% label for first subfigure
\includegraphics[width=0.35\textwidth]{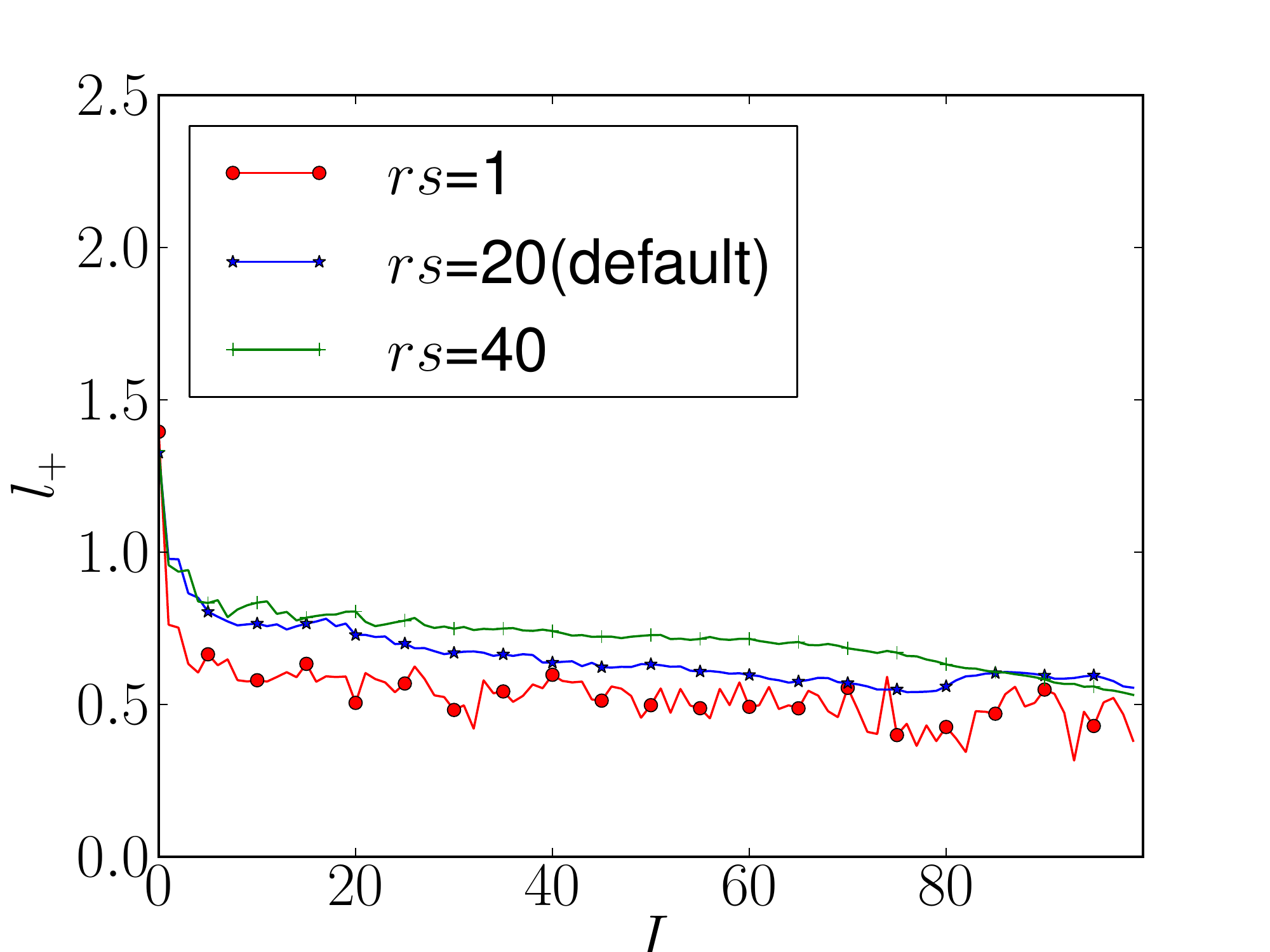}}
\hspace*{-2em}
\subfigure[$l_+$ vs $ep$]{\label{fig:subfigparasweep:epoch} %% label for first subfigure
\includegraphics[width=0.35\textwidth]{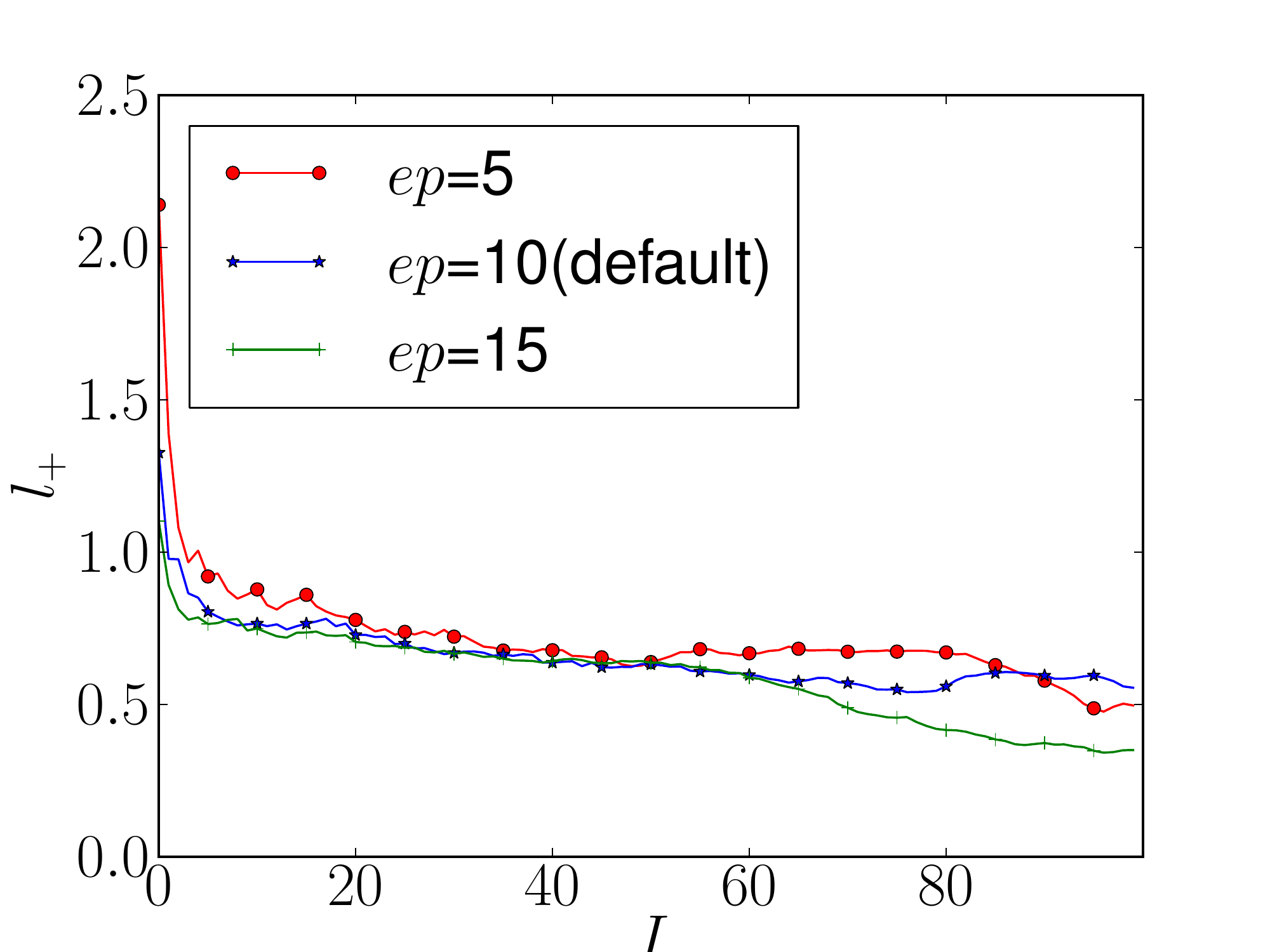}}
\hspace*{-1.5em}
\subfigure[$l_+$ vs $bs$]{\label{fig:subfigparasweep:batchsize} %% label for first subfigure
\includegraphics[width=0.35\textwidth]{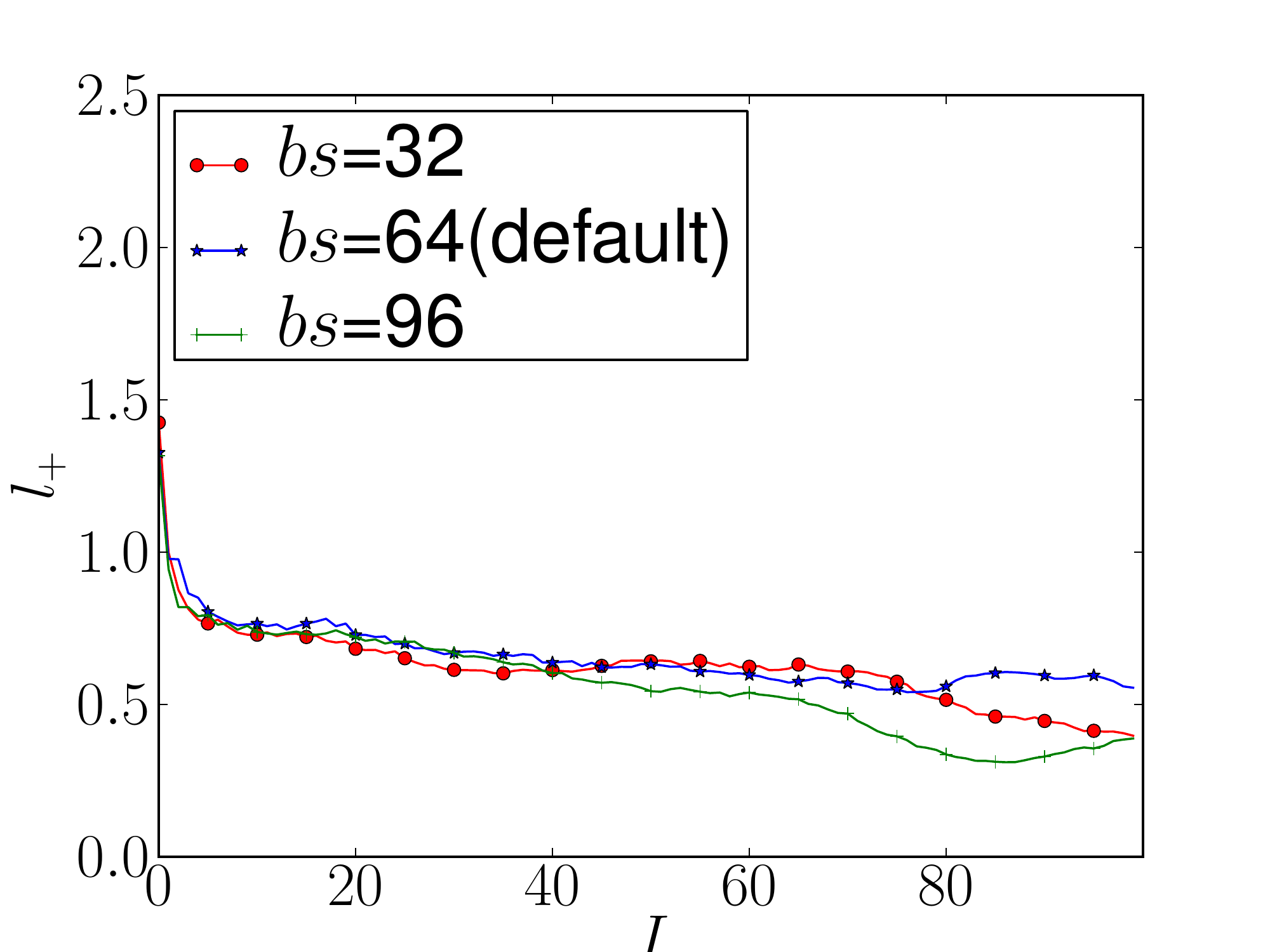}}
\hspace*{-1.5em}
\subfigure[$l_+$ vs $lr$]{\label{fig:subfigparasweep:learningrate} %% label for first subfigure
\includegraphics[width=0.35\textwidth]{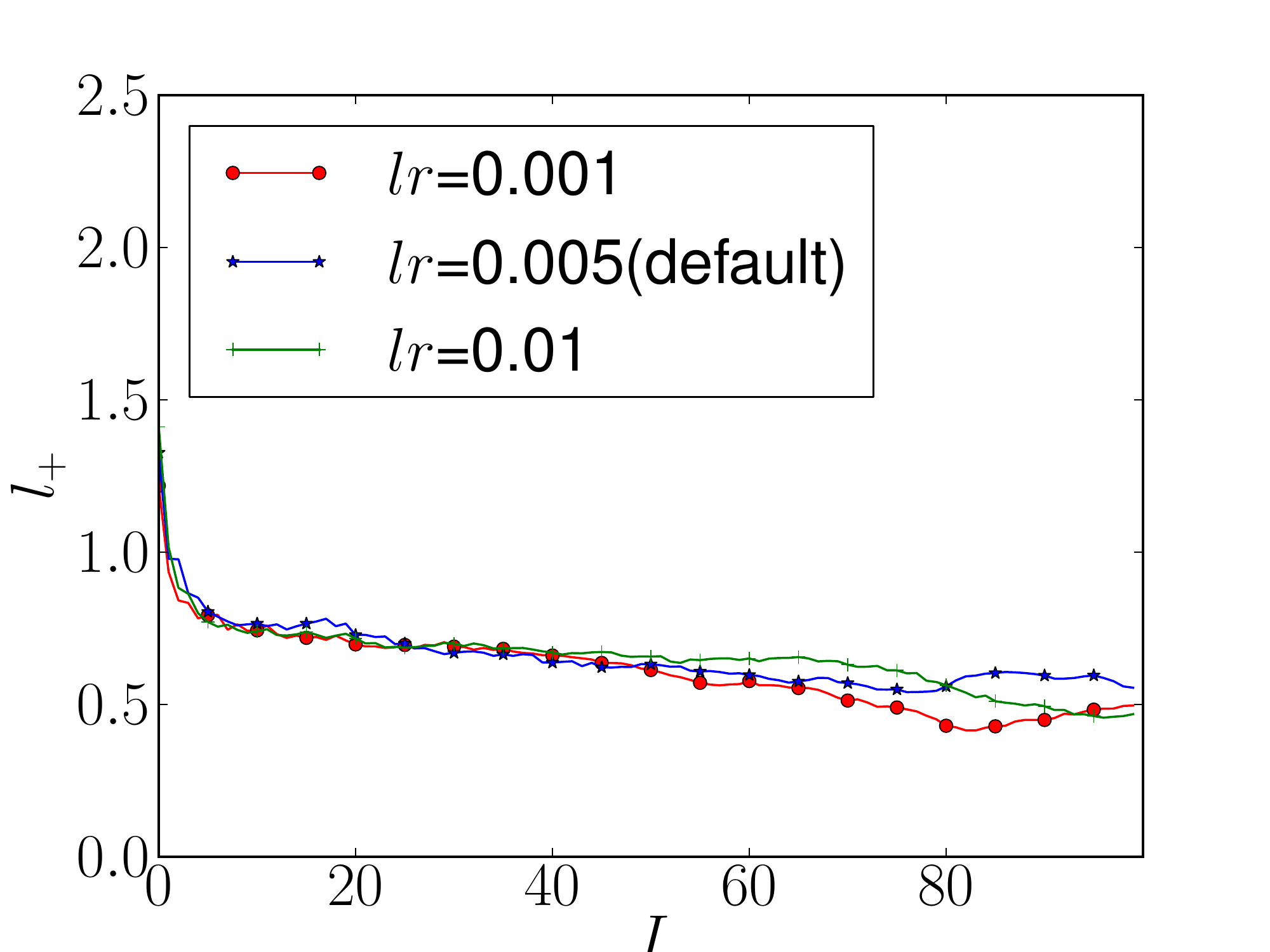}}
\hspace*{-2em}
\subfigure[$l_+$ vs $d$]{\label{fig:subfigparasweep:dropout} %% label for first subfigure
\includegraphics[width=0.35\textwidth]{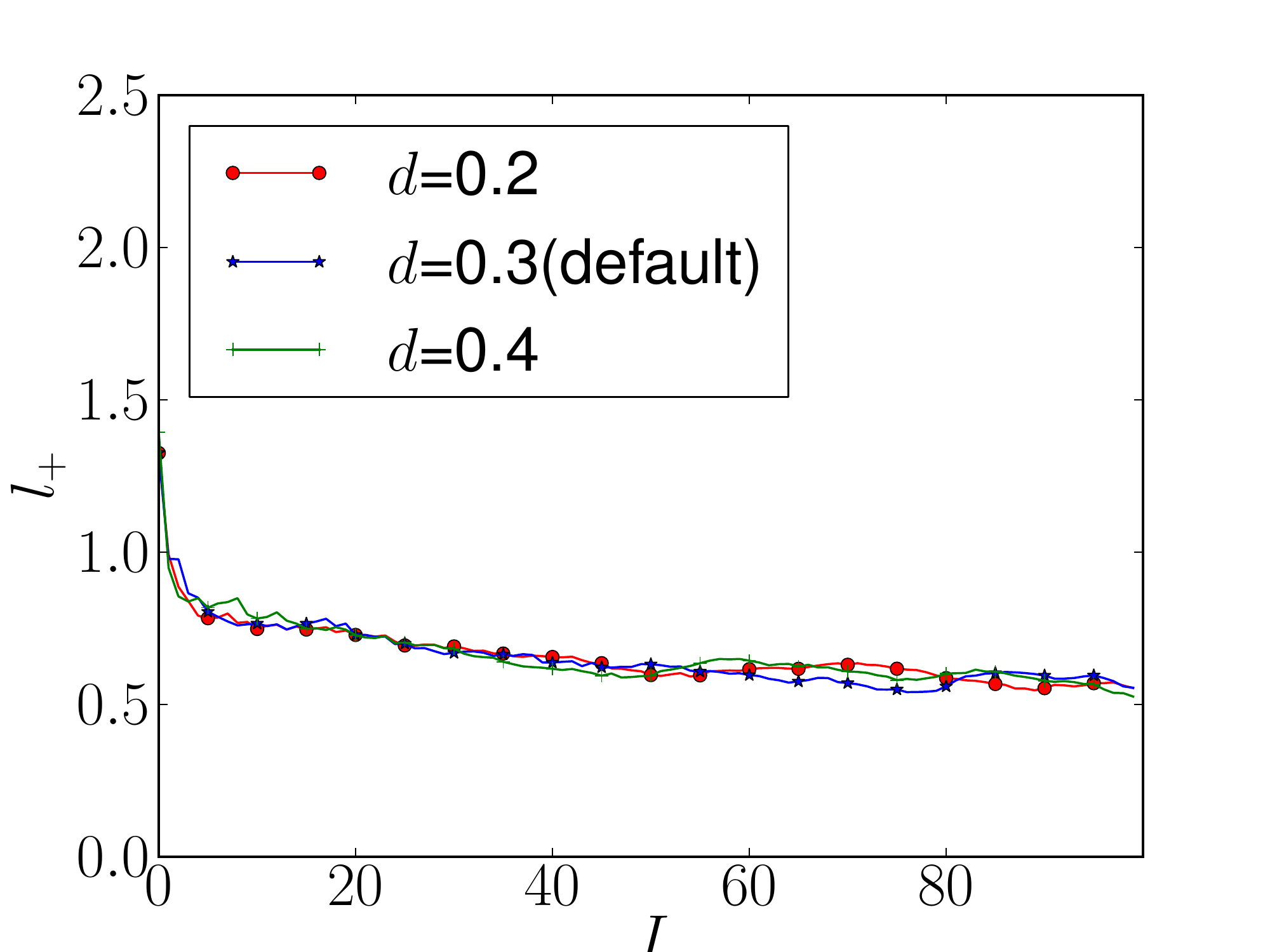}}
\hspace*{-1.5em}
\subfigure[$l_+$ vs $n$]{\label{fig:subfigparasweep:arenacompare} %% label for first subfigure
\includegraphics[width=0.35\textwidth]{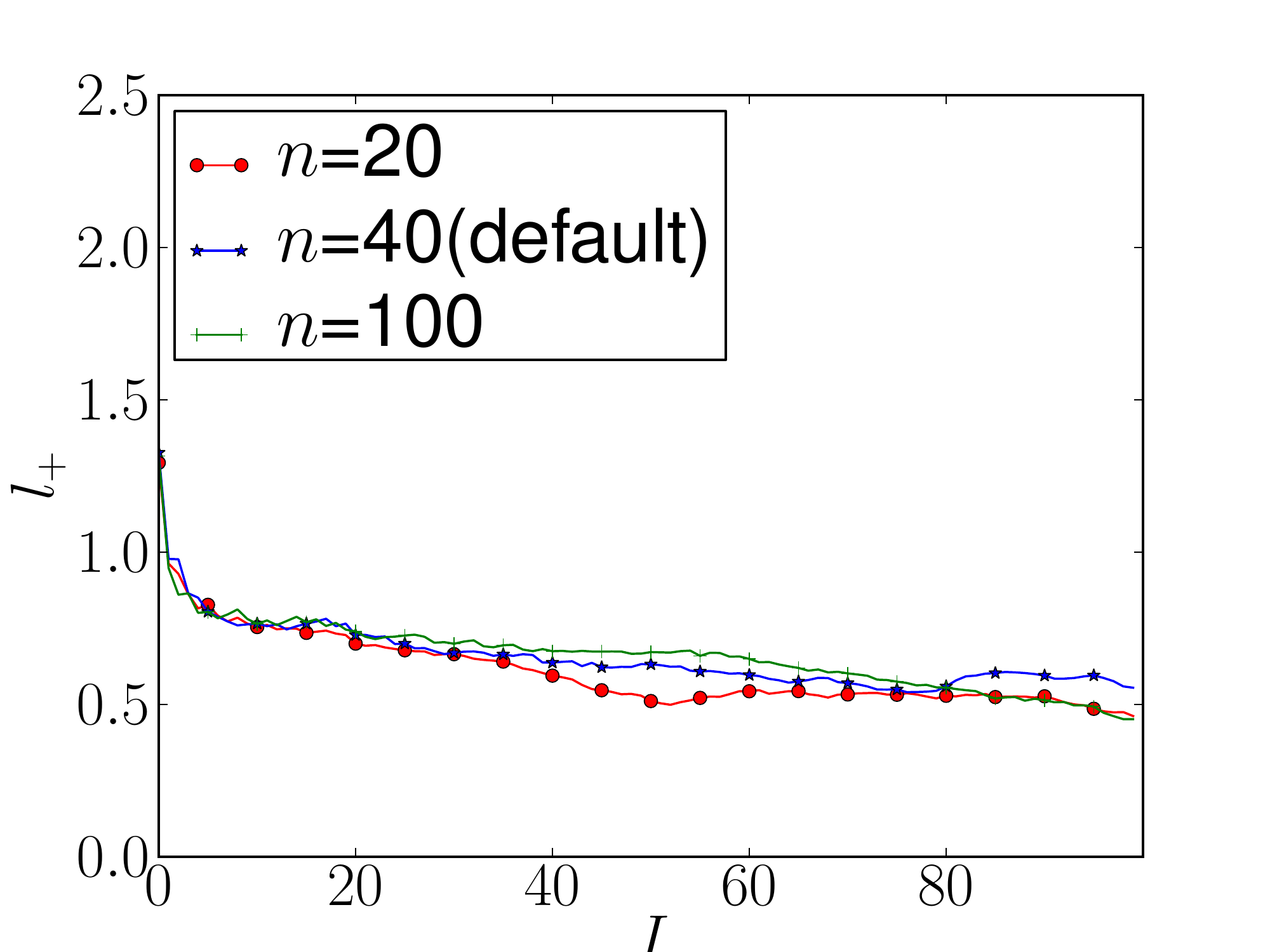}}
\hspace*{-1.5em}
\subfigure[$l_+$ vs $u$]{\label{fig:subfigparasweep:updateThreshold} %% label for first subfigure
\includegraphics[width=0.35\textwidth]{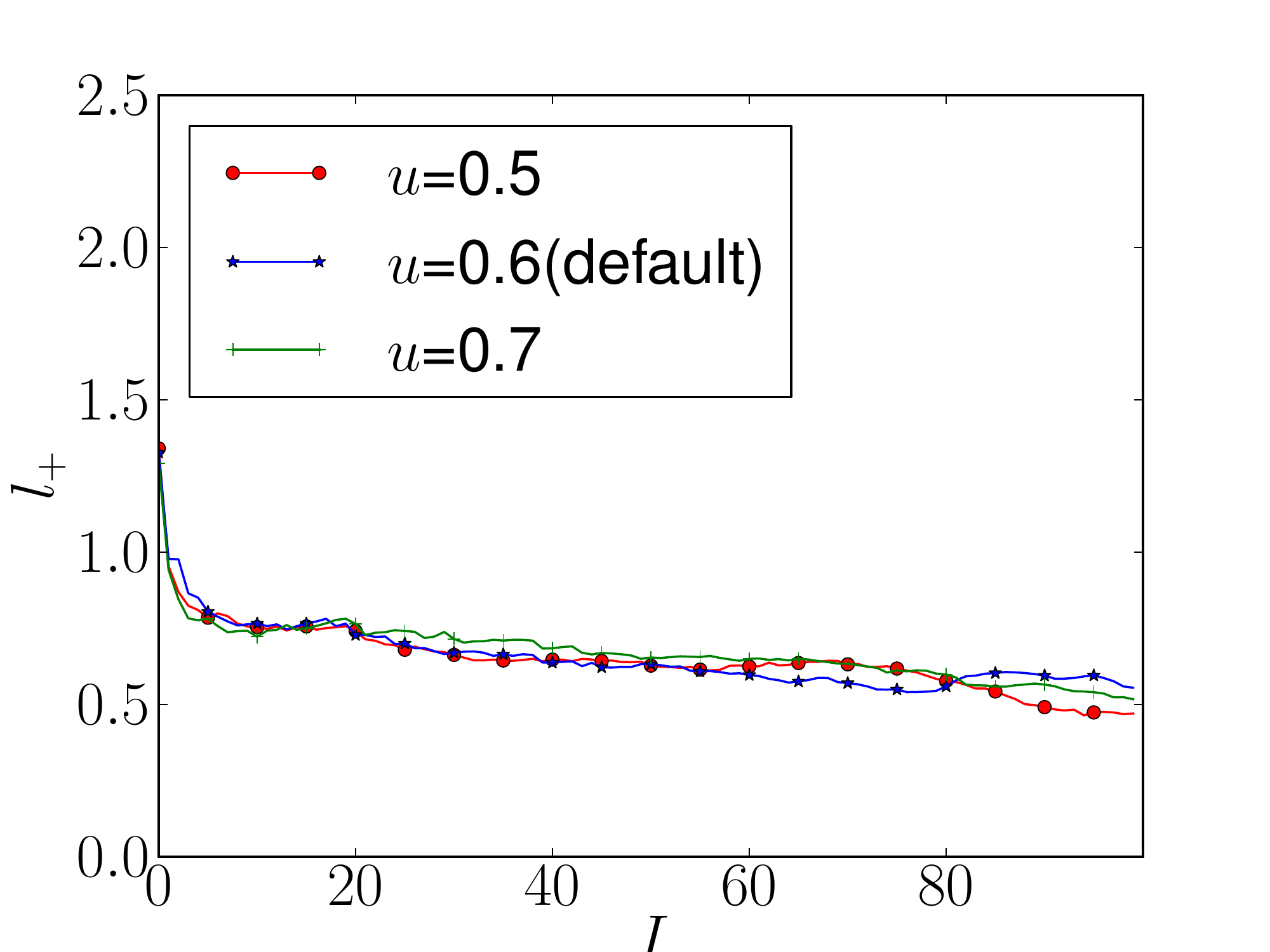}}
\caption{Training loss for different parameter settings over iterations}
\label{fig:subfigparasweeploss} %% label for entire figure
\end{figure}

\textbf{\emph{I:}} In order to find a good value for \emph{I} (iterations), we train 3 different models to play 6$\times$6 Othello by setting \emph{I} at minimum, default and maximum value respectively. We keep the other hyper-parameters at their default values.  Fig.~\ref{fig:subfigparasweep:iteration} shows that training loss decreases to a relatively stable level. However, after iteration 120, the training loss unexpectedly increases to the same level as for iteration 100 and further decreases. This surprising behavior could be caused by a too high learning rate, an improper update threshold, or overfitting. This is an unexpected result since in theory more iterations lead to better performance. %This illustrates the importance of proper hyper-parameter values.

\textbf{\emph{E:}} Since more episodes mean more training examples, it can be expected that more training examples lead to more accurate results. However, collecting more training examples also needs more resources. This shows again that hyper-parameter optimization is necessary to find a reasonable value of for {\em E}. In Fig.~\ref{fig:subfigparasweep:episode}, for \emph{E}=100, the training loss curve is almost the same as the 2 other curves for a long time before eventually going down.

\textbf{\emph{T':}} The step threshold controls when to choose a random action or the one suggested by MCTS. This parameter controls exploration in self-play, to prevent  deterministic policies from generating training examples. Small \emph{T'} results in more deterministic policies, large \emph{T'} in policies more different from the model. In Fig.~\ref{fig:subfigparasweep:tempThreshold}, we see that \emph{T'}=10 is a good value.

\textbf{\emph{m:}} In theory, more MCTS simulations \emph{m} should provide better policies. However, higher \emph{m} requires more time to get such a policy.  Fig.~\ref{fig:subfigparasweep:mctssimulation} shows that a value for 200 MCTS simulations achieves the best performance in the 70th iteration,  then has a drop, to reach a similar level as 100 simulations in  iteration 100.

\textbf{\emph{c:}} This hyper-parameter $C_p$ is used to balance the exploration and exploitation during tree search. It is often set at 1.0. However, in Fig.~\ref{fig:subfigparasweep:Cpuct}, our experimental results show that more exploitation~(\emph{c}=0.5) can provide smaller training loss. 

\textbf{\emph{rs:}} In order to reduce overfitting, it is important to retrain models using previous training examples. Finding a good retrain length of historical training examples is necessary to reduce training time. In Fig.~\ref{fig:subfigparasweep:retrainlength}, we see that using training examples from the most recent single previous iteration achieves the smallest training loss. This is an unexpecrted result, suggesting that overfitting is prevented by other means and that the time saving works out best overall. 

\textbf{\emph{ep:}} The training loss of different \emph{ep} is shown in Fig.~\ref{fig:subfigparasweep:epoch}. For  \emph{ep}=15 the training loss  is the lowest. This result shows that along with the increase of epoch, the training loss decreases, which is as expected.

\textbf{\emph{bs:}} a smaller batch size \emph{bs} increases the number of batches, leading to higher time cost. However, smaller \emph{bs} means less training examples in each batch, which may cause more fluctuation~(larger variance) of training loss.  Fig.~\ref{fig:subfigparasweep:batchsize} shows that \emph{bs}=96 achieves the smallest training loss in iteration 85.

\textbf{\emph{lr:}} In order to avoid skipping over optima, a small learning rate is generally suggested. However, a smaller learning rate learns~(accepts) new knowledge slowly. In Fig.~\ref{fig:subfigparasweep:learningrate}, \emph{lr}=0.001 achieves the lowest training loss around iteration 80.

\textbf{\emph{d:}} Dropout is a popular method to prevent overfitting. Srivastava et al. claim that dropping out 20\% of the input units and 50\% of the hidden units is often found to be good~\cite{Srivastava2014}. In Fig.~\ref{fig:subfigparasweep:dropout}, however, we can not see a significant difference. 

\textbf{\emph{n:}} The number of games in the arena comparison is a key factor of time cost. A small value may miss accepting good new models and too large a value is a waste of time. Our experimental results in Fig.~\ref{fig:subfigparasweep:arenacompare} show that there is no significant difference. A combination with \emph{u} can be used to determine the acceptance or rejection of a newly learnt model. In order to reduce time cost, a  small \emph{n} combined with a  large \emph{u} may be a good choice.

\textbf{\emph{u:}} This hyper-parameter is the update threshold. Normally, in two-player games, player A is better than  player B if it wins more than 50\% games. A higher threshold avoids fluctuations.  However, if we set it too high, it becomes too difficult to  accept better models. Fig.~\ref{fig:subfigparasweep:updateThreshold} shows that \emph{u}=0.7 is too high, 0.5 and 0.6 are acceptable.

\begin{table}[H]
\centering\hspace*{-2.3em}
%\linebreak
\caption{Time Cost~(hr) of Different Parameter Setting}\label{timecosttab}
\begin{tabular}{|l|l|l|l|l|}
\hline
Parameter& Minimum	& Default& Maximum& Type\\
\hline
\emph{I}	&\textbf{23.8}	&44.0	&60.3&time-sensitive\\
\hline
\emph{E}	&\textbf{17.4}	&44.0&87.7&time-sensitive\\
\hline
\emph{T'}	&41.6		&44.0&40.4&time-friendly\\
\hline
\emph{m}	&\textbf{26.0}	&44.0&	64.8&time-sensitive\\
\hline
\emph{c}	&50.7	&44.0&	49.1&time-friendly\\
\hline
\emph{rs}	&\textbf{26.5}	&44.0&50.7&time-sensitive\\
\hline
\emph{ep}	&\textbf{43.4}	&44.0	&55.7&time-sensitive\\
\hline
\emph{bs}	&47.7	&44.0&\textbf{37.7}&time-sensitive\\
\hline
\emph{lr}	&47.8&	44.0&40.3&time-friendly\\
\hline
\emph{d}&	51.9&44.0&51.4&time-friendly\\
%\hline
%channel&	44.1&	44.0&70.9\\
\hline
\emph{n}	&\textbf{33.5}&44.0	&57.4&time-sensitive\\
\hline
\emph{u}	&39.7	&44.0 &	40.4&time-friendly\\
\hline
\end{tabular}
\end{table}

To investigate the impact on running time, we present the effect of different values for each hyper-parameter in Table~\ref{timecosttab}. We see that for parameter \emph{I}, \emph{E}, \emph{m}, \emph{rs}, \emph{n}, smaller values lead to quicker training, which is as expected. For \emph{bs}, larger values result in quicker training. The other hyper-parameters are indifferent, changing their values will not lead to significant changes in training time. Therefore, tuning these hyper-parameters shall reduce training time or achieve better quality in the same time.

Based on the aforementioned results and  analysis, we summarize the importance by evaluating contributions of each parameter to training loss and time cost, respectively, in Table~\ref{losstimetab}~(best values in bold font). For training loss, different values of \emph{n} and \emph{u} do not result in a significant difference. Modifying time-indifferent hyper-parameters does not much change training time, whereas larger value of time-sensitive hyper-parameters lead to higher time cost.
\begin{table}[H]
\centering\hspace*{-2.3em}
%\linebreak
\caption{A Summary of Importance in Different Objectives}\label{losstimetab}
\begin{tabular}{|l|l|l|l|l|}
\hline
Parameter	&Default Value &	Loss  & Time Cost\\
\hline
\emph{I}	& 100&100&\textbf{50}\\
\hline
\emph{E}	& 50 &\textbf{10}&\textbf{10}\\
\hline
\emph{T'}	&15 &\textbf{10}&similar\\
\hline
\emph{m}	& 100&\textbf{200}&\textbf{25}\\
\hline
\emph{c}	& 1.0&\textbf{0.5}&similar\\
\hline
\emph{rs}	&20 &\textbf{1}&\textbf{1}\\
\hline
\emph{ep}	& 10&\textbf{15}&\textbf{5}\\
\hline
\emph{bs}	& 64&\textbf{96}&\textbf{96}\\
\hline
\emph{lr} &0.005 &\textbf{0.001}&similar\\
\hline
\emph{d}   &0.3 &0.3&similar\\
\hline
\emph{n}	& 40&insignificant&\textbf{20}\\
\hline
\emph{u}	&0.6&insignificant&similar\\
\hline
\end{tabular}
\end{table}

\subsection{Hyper-Parameter Correlation Evaluation Results}\label{correlationevaluationresults}
In this part, we investigate the correlation between promising hyper-parameters in terms of time cost and playing strength. There are $3^4=81$ final best players trained based on 3 different values of 4 hyper-parameters~(\emph{I}, \emph{E}, \emph{m} and \emph{ep}) in Table~\ref{correlationparatab} plus a random player~(i.e. 82 in total). Any 2 of these 82 players play with each other. Therefore, there are 82$\times$81/2=3321 pairs, and for each of these, 10 games are played.

\begin{figure}[tbh]
\centering
\hspace*{-2.3em}
\includegraphics[width=1\textwidth]{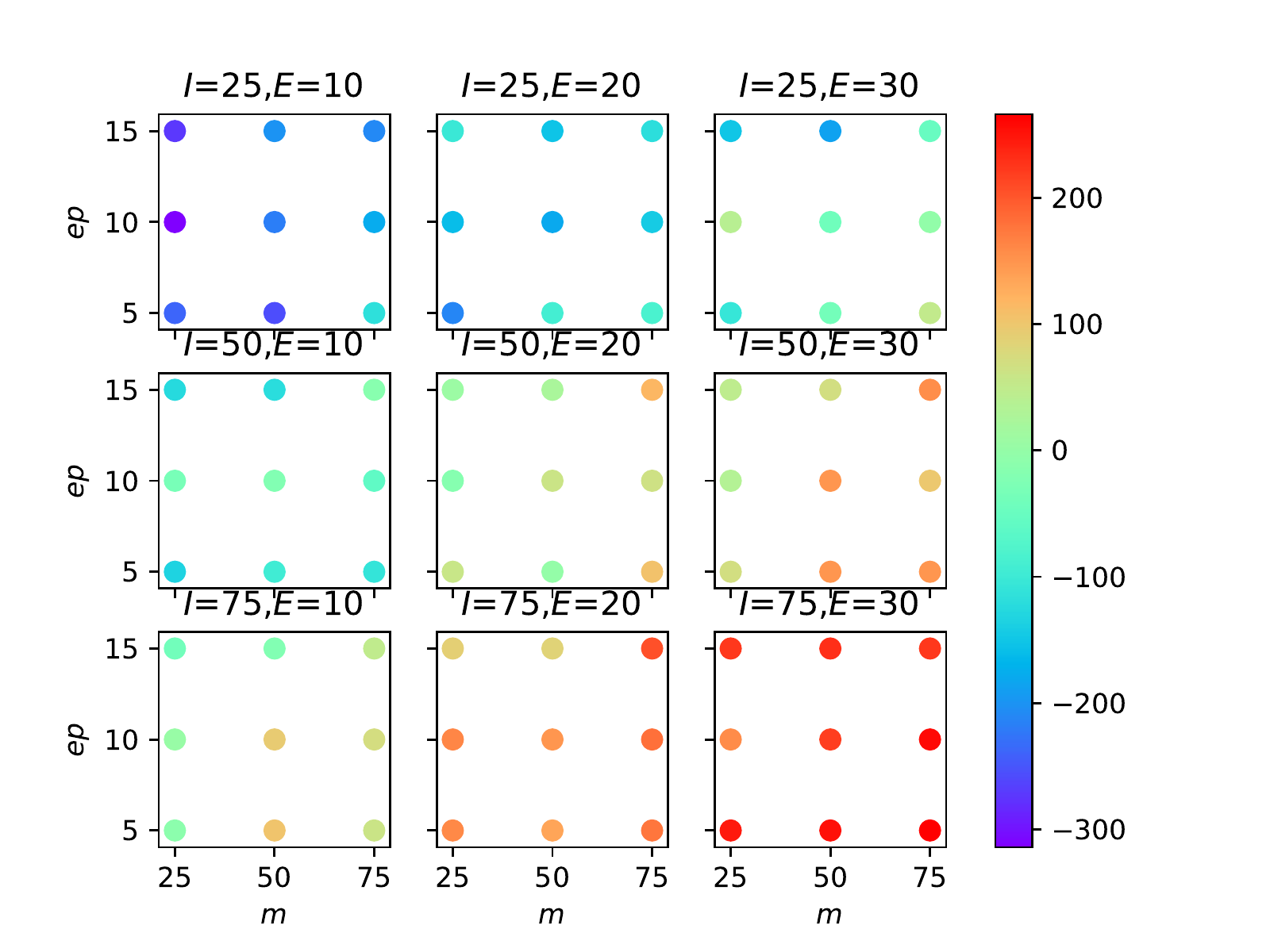}
\caption{Elo ratings of the final best players of the full tournament (3 parameters, 1 target value)}
\label{fig:fig2delosubt}
\end{figure}

In each sub-figures of Fig.~\ref{fig:fig2delosubt}, all models are trained from the same value of \emph{I} and \emph{E}, according to the different values in x-axis and y-axis, we find that, generally, larger \emph{m} and larger \emph{ep} lead to higher Elo ratings. However, in the last sub-figure, we can clearly notice that the Elo rating of \emph{ep}=10 is higher than that of \emph{ep}=15 for \emph{m}=75, which shows that sometimes more training can not improve the playing strength but decreases the training performance. We suspect that this is caused by overfitting. Looking at the sub-figures, the results also show that more (outer) training iterations can significantly improve the playing strength, also more  training examples in each iteration~(bigger \emph{E}) helps. These outer iterations are clearly more important than optimizing the inner hyper-parameters of {\em m} and {\em ep}. Note that higher values for the outer hyper-parameters imply more MCTS simulations and more training epochs, but not vice versa. This is an important insight regarding tuning hyper-parameters for self-play.

According to (\ref{timecostfunction}) and Table.~\ref{losstimetab}, we know that smaller values of time sensitive hyper-parameters result in quicker training. However, some time sensitive hyper-parameters influence the training of better models. Therefore, we analyze training time versus Elo rating of the hyper-parameters, to achieve the best training performance for a fixed time budget.

\begin{figure}[tbh]
\centering
\hspace*{-2.3em}
\includegraphics[width=0.8\textwidth]{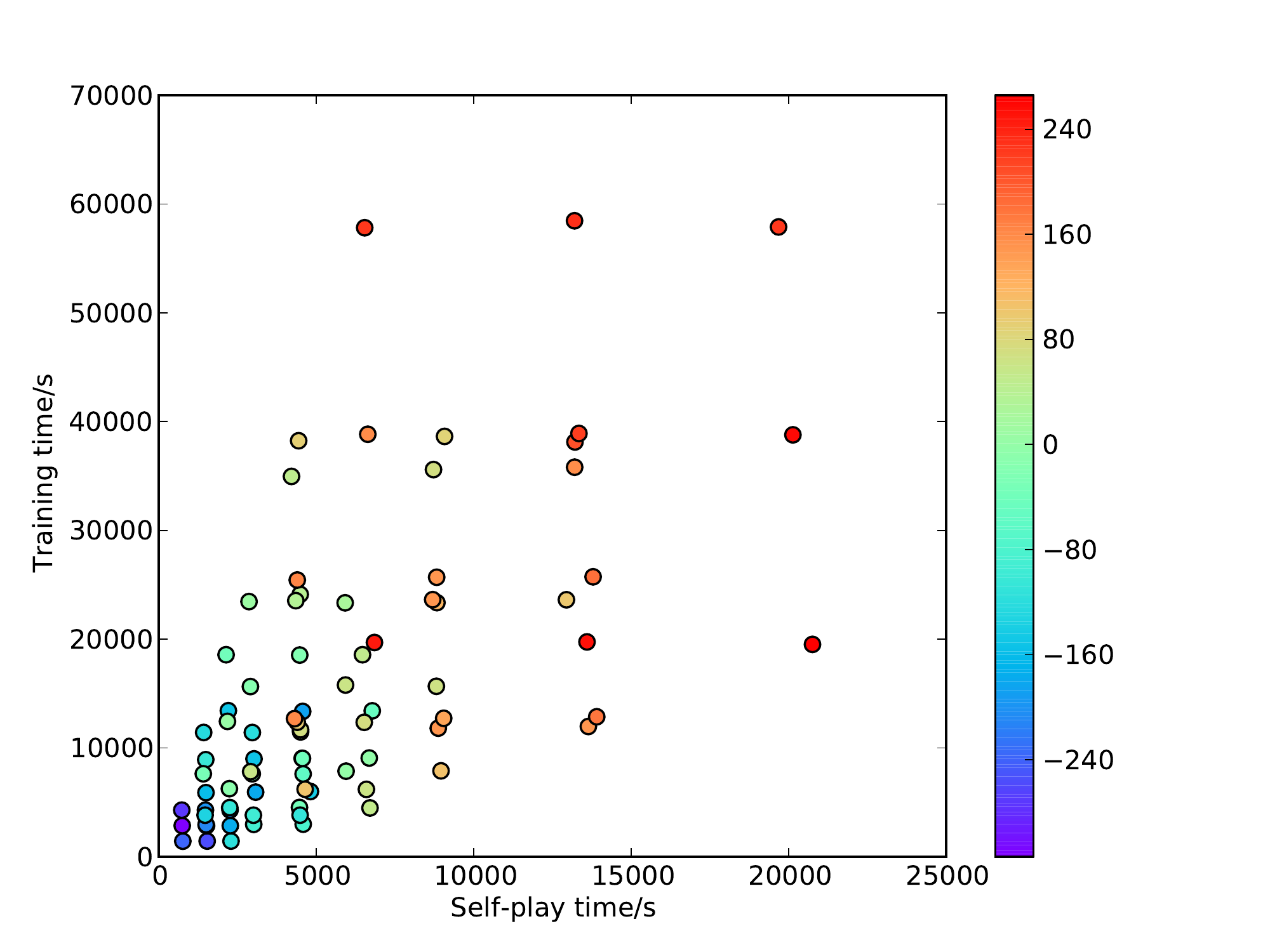}
\caption{Elo ratings of the final best players with different time cost of Self-play and neural network training (same base data as in Fig.~\ref{fig:fig2delosubt})}
\label{fig:figelowithtimecost} %% label for entire figure
\end{figure}

In order to find a way to assess the relationship between time cost and Elo ratings, we categorize the time cost into two parts, one part is the self-play ~(stage 1 in Algorithm~\ref{alg:a0g}, iterations and episodes) time cost, the other is the training part~(stage 2 in Algorithm~\ref{alg:a0g}, training epochs). In general, spending more time in training and in self-play gives higher Elo. In self-play time cost, there is also an other interesting variable, searching time cost, which is influenced by the value of \emph{m}.

In Fig.~\ref{fig:figelowithtimecost} we also find high Elo points closer to the origin, confirming that high Elo combinations of low self-play time and low training time exist, as was indicated above, by choosing low epoch {\em ep} and simulation {\em m} values, since the outer iterations already imply adequate training and simulation.   %going from left to right, we find that, more self-play for generating training data can improve the Elo ratings. However, in vertical comparison, more points show that more training time does not always get higher Elo ratings, which indicates that setting proper value for training part can save much time. More specifically, the yellow point closing to x-axis and y-axis provides us more deep information of using similar time budget to achieve the best training performance by setting proper parameter values.

\begin{figure}[H]
\centering
\hspace*{-1.8em}
\subfigure[Self-play time vs Elo]{\label{fig:subfigelowithselfplaytime} %% label for first subfigure
\includegraphics[width=0.5\textwidth]{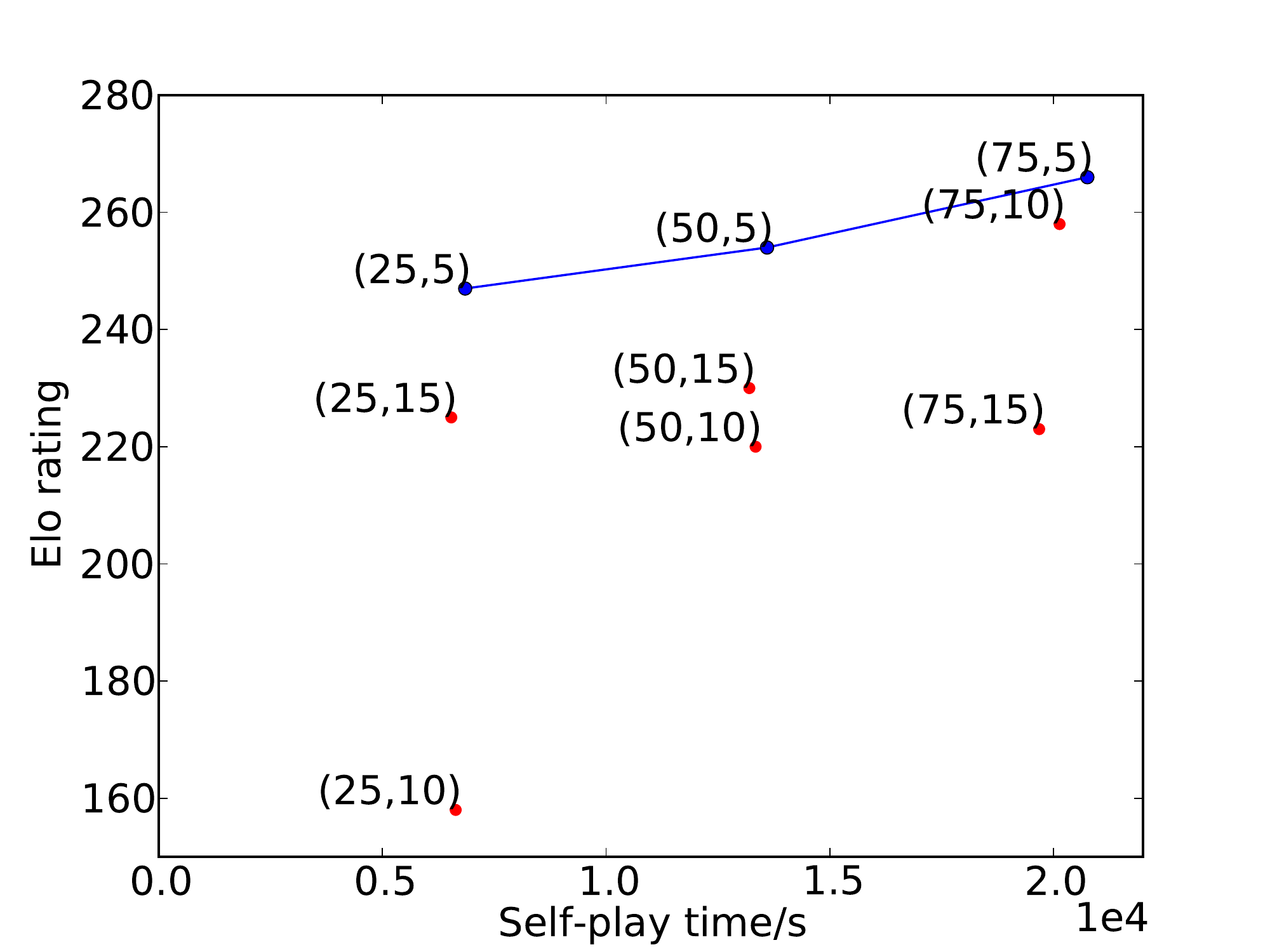}}
\hspace*{-2.4em}
%\subfigure[Training time vs Elo]{\label{fig:subfigelowithtrainingtime} %% label for first subfigure
%\includegraphics[width=0.45\textwidth]{tnntime2elo.pdf}}
%\hspace*{-2em}
\subfigure[Total time vs Elo]{\label{fig:subfigelowithtotaltime} %% label for first subfigure
\includegraphics[width=0.5\textwidth]{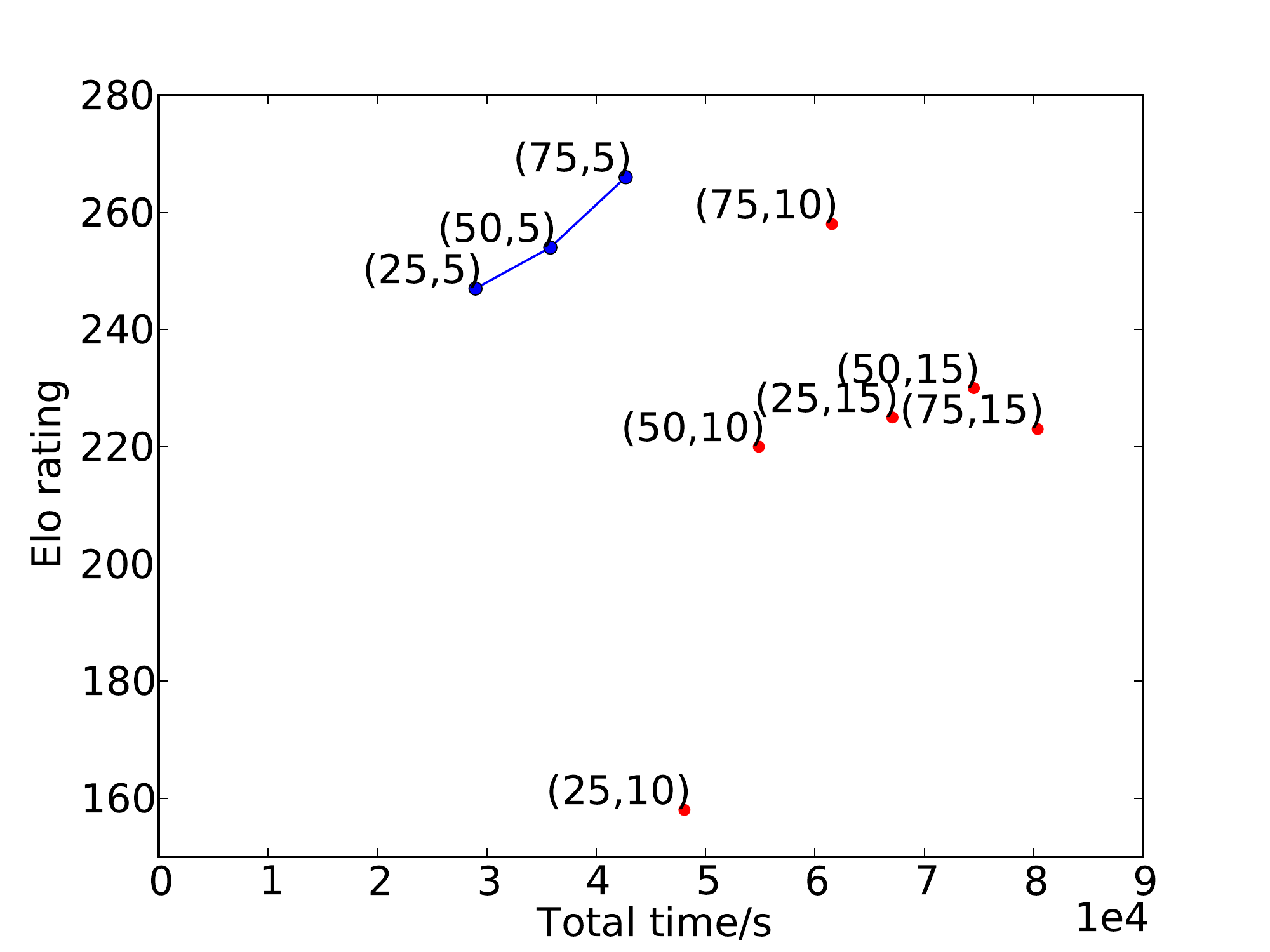}}
\caption{ Elo ratings of final best players to self-play, training and total time cost while $I$=75 and $E$=30.The values of tuple ($m$, $ep$) are given in the figures for every data point. In long total training, for $m$, larger values cost  more time and generally improve the playing strength. For $ep$, more training within one iteration does not show improvement for Elo ratings. The lines indicate the Pareto fronts of Elo rating vs. time.}
\label{fig:figsubtimecostwithelo}
\end{figure}

In order to further analyze the influence of self-play  and training on time, we present in Fig.~\ref{fig:subfigelowithselfplaytime} the full-tournament Elo ratings of  the lower right panel in Fig.~\ref{fig:fig2delosubt}. The blue line indicates the Pareto front of these combinations. We find that low  epoch values achieves the highest Elo in a high iteration training session: more outer self-play iterations implies more training epochs, and the data generated is more diverse such that training reaches more efficient stable state (no overfitting). 

\subsection{Alternative Loss Function Results}\label{lossfunctionresults}
In the following, we present the results of different loss functions. We have measured individual value loss, individual policy loss, the sum of thee two, and the product of the two, for the three games. We report training loss, the training Elo rating and the tournament Elo rating of the final best players. Error bars indicate standard deviation of 8 runs.

\subsubsection{Training Loss}\label{lossfunctionlossresults}

\begin{figure}[!tbh]
\centering
\hspace*{-1.5em}
\subfigure[Minimize $l_\textbf{p}$]{\label{fig:subfigloss55othello:a}
\includegraphics[width=0.45\textwidth]{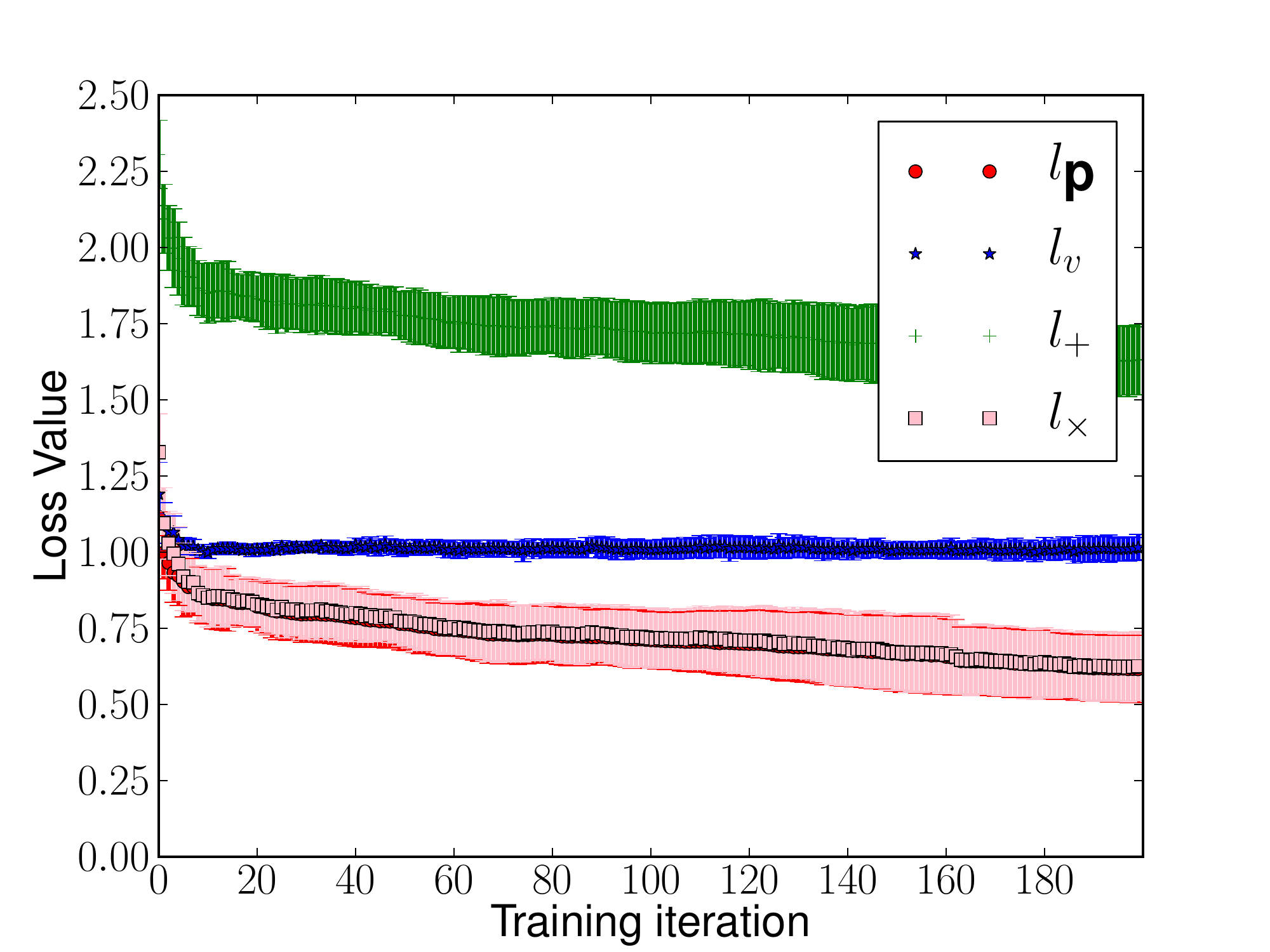}}
\hspace*{-1.8em}
\subfigure[Minimize $l_v$]{\label{fig:subfigloss55othello:b}
\includegraphics[width=0.45\textwidth]{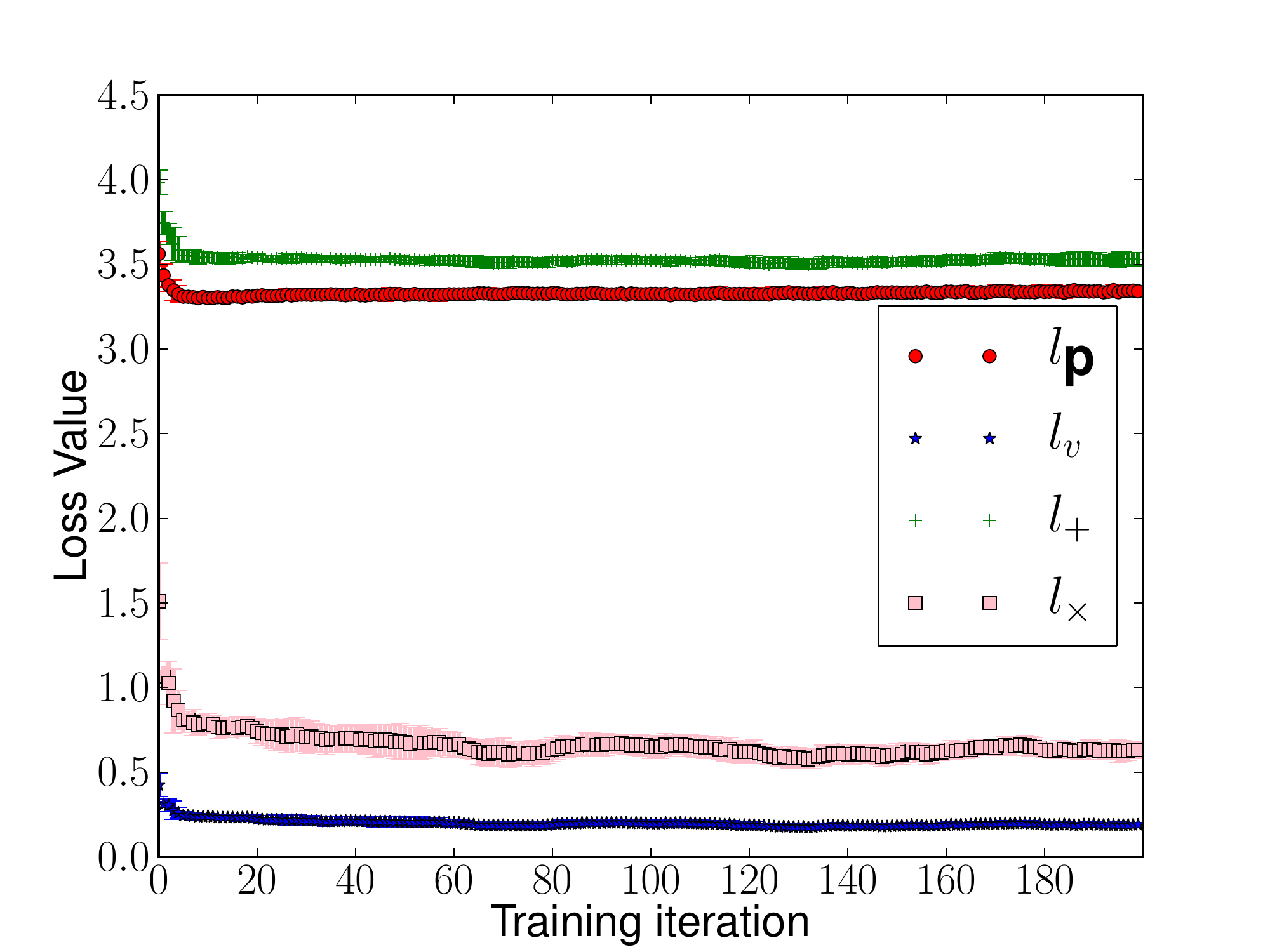}}
\hspace*{-1.5em}
\subfigure[Minimize $l_+$]{\label{fig:subfigloss55othello:c}
\includegraphics[width=0.45\textwidth]{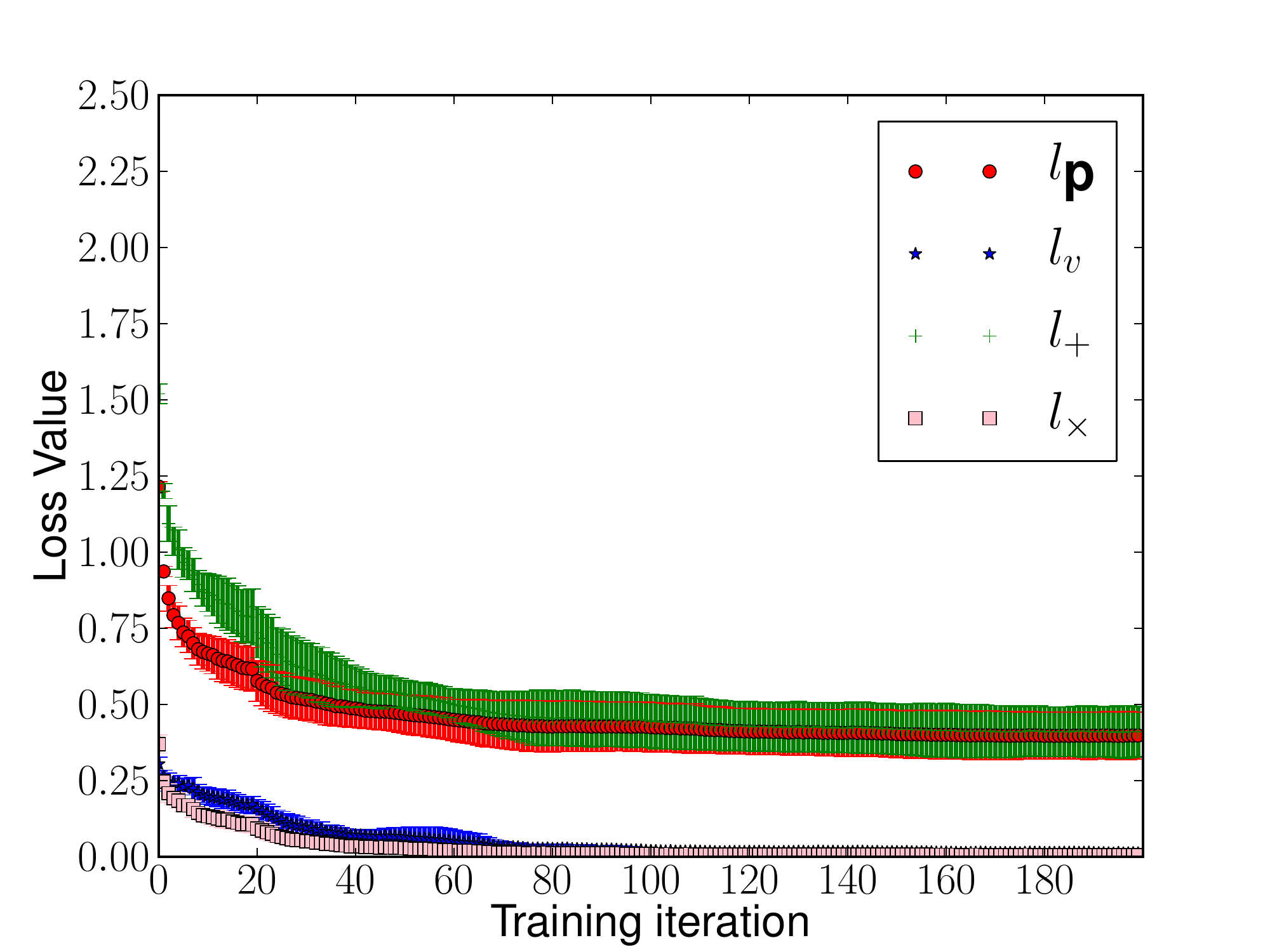}}
\hspace*{-1.8em}
\subfigure[Minimize $l_\times$]{\label{fig:subfigloss55othello:d}
\includegraphics[width=0.45\textwidth]{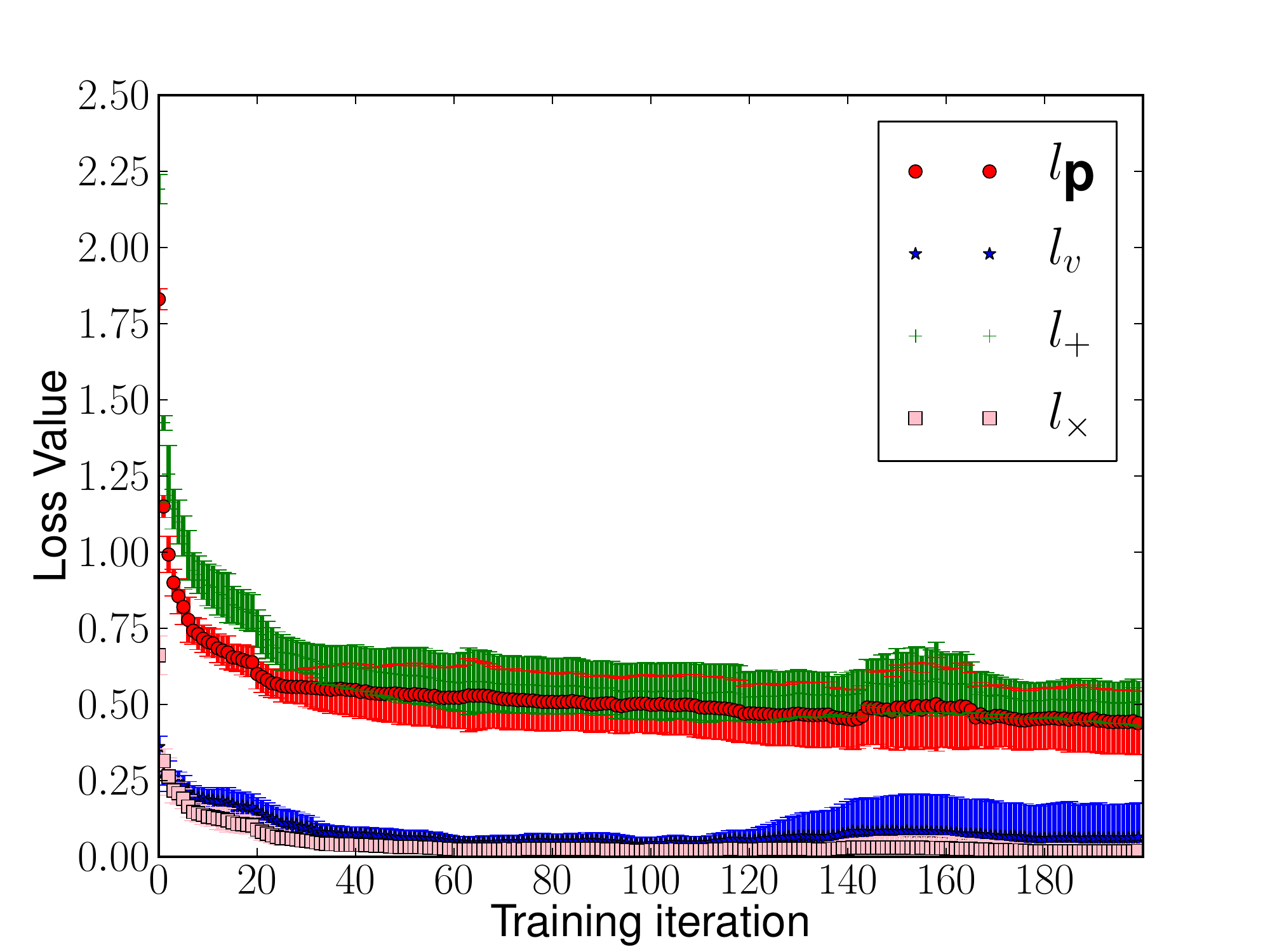}}
\caption{Training losses for minimizing different targets in 5$\times$5 Othello, averaged over 8 runs. All measured losses are shown, but only one of these is minimized for. Note the different scaling for subfigure (b). Except for the $l_+$, the target that is minimized for is also the lowest}
\label{fig:subfigloss55othello} %% label for entire figure
\end{figure}

% \begin{figure}[!tbh]
% %\vspace{-3em}
% \centering
% \hspace*{-1.5em}
% \subfigure[Minimize $l_\textbf{p}$]{\label{fig:subfigloss66othello:a}
% \includegraphics[width=0.45\textwidth]{errorbar_onlypi_66othello_loss_8runs}}
% \hspace*{-1.8em}
% \subfigure[Minimize $l_v$]{\label{fig:subfigloss66othello:b}
% \includegraphics[width=0.45\textwidth]{errorbar_onlyv_66othello_loss_8runs}}
% \hspace*{-1.5em}
% \subfigure[Minimize $l_+$]{\label{fig:subfigloss66othello:c}
% \includegraphics[width=0.45\textwidth]{errorbar_pi_v_66othello_loss_8runs}}
% \hspace*{-1.8em}
% \subfigure[Minimize $l_\times$]{\label{fig:subfigloss66othello:d}
% \includegraphics[width=0.45\textwidth]{errorbar_piv_66othello_loss_8runs}}
% \caption{Training losses for minimizing different targets in 6$\times$6 Othello, averaged from 8 runs. All losses are shown while we minimize only one (similar to Fig~\ref{fig:subfigloss55othello}). Note the different scaling for subfigure (b).
% Except for $l_+$, the target that is minimized for is the lowest}
% \label{fig:subfigloss66othello} %% label for entire figure
% \end{figure}

\begin{figure}[!tbh]
%\vspace{-3em}
\centering
\hspace*{-1.5em}
\subfigure[Minimize $l_\textbf{p}$]{\label{fig:subfigloss55connect4:a}
\includegraphics[width=0.45\textwidth]{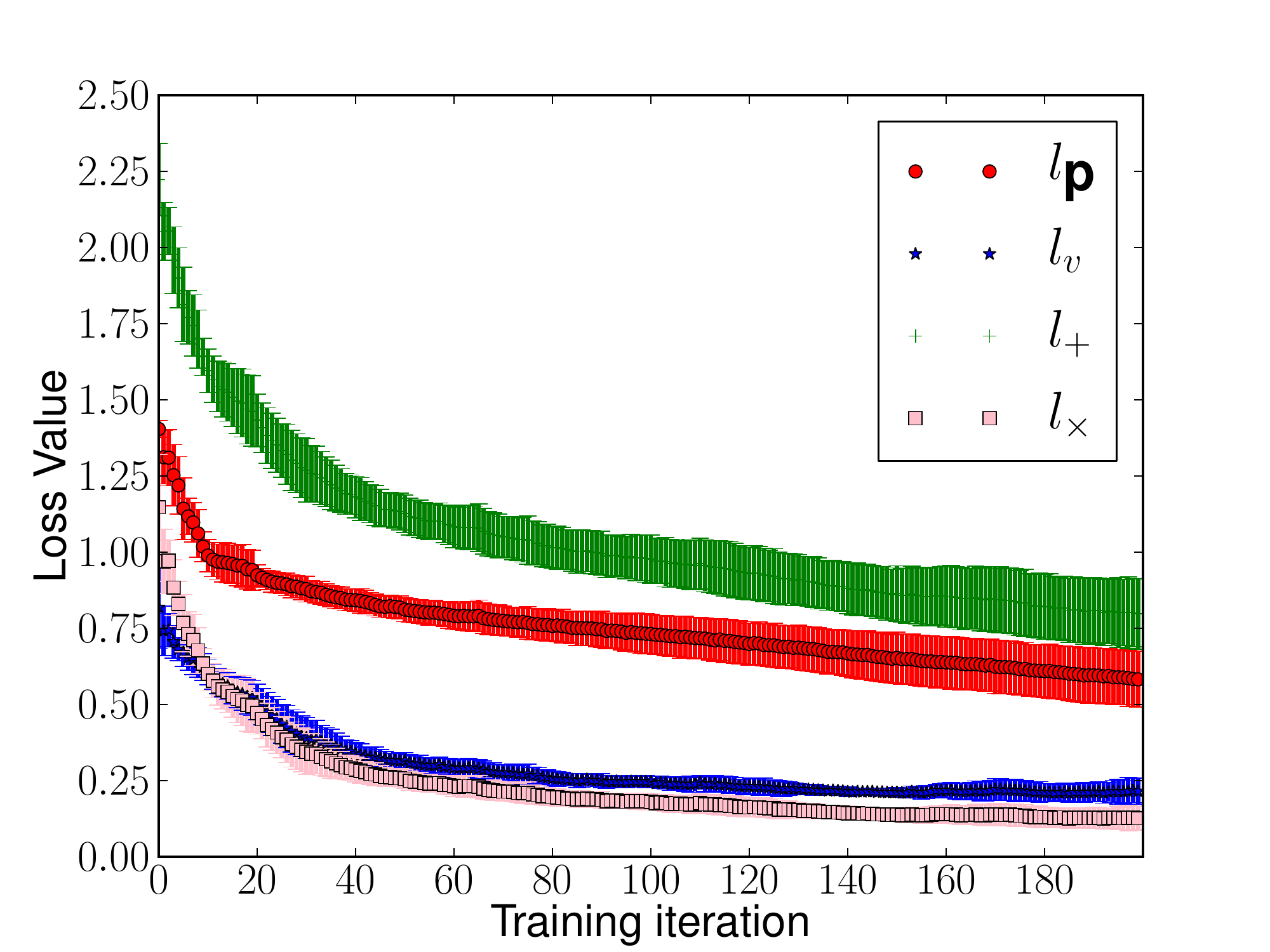}}
\hspace*{-1.8em}
\subfigure[Minimize $l_v$]{\label{fig:subfigloss55connect4:b}
\includegraphics[width=0.45\textwidth]{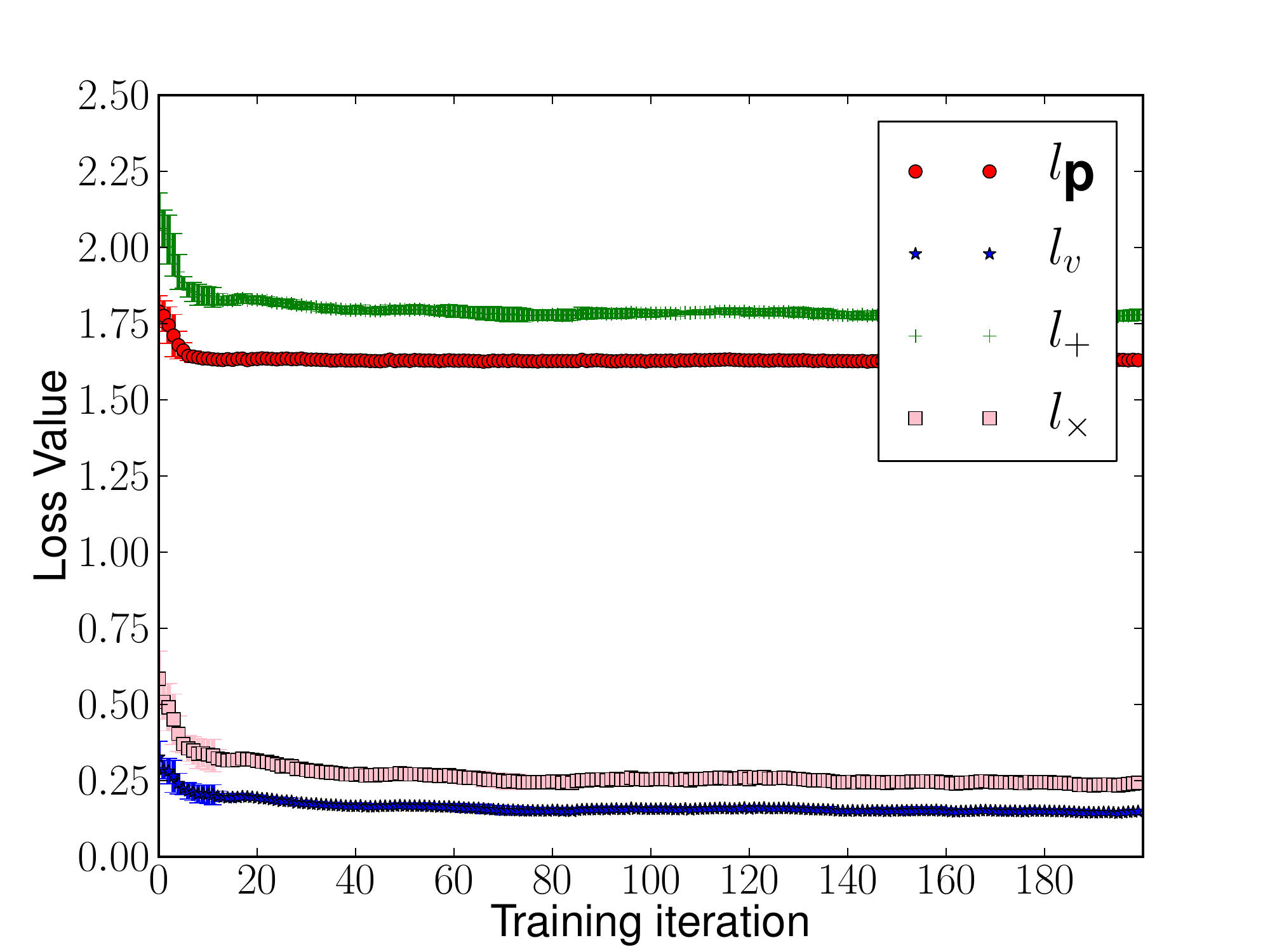}}
\hspace*{-1.5em}
\subfigure[Minimize $l_+$]{\label{fig:subfigloss55connect4:c}
\includegraphics[width=0.45\textwidth]{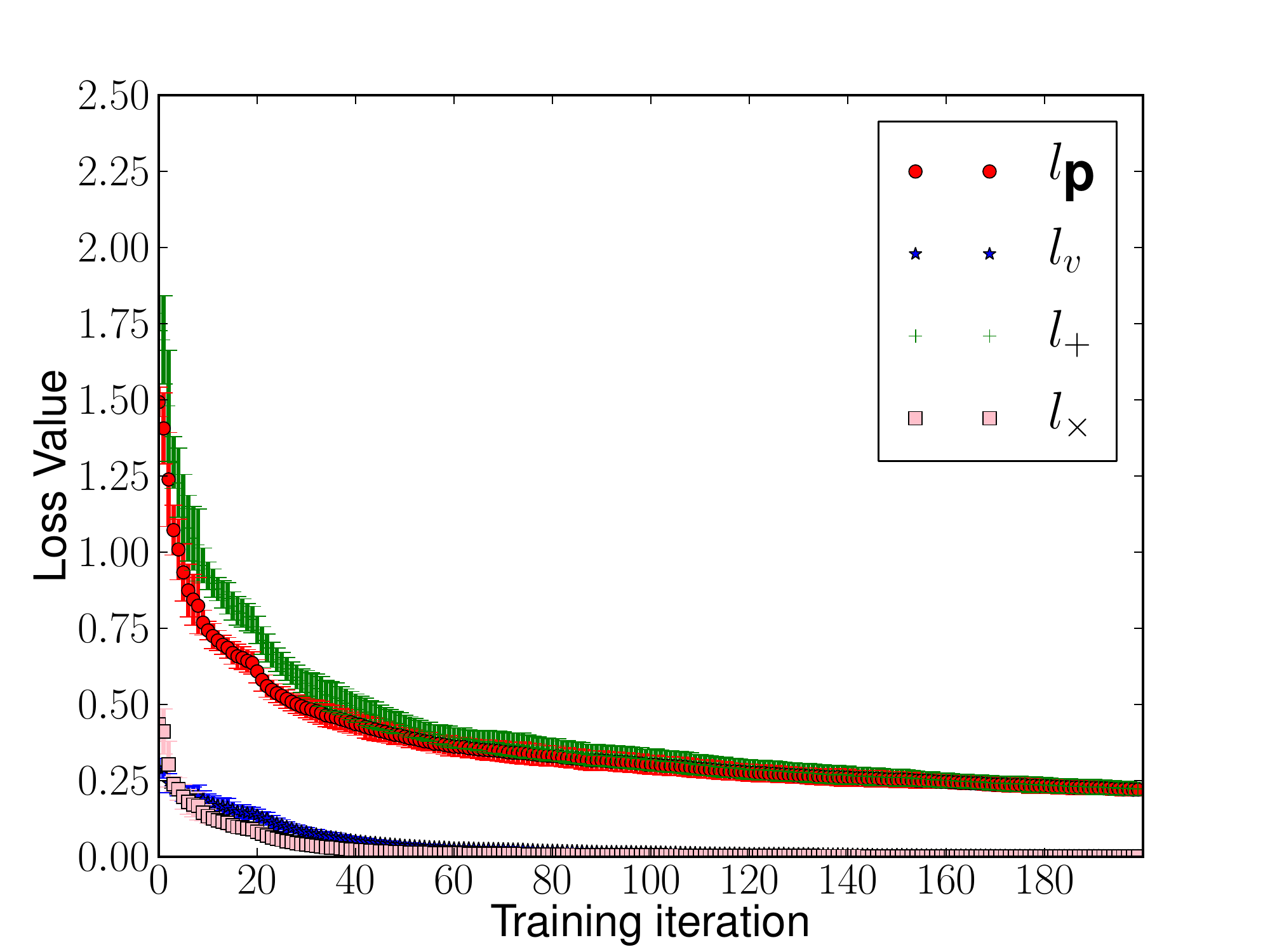}}
\hspace*{-1.8em}
\subfigure[Minimize $l_\times$]{\label{fig:subfigloss55connect4:d}
\includegraphics[width=0.45\textwidth]{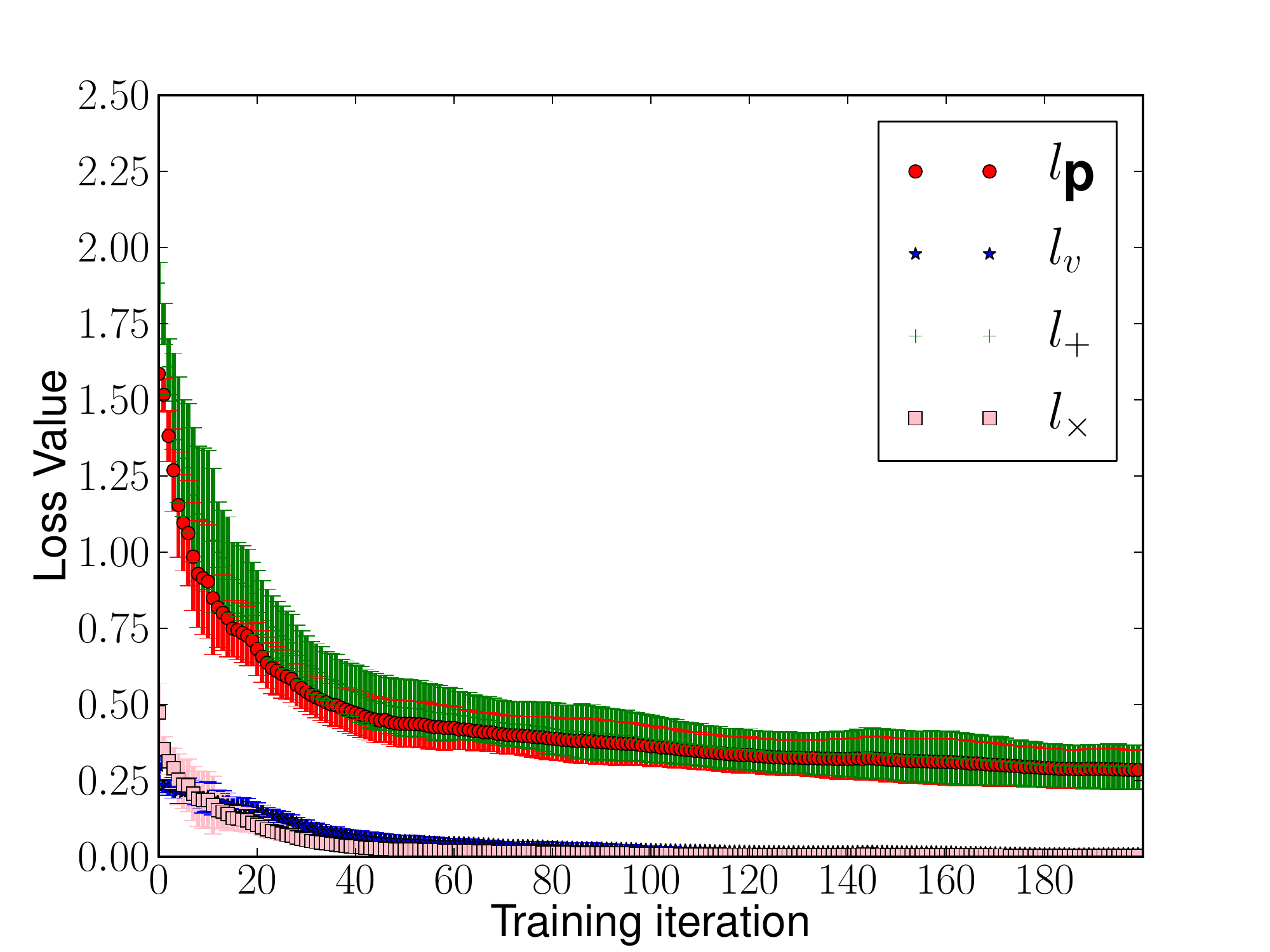}}
\caption{Training losses for minimizing the four different targets in 5$\times$5 Connect Four, averaged from 8 runs. $l_v$ is always the lowest}
\label{fig:subfigloss55connect4} %% label for entire figure
\end{figure}

\begin{figure}[!tbh]
%\vspace{-2em}
\centering
\hspace*{-1.5em}
\subfigure[Minimize $l_\textbf{p}$]{\label{fig:subfigloss55Gobang:a}
\includegraphics[width=0.45\textwidth]{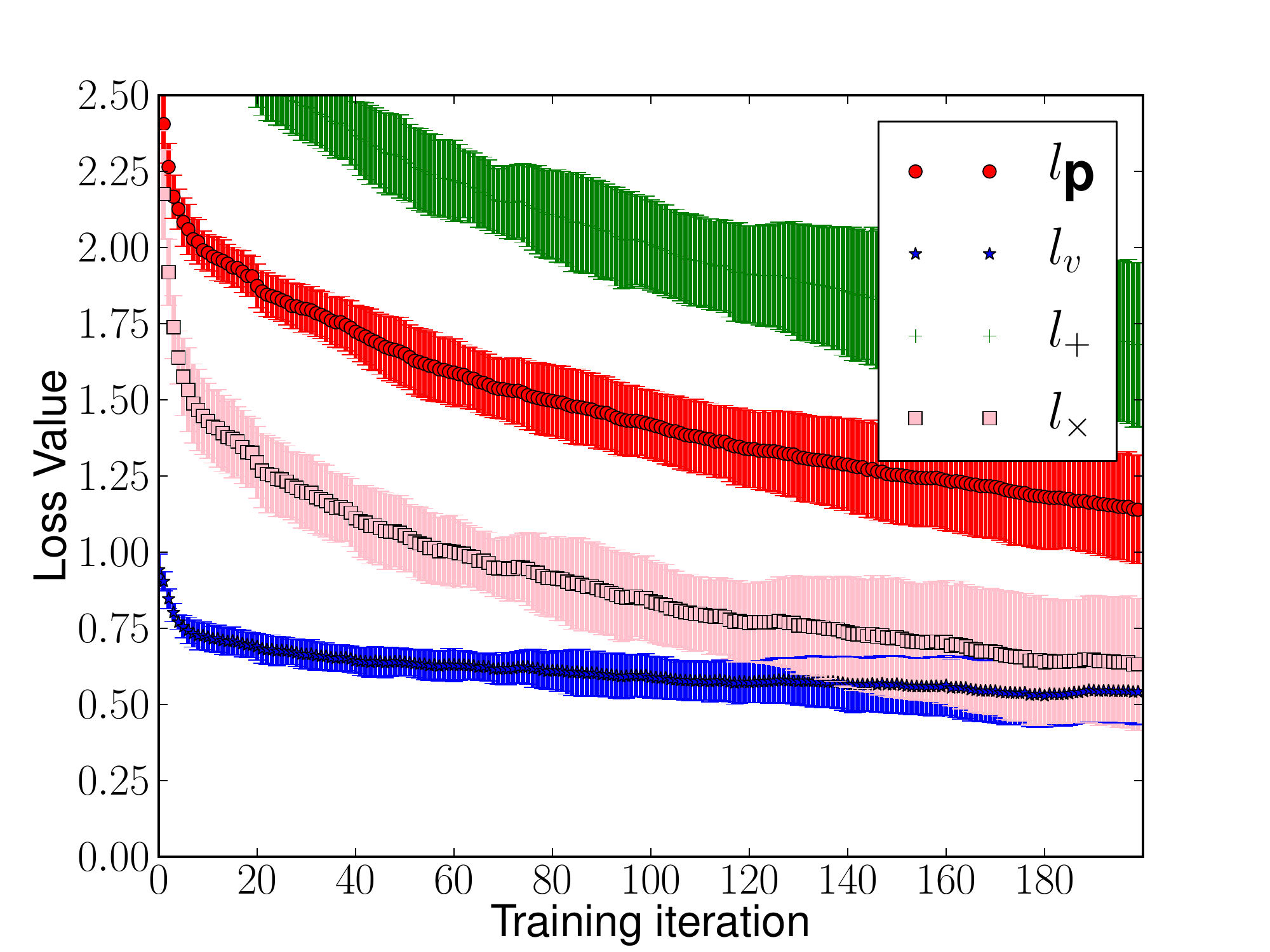}}
\hspace*{-1.8em}
\subfigure[Minimize $l_v$]{\label{fig:subfigloss55Gobang:b}
\includegraphics[width=0.45\textwidth]{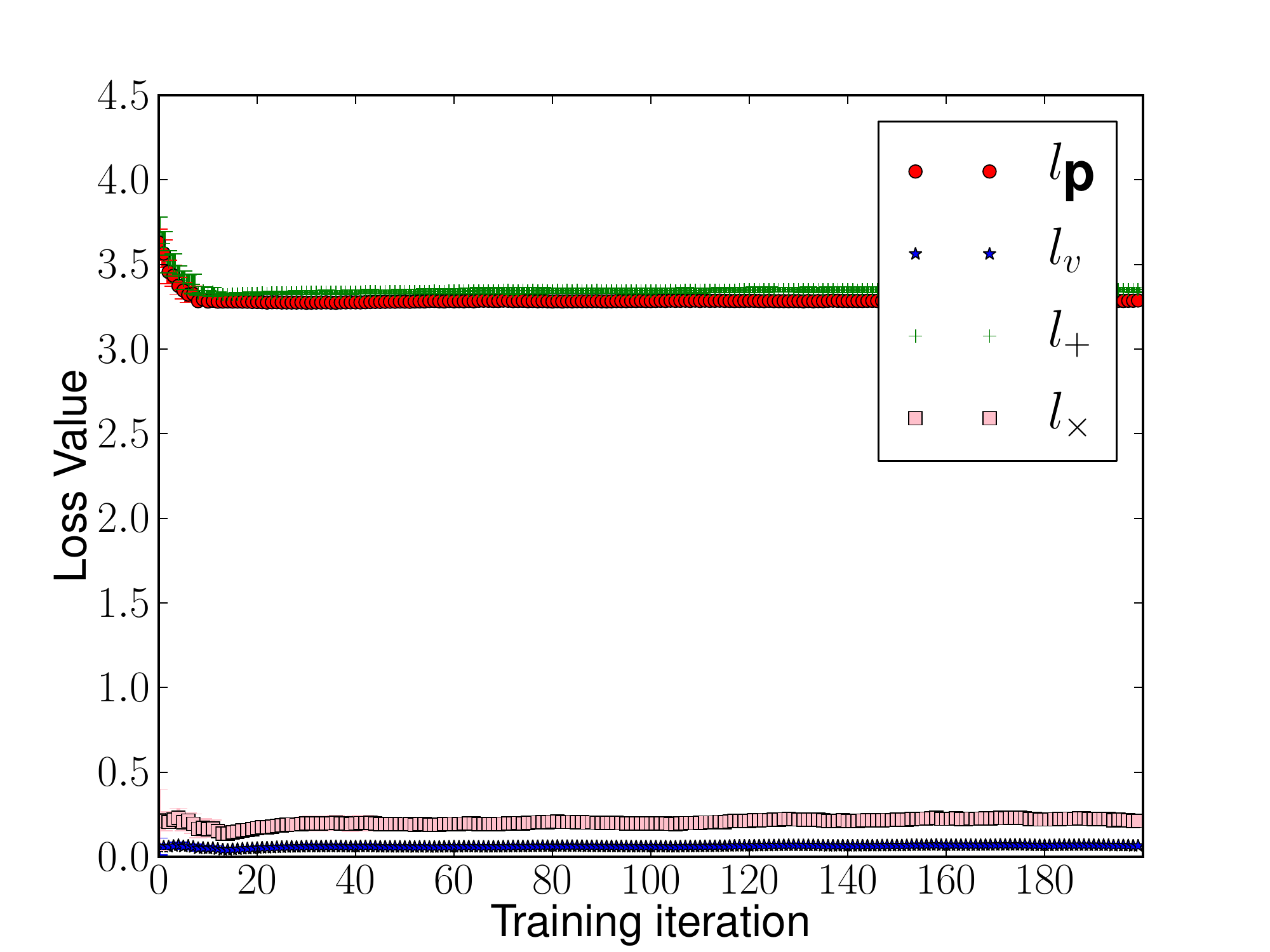}}
\hspace*{-1.5em}
\subfigure[Minimize $l_+$]{\label{fig:subfigloss55Gobang:c}
\includegraphics[width=0.45\textwidth]{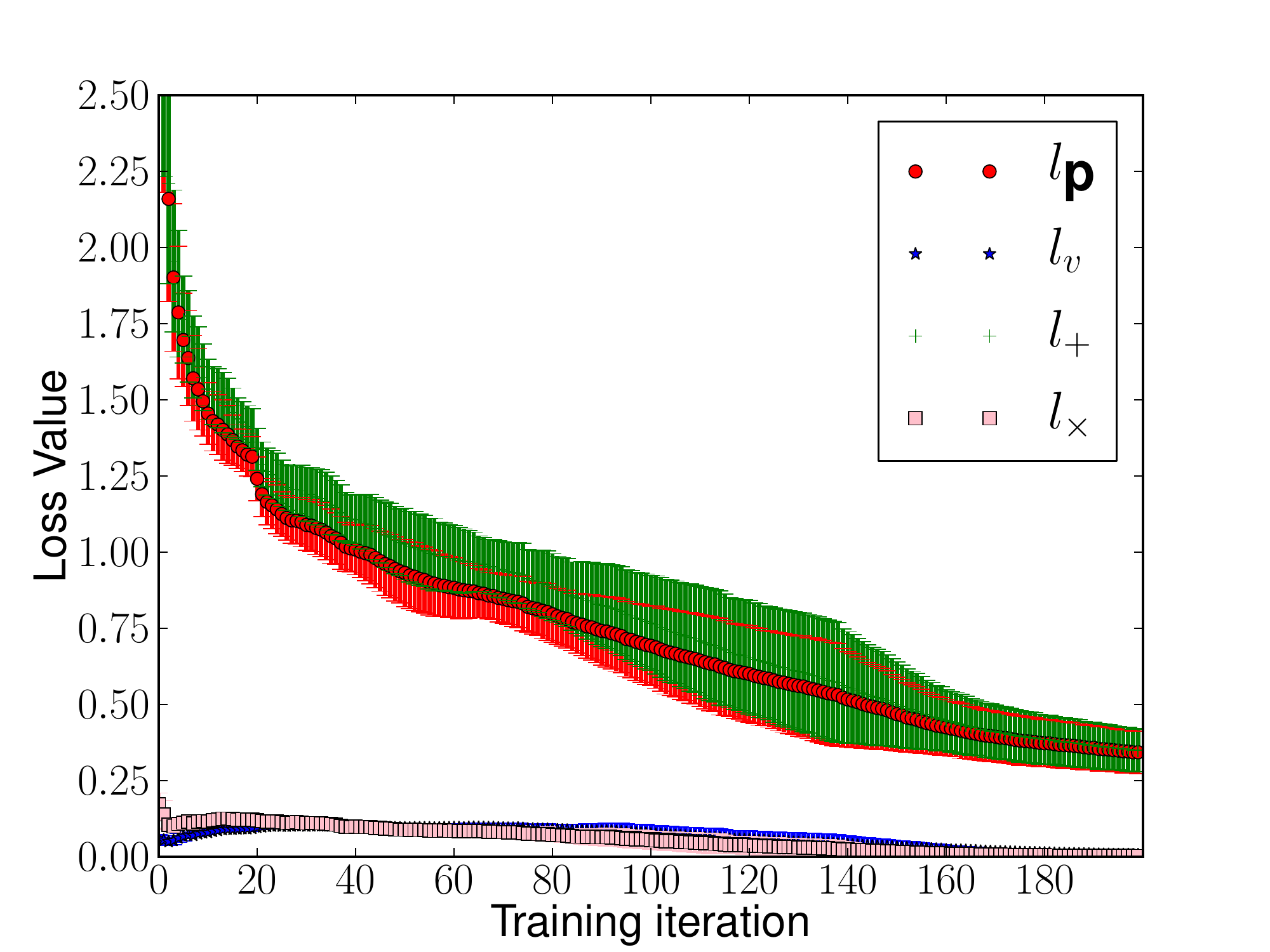}}
\hspace*{-1.8em}
\subfigure[Minimize $l_\times$]{\label{fig:subfigloss55Gobang:d}
\includegraphics[width=0.45\textwidth]{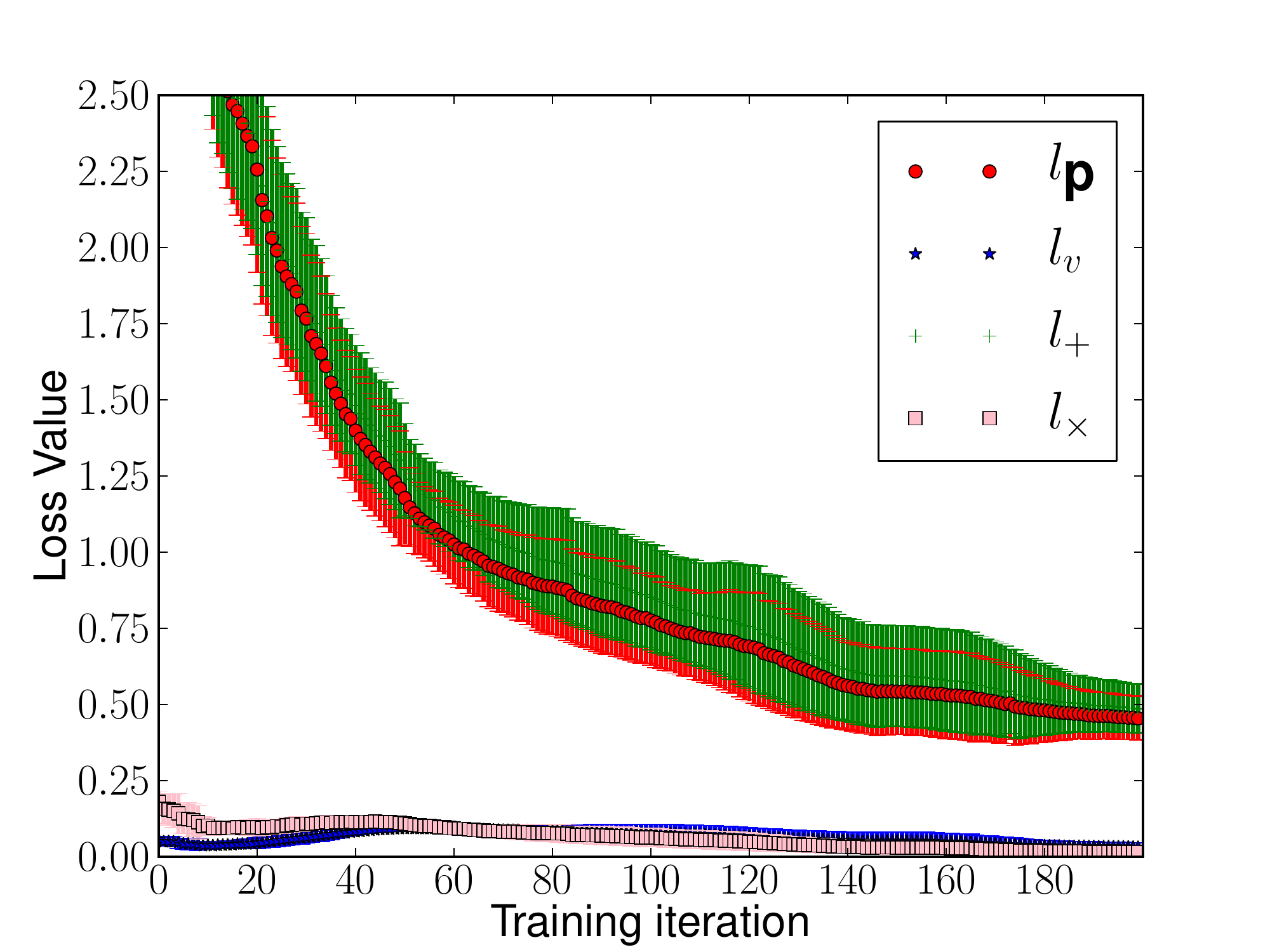}}
\caption{Training losses for minimizing the four different targets in 5$\times$5 Gobang, averaged from 8 runs. $l_v$ is always the lowest}
\label{fig:subfigloss55Gobang} %% label for entire figure
\end{figure}

% \begin{figure}[!tbh]
% %\vspace{-2em}
% \centering
% \hspace*{-1.5em}
% \subfigure[Minimize $l_\textbf{p}$]{\label{fig:subfigloss66Gobang:a}
% \includegraphics[width=0.45\textwidth]{errorbar_onlypi_66gobang_loss_8runs.pdf}}
% \hspace*{-1.8em}
% \subfigure[Minimize $l_v$]{\label{fig:subfigloss66Gobang:b}
% \includegraphics[width=0.45\textwidth]{errorbar_onlyv_66gobang_loss_8runs.pdf}}
% \hspace*{-1.5em}
% \subfigure[Minimize $l_+$]{\label{fig:subfigloss66Gobang:c}
% \includegraphics[width=0.45\textwidth]{errorbar_pi_v_66gobang_loss_8runs.pdf}}
% \hspace*{-1.8em}
% \subfigure[Minimize $l_\times$]{\label{fig:subfigloss66Gobang:d}
% \includegraphics[width=0.45\textwidth]{errorbar_piv_66gobang_loss_8runs.pdf}}
% \caption{Training losses for minimizing the four different targets in 6$\times$6 Gobang, aggregated from 8 runs. $l_v$ is always the lowest}
% \label{fig:subfigloss55Gobang} %% label for entire figure
% \end{figure}

We first show the training losses in every iteration with one minimization task per diagram, hence we need four of these per game. In these graphs we see what minimizing for a specific target actually means for the other loss types.

For 5$\times$5 Othello, from Fig.~\ref{fig:subfigloss55othello:a}, we find that when minimizing $l_\textbf{p}$ only, the loss decreases significantly to about 0.6 at the end of each training, whereas $l_v$ stagnates at 1.0 after 10 iterations. Minimizing only $l_v$ (Fig.~\ref{fig:subfigloss55othello:b}) brings it down from $0.5$ to $0.2$, but $l_\textbf{p}$ remains stable at a high level. In  Fig.~\ref{fig:subfigloss55othello:c}, we see that when the $l_+$ is minimized, both losses are reduced significantly. The $l_\textbf{p}$ decreases from about 1.2 to 0.5,  $l_v$ surprisingly decreases to 0.  Fig.~\ref{fig:subfigloss55othello:d}, it is similar to Fig.~\ref{fig:subfigloss55othello:c}, while the $l_\times$ is minimized, the $l_\textbf{p}$ and $l_v$ are also reduced. The $l_\textbf{p}$ decreases to 0.5, the $l_v$ also surprisingly decreases to about 0. Figures for 6$\times$6 Othello are not shown since they are very similar to 5$\times$5 ~(for the same reason we do not show loss pictures for 6$\times$6 Connect Four and 6$\times$6 Gobang).

% For the larger 6$\times$6 Othello, we find that minimizing only $l_\textbf{p}$ reduces it significantly to about 0.75, where $l_v$ is stable again after about 10 iterations (Fig.~\ref{fig:subfigloss66othello:a}). For minimizing $l_v$ (Fig.~\ref{fig:subfigloss66othello:b}), the results show that $l_v$ is reduced from more than 0.5 to about 0.25 at the end of each training, but $l_\textbf{p}$ seems to remain almost unchanged.
% For minimizing the $l_+$ (Fig.~\ref{fig:subfigloss66othello:c}), we find in contrast to 5$\times$5 Othello that $l_\textbf{p}$ decreases from about 1.1 to 0.4, whereas $l_v$ increases slightly from about 0.2 and then decreases to about 0.2 again. We also find a similar behavior of $l_v$ when minimizing the $l_\times$ (Fig.~\ref{fig:subfigloss66othello:d}), with the difference that the final computed loss is much lower as the values are usually smaller than one. However, the similarity of the single losses is striking.

For 5$\times$5 Connect Four (see Fig.~\ref{fig:subfigloss55connect4:a}), we find that when only minimizing $l_\textbf{p}$, it  significantly reduces from 1.4 to about 0.6, whereas $l_v$ is minimized much quicker from 1.0 to about 0.2, where it is almost stationary.
Minimizing $l_v$ (Fig.~\ref{fig:subfigloss55connect4:b}) leads to some reduction from more than 0.5 to about 0.15, but
$l_\textbf{p}$ is not moving much after an initial slight decrease to about 1.6.
For minimizing the $l_+$ (Fig.~\ref{fig:subfigloss55connect4:c}) and the $l_\times$ (Fig.~\ref{fig:subfigloss55connect4:d}), the behavior of $l_\textbf{p}$ and $l_v$ is very similar, they both
decrease steadily, until $l_v$ surprisingly reaches 0.
Of course the $l_+$ and the $l_\times$ arrive at different values, but in terms of both $l_\textbf{p}$ and $l_v$ they are not different.

For 5$\times$5 Gobang game, we find that, in Fig.~\ref{fig:subfigloss55Gobang}, when only minimizing $l_\textbf{p}$, $l_\textbf{p}$ value decreases from around 2.5 to about 1.25 while the $l_v$ value reduces from 1.0 to 0.5~(see Fig.~\ref{fig:subfigloss55Gobang:a}). When minimizing $l_v$, $l_v$ value quickly reduces to a very level which is lower than 0.1~(see Fig.~\ref{fig:subfigloss55Gobang:b}). Minimizing $l_+$ and $l_\times$ both lead to stationary low $l_v$ values from the beginning of training which is different from Othello and Connect Four. 

\subsubsection{Training Elo Rating}\label{lossfunctionwholehistoryelo}

\begin{figure}[!tbh]
\centering
\hspace*{-1.5em}
\subfigure[5$\times$5 Othello]{\label{fig:subfigElowholehistory:a}
\includegraphics[width=0.45\textwidth]{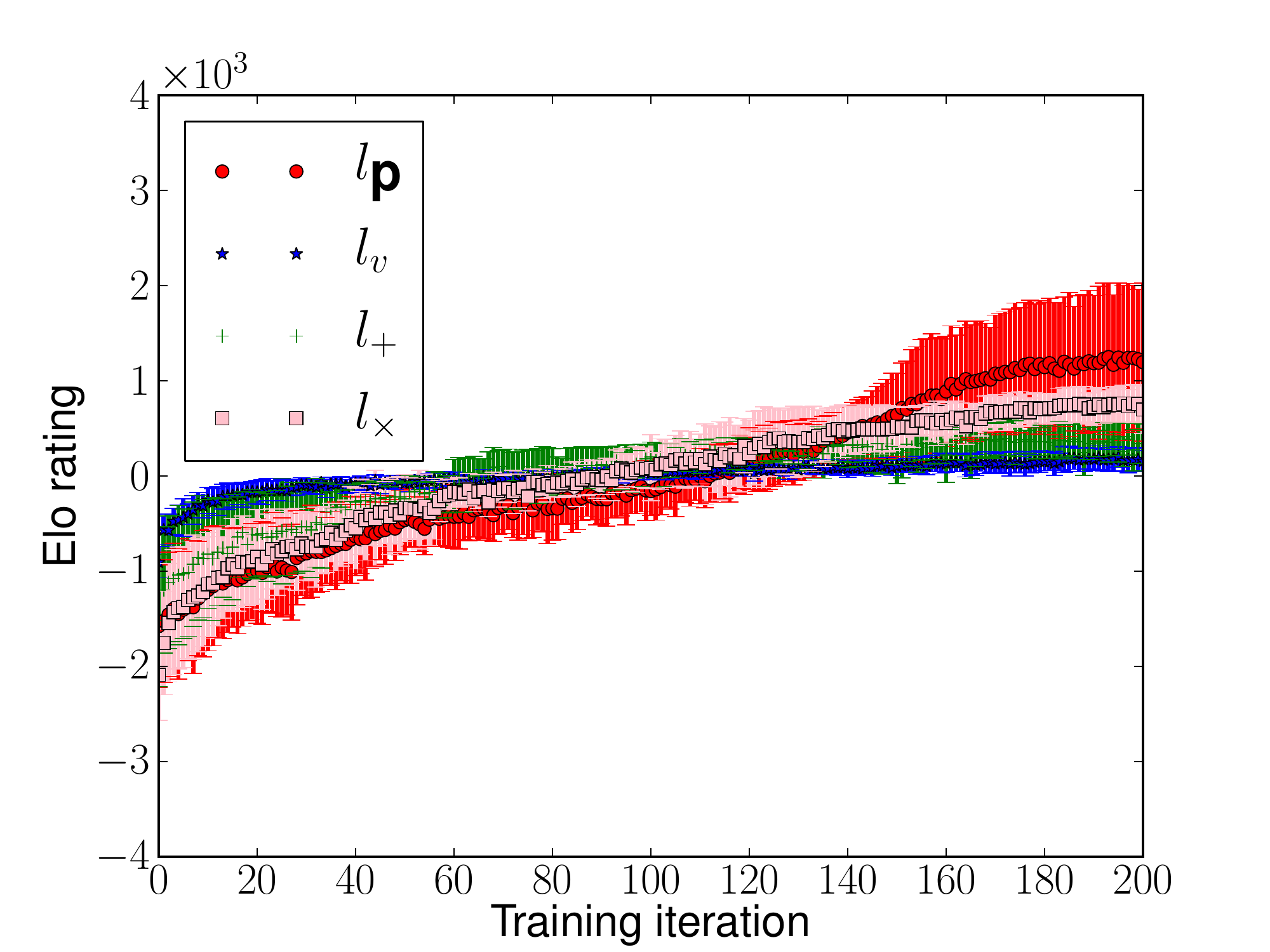}}
\hspace*{-1.8em}
\subfigure[6$\times$6 Othello]{\label{fig:subfigElowholehistory:b}
\includegraphics[width=0.45\textwidth]{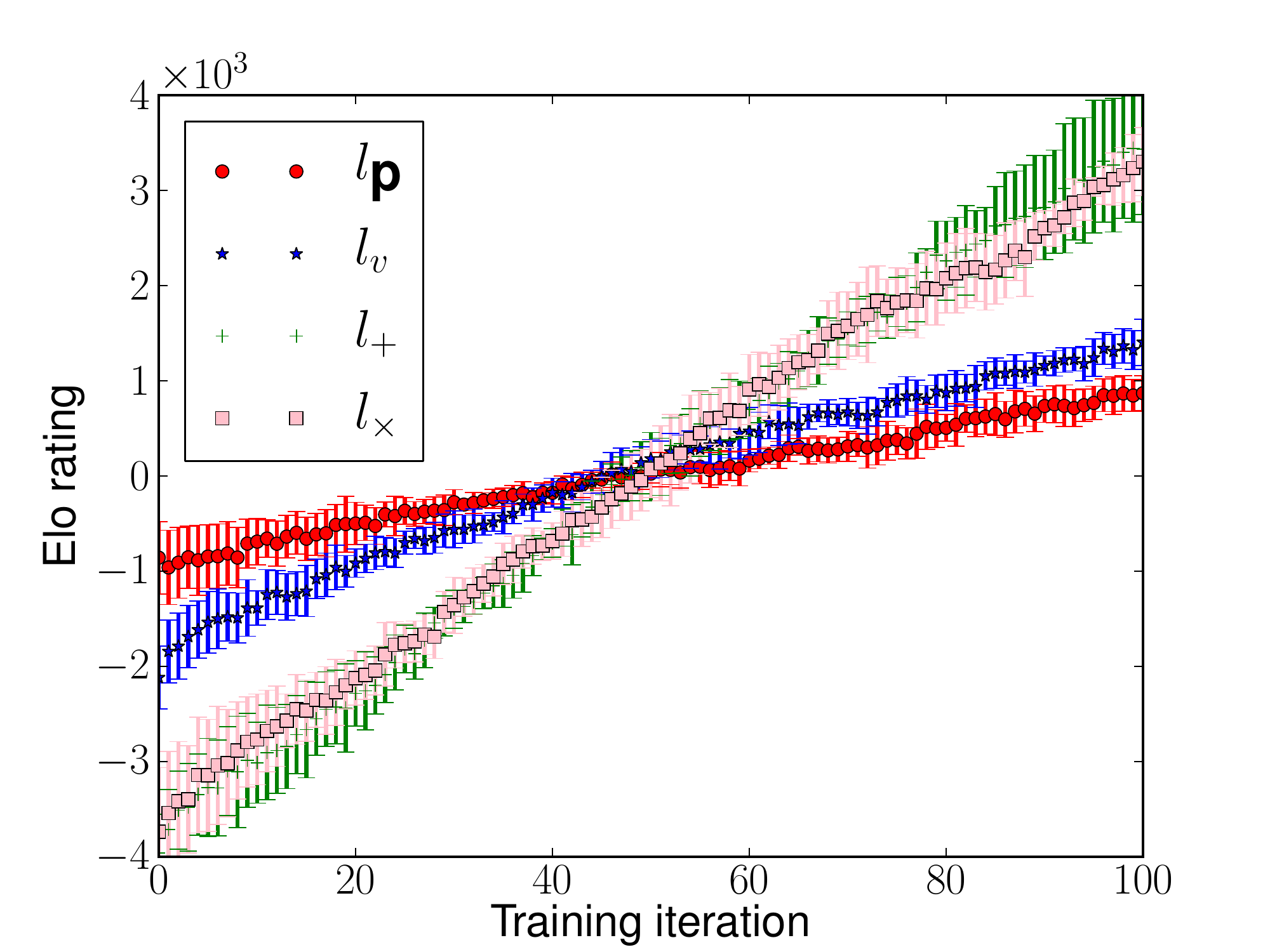}}
\hspace*{-1.5em}
\subfigure[5$\times$5 Connect Four]{\label{fig:subfigElowholehistory:c}
\includegraphics[width=0.45\textwidth]{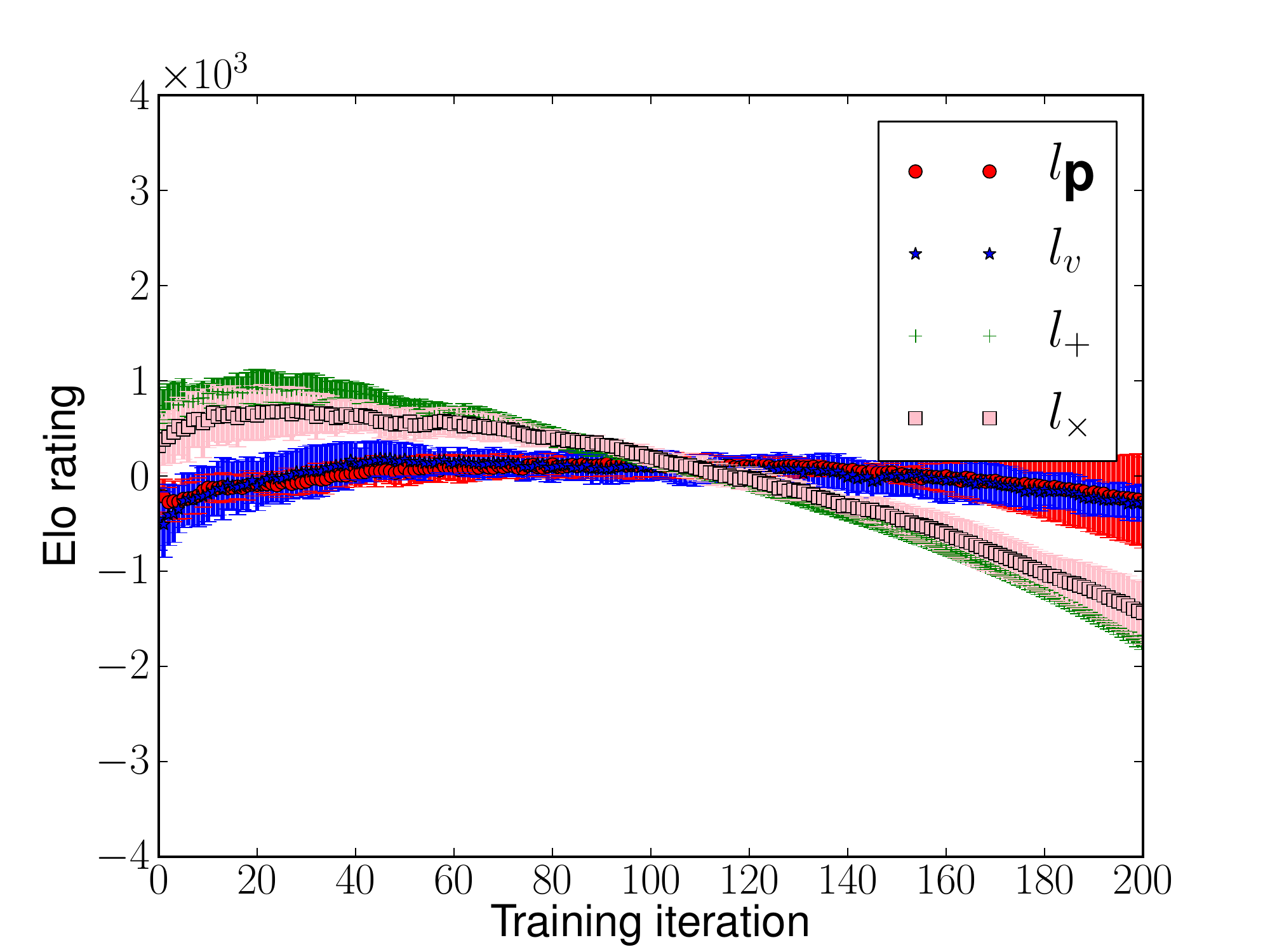}}
\hspace*{-1.8em}
\subfigure[6$\times$6 Connect Four]{\label{fig:subfigElowholehistory:d}
\includegraphics[width=0.45\textwidth]{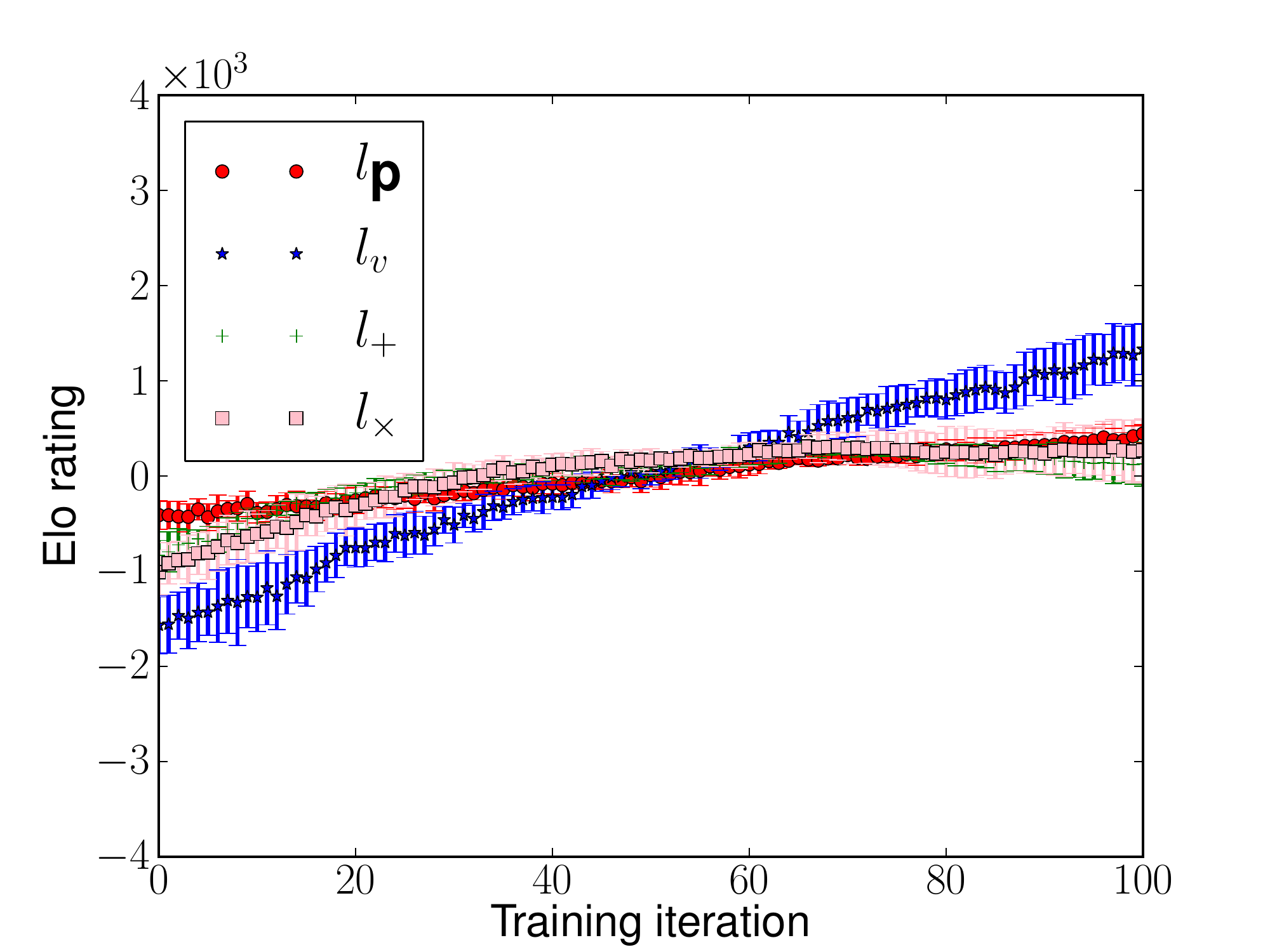}}
\hspace*{-1.5em}
\subfigure[5$\times$5 Gobang]{\label{fig:subfigElowholehistory:e}
\includegraphics[width=0.45\textwidth]{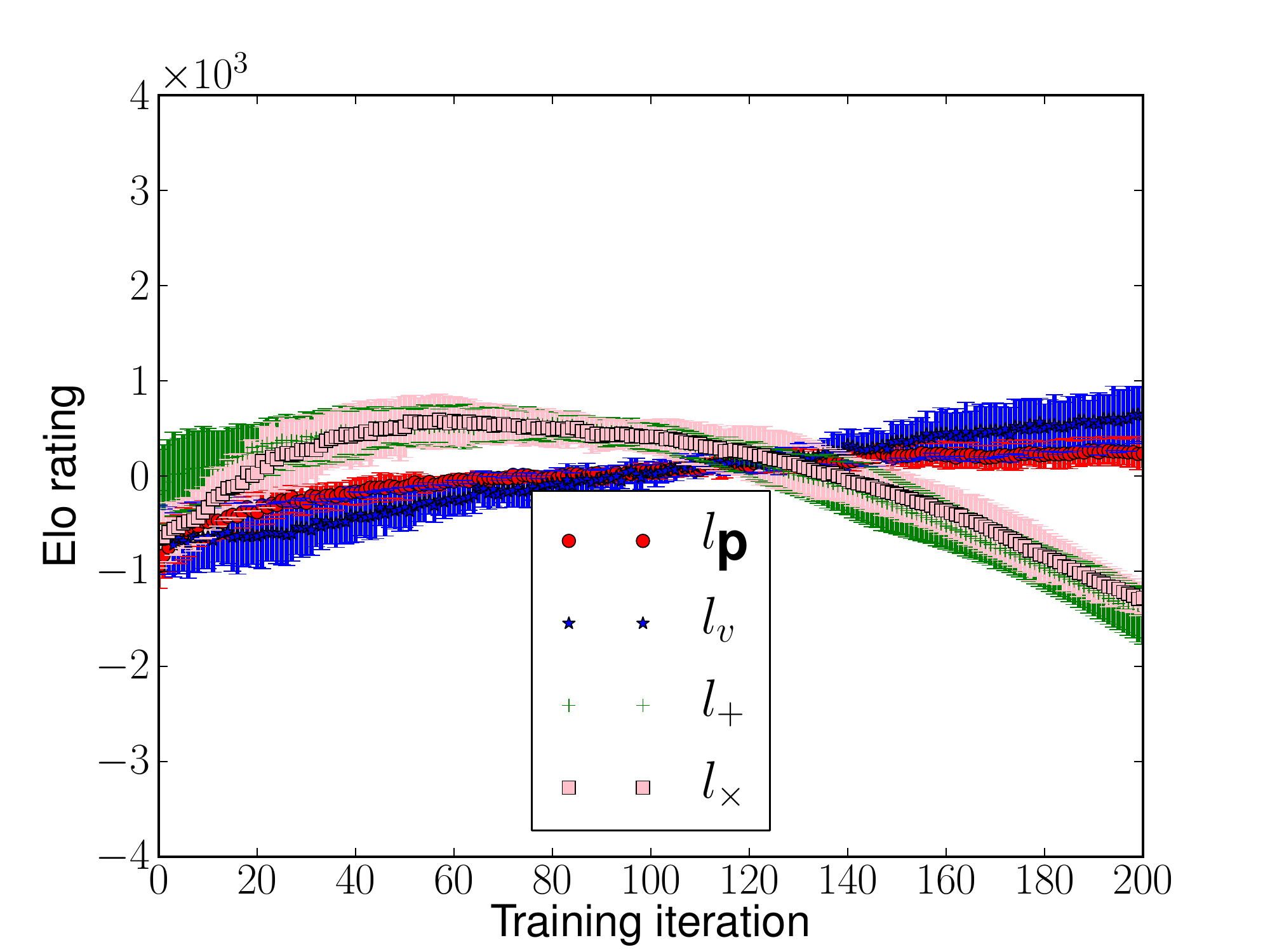}}
\hspace*{-1.8em}
\subfigure[6$\times$6 Gobang]{\label{fig:subfigElowholehistory:f}
\includegraphics[width=0.45\textwidth]{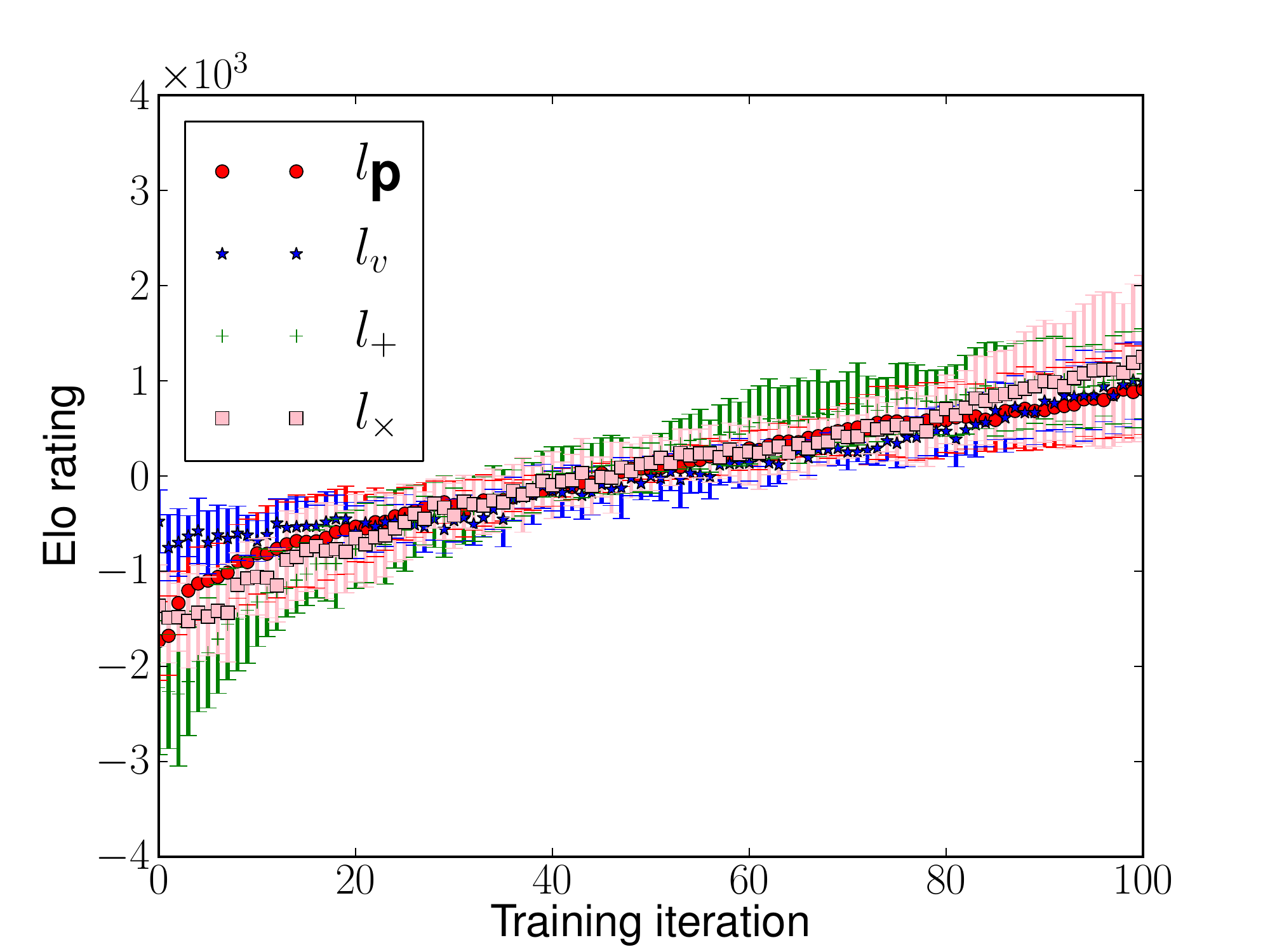}}
\caption{The whole history Elo rating at each iteration during training for different games, aggregated from 8 runs. The training Elo for $l_+$ and $l_\times$ in panel b and c for example shows inconsistent results}
\label{fig:subfigElowholehistory} %% label for entire figure
\end{figure}

Following the AlphaGo  papers, we also investigate the   training Elo rating of every iteration during training. Instead of showing results form single runs, we provide means and variances for 8 runs for each target, categorized by different games in Fig.~\ref{fig:subfigElowholehistory}.

From Fig.~\ref{fig:subfigElowholehistory:a} (small 5$\times$5 Othello) we see that for all minimization tasks, Elo values steadily improve, while they raise fastest for $l_\textbf{p}$. In Fig.~\ref{fig:subfigElowholehistory:b}, we find that for 6$\times$6 Othello version, Elo values also always improve, but much faster for the $l_+$ and $l_\times$ target, compared to the single loss targets.

Fig.~\ref{fig:subfigElowholehistory:c} and
Fig.~\ref{fig:subfigElowholehistory:d} show the Elo rate progression for training players with the four different targets on the small and larger Connect Four setting. This looks a bit different from the Othello results, as we find stagnation (for 6$\times$6 Connect Four) as well as even degeneration (for 5$\times$5 Connect Four). The latter actually means that for decreasing loss in the training phase, we achieve decreasing Elo rates, such that the players get weaker and not stronger. In the larger Connect Four setting, we still have a clear improvement, especially if we minimize for $l_v$. Minimizing for $l_\textbf{p}$ leads to stagnation quickly, or at least a very slow improvement.

Overall, we display the Elo progression obtained from the different minimization targets for one game together. However, one must be aware that their numbers are not directly comparable due to the high self-play bias~(as they stem from players who have never seen each other). Nevertheless, the trends are important, and it is especially interesting to see if Elo values correlate with the progression of losses. Based on the experimental results, we can conclude that the training Elo rating is certainly good for assessing if training actually works, whereas the losses alone do not always show that. We may even experience contradicting outcomes as stagnating losses and rising Elo ratings (for the big Othello setting and $l_v$) or completely counterintuitive results as for the small Connect Four setting where Elo ratings and losses are partly anti-correlated. We  have experimental evidence for the fact that training losses and Elo ratings are by no means exchangeable as they can provide very different impressions of what is actually happening.

\subsubsection{The Final Best Player Tournament Elo Rating}\label{lossfunctionfulltounamentresults}
\begin{figure}[!tbh]
\centering
\hspace*{-1.5em}
\subfigure[5$\times$5 Othello]{\label{fig:subfigEloallgamesplayers:a}
\includegraphics[width=0.45\textwidth]{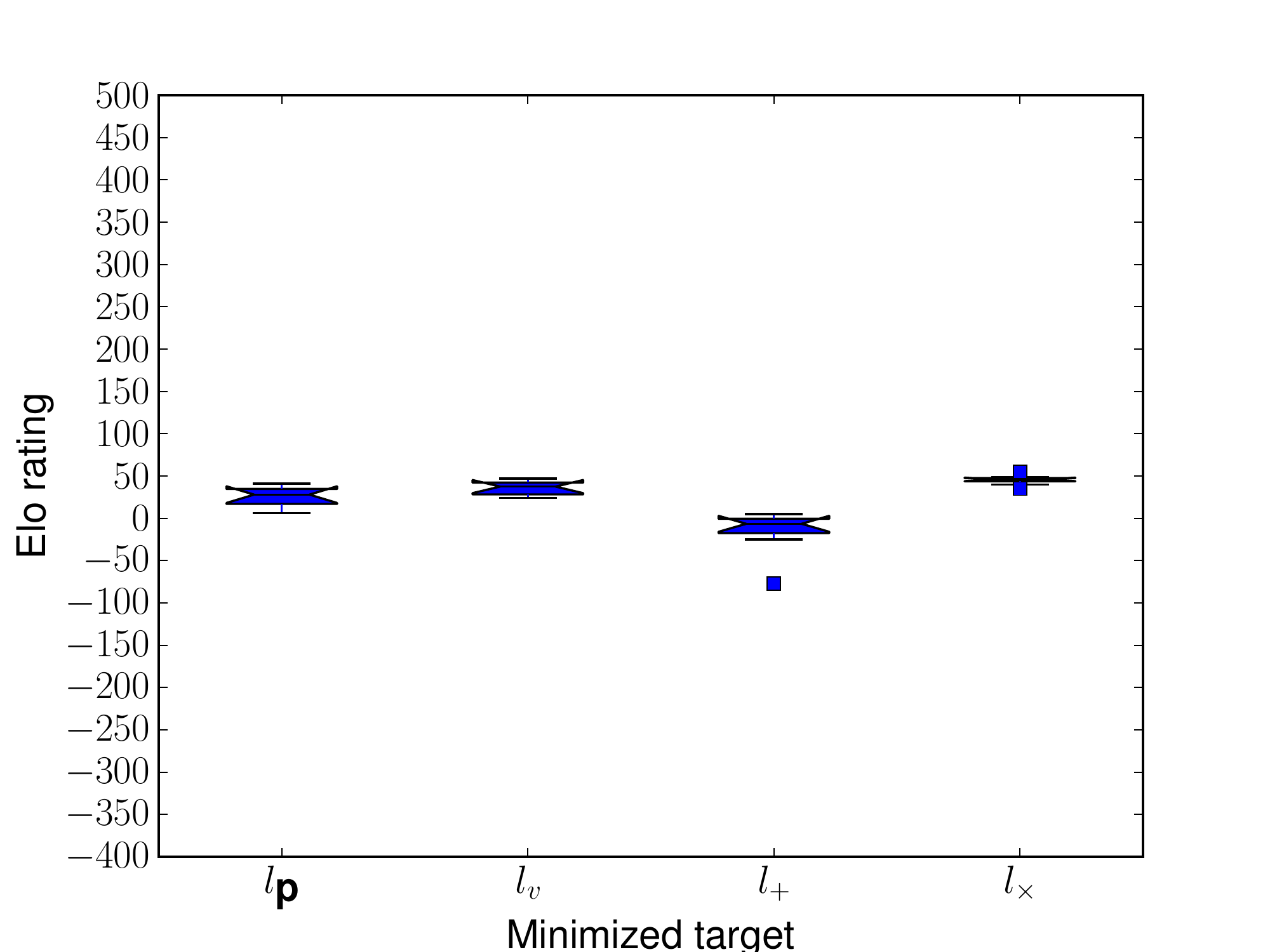}}
\hspace*{-1.8em}
\subfigure[6$\times$6 Othello]{\label{fig:subfigEloallgamesplayers:b}
\includegraphics[width=0.45\textwidth]{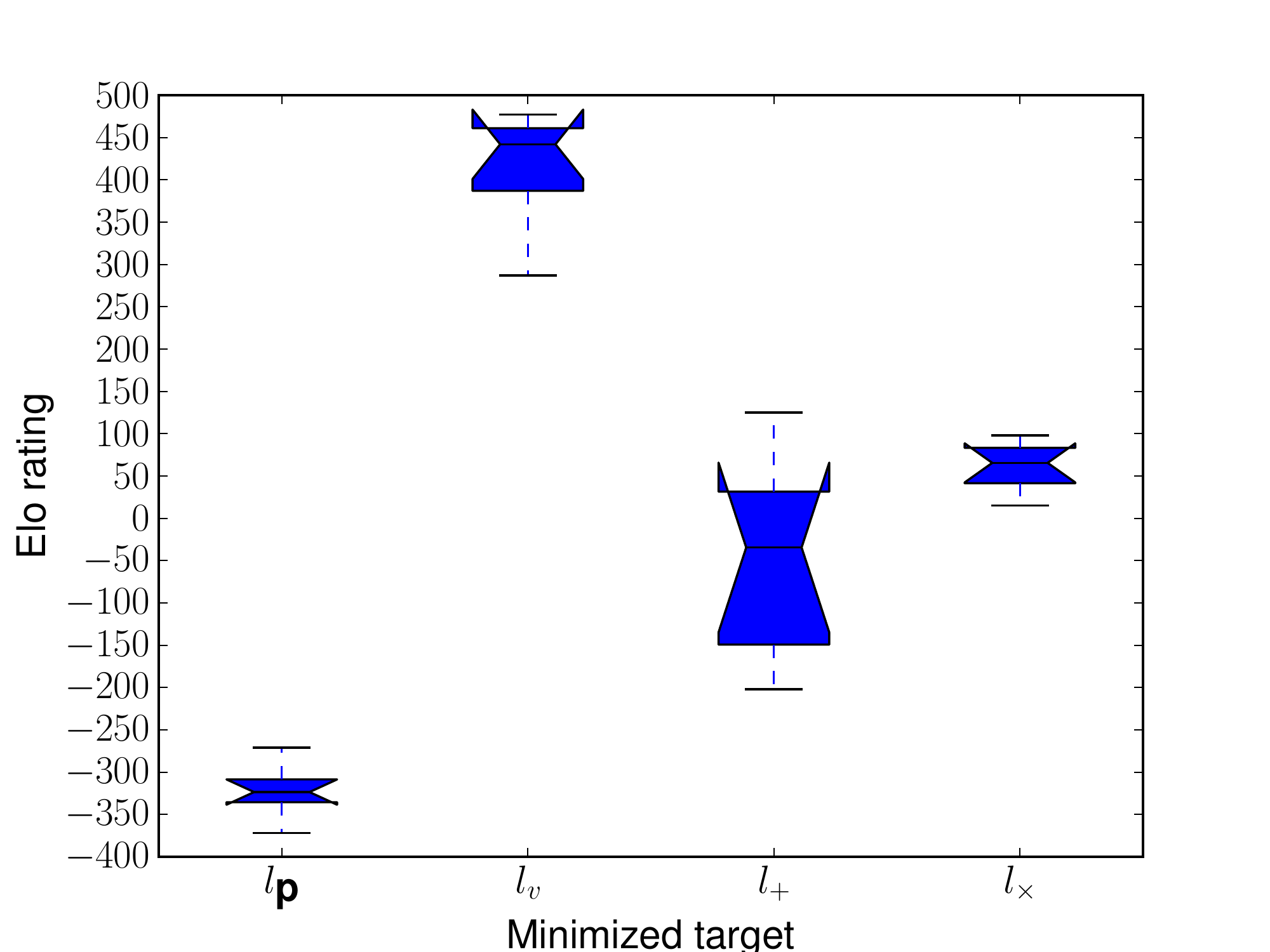}}
\hspace*{-1.5em}
\subfigure[5$\times$5 Connect Four]{\label{fig:subfigEloallgamesplayers:c}
\includegraphics[width=0.45\textwidth]{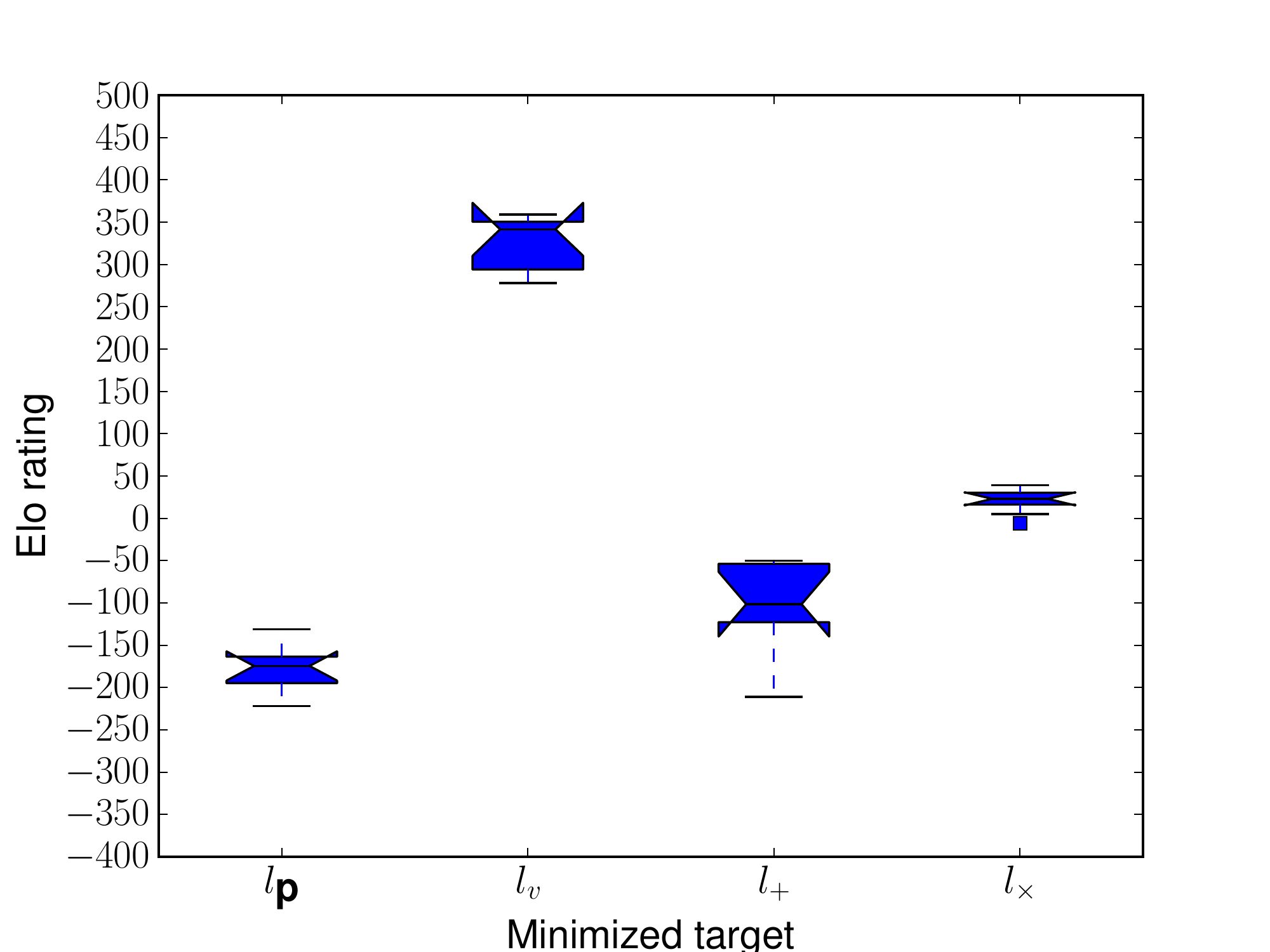}}
\hspace*{-1.8em}
\subfigure[6$\times$6 Connect Four]{\label{fig:subfigEloallgamesplayers:d}
\includegraphics[width=0.45\textwidth]{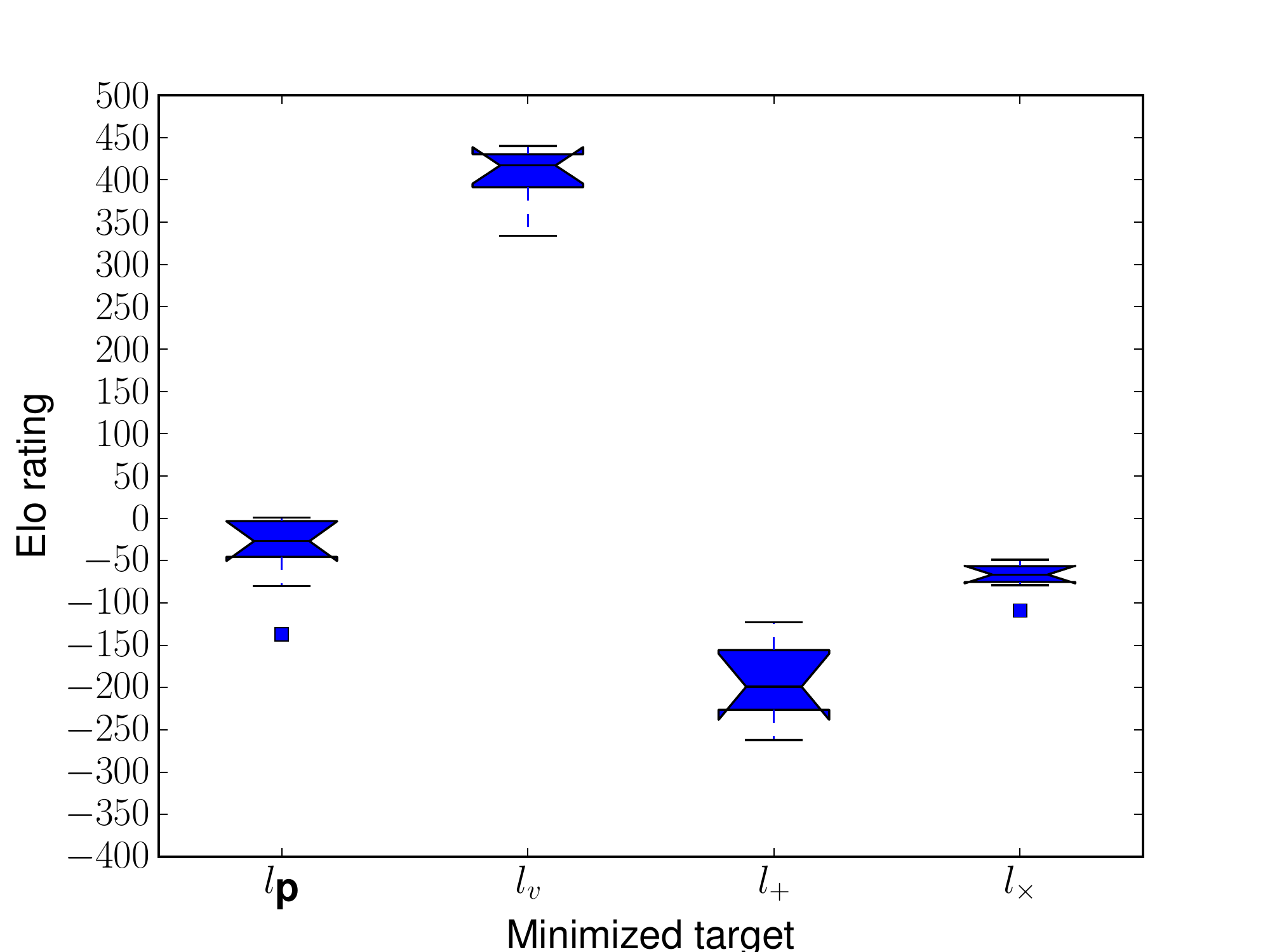}}
\hspace*{-1.5em}
\subfigure[5$\times$5 Gobang]{\label{fig:subfigEloallgamesplayers:e}
\includegraphics[width=0.45\textwidth]{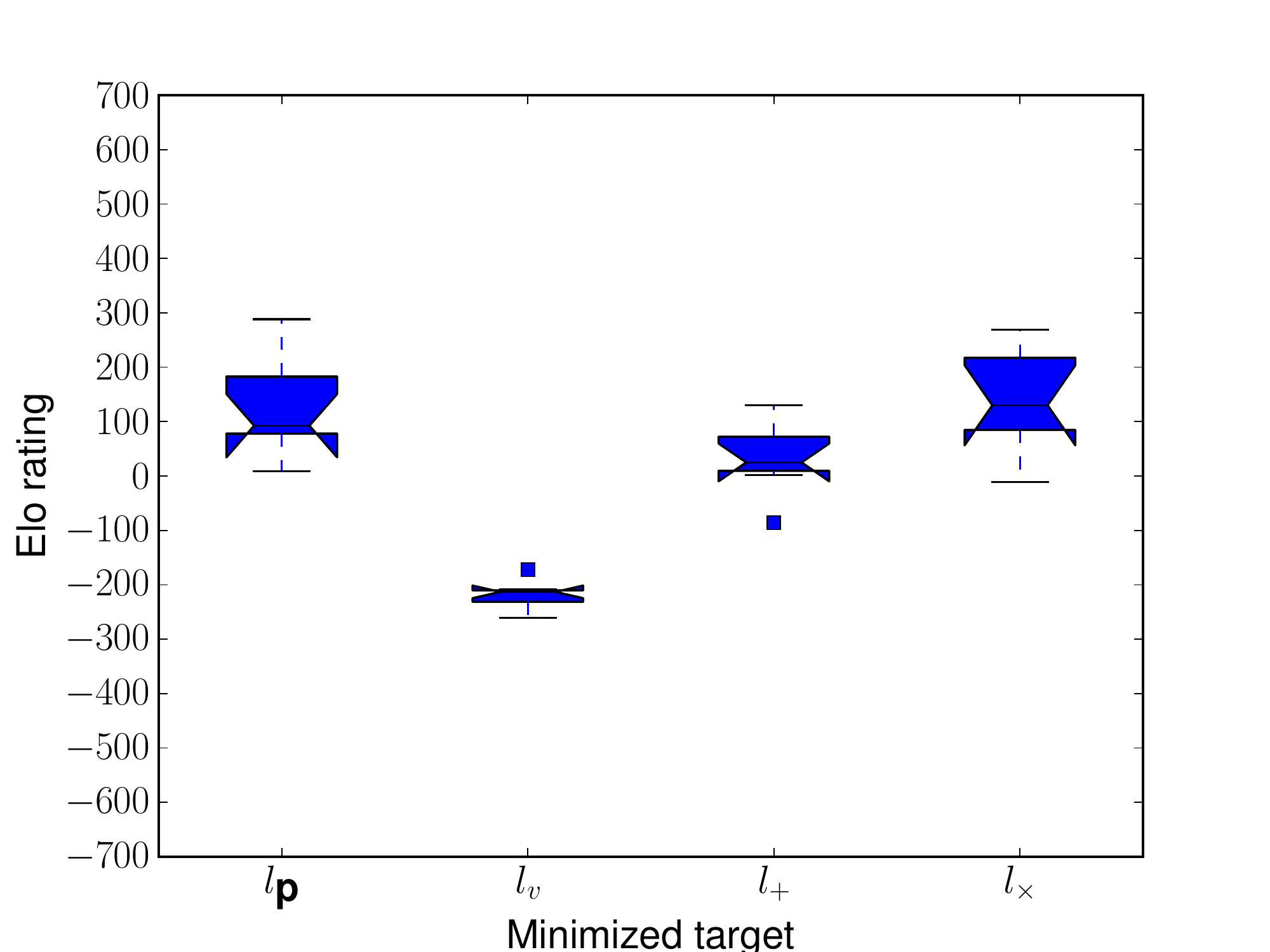}}
\hspace*{-1.8em}
\subfigure[6$\times$6 Gobang]{\label{fig:subfigEloallgamesplayers:f}
\includegraphics[width=0.45\textwidth]{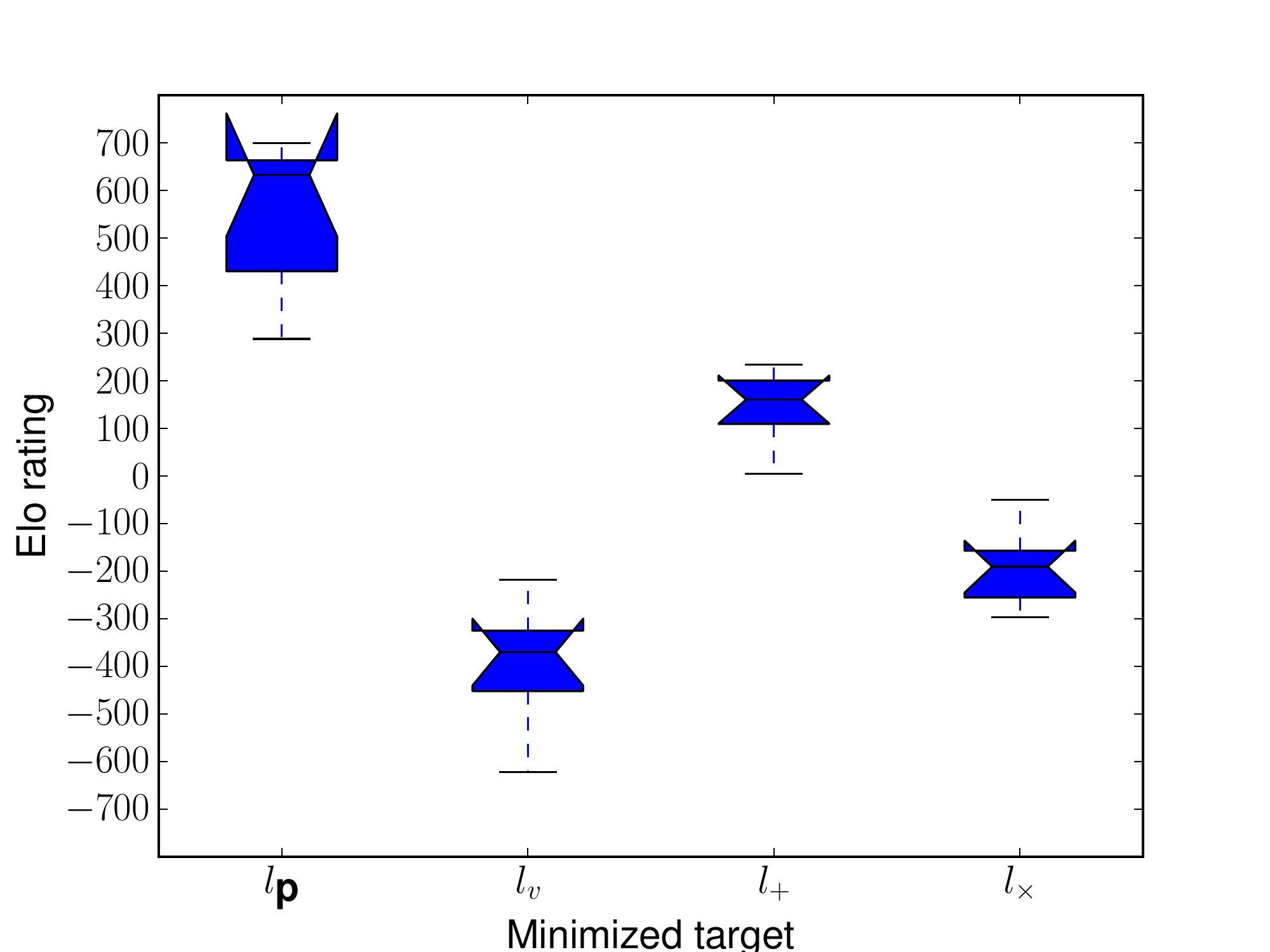}}
\caption{Round-robin tournament of all final models from minimizing different targets. For each game 8 final models from 4 different targets plus a random player~(i.e. 33 in total). In panel (a) the difference is small. In panel b, c, and d, the Elo rating of $l_v$ minimized players clearly dominates. However, in panel (f), the Elo rating of $l_\textbf{p}$ minimized players clearly achieve the best performance.}
\label{fig:subfigEloallgamesplayers}
\end{figure}
In order to measure which target can achieve better playing strength, we let all final models trained from 8 runs and 4 targets plus a random player pit against each other for 20 times in a full round robin tournament. This enables a direct comparison of the final outcomes of the different training processes with different targets. It is thus more informative than the training Elo due to the self-play bias, but provides no information during the self-play training process. In principle, it is possible to do this also during the training at certain iterations, but this is computationally very expensive.

The results are presented in Fig.~\ref{fig:subfigEloallgamesplayers}.
and show that minimizing $l_v$ achieves the highest Elo rating with small variance for 6$\times$6 Othello, 5$\times$5 Connect Four and 6$\times$6 Connect Four. For 5$\times$5 Othello, with 200 training iterations, the difference between the results is small. We therefore presume that minimizing $l_v$ is the best choice for the games we focus on. This is  surprising because we expected the $l_+$ to perform best as documented in the literature. However, this may apply to smaller games only, and 5$\times$5 Othello already seems to be a border case where overfitting levels out all differences.

In conclusion, we find that minimizing $l_v$ only is an alternative to the $l_+$ target for certain cases.
We also report exceptions, especially in relation to the Elo rating as calculated during training. The relation between Elo and loss during training is sometimes inconsistent~(5$\times$5 Connect Four training shows Elo decreasing while the losses are actually minimized) due to training bias. And for Gobang game, only minimizing $l_\textbf{p}$ is the best alternative. A combination achieves lowest loss, but $l_v$ achieves the highest training Elo. If we minimize product loss $l_\times$, this can result in higher Elo rating for certain games. More research into training bias is needed.

\section{Conclusion}\label{sec:conclusion}
% 1. alphago many hyperparameters
%    also loss functions
%    exciting results, holding promiss for future research. However, very computationally demanding. Therefore, insight needed in how to get good results quickly. 
% 2. first experiments to determine how important to find optimum of hyper parameters
%    through the use of small games we can vary
%    parameters to sweep through space
% 3. first contribution: whole history Elo, not running Elo 
% 4. second contribution: value loss function better
% 5. third contribution: determine the most important hyper parameters, show non-trivial interaction effects between epoch and simulations, and show that the outer loop iterations dominate the inner epoch, simulation, and episode parameters.
% This means that in practice, researchers can minimaze values for epoch, simulations and episodes, since the slef play training will compensate with being able to run more iteraitons, that will get the necessary simulaitons, network traning and replay buffer fulfilment anyway. In effect, our analuysis has shown that fine tuning is less important, since running many iterations compensates for too low inner training and search. It will be alrihgt, just let it run long enough.
AlphaGo has taken reinforcement learning by storm. The performance of the novel approach to self-play is stunning, yet the computational demands are  high, prohibiting the wider applicability of this method. Little is known about the impact of the values of the many hyper-parameters on the speed and quality of learning.  In this work, we  analyze  important hyper-parameters and combinations of loss-functions. We gain more insight and  find recommendations for faster and better self-play. We have used small games to allow us to perform a thorough  sweep using a large number of hyper-parameters, within a reasonable computational budget.  
We sweep 12 parameters in AlphaZeroGeneral~\cite{Surag2018} and analyse loss and time cost for 6$\times$6 Othello, 
and select the 4 most promising parameters for further optimization.

We more thoroughly evaluate the interaction between these 4 time-related hyper-parameters, and find that i) generally, higher values lead to higher playing strength; ii) within a limited budget, a higher number of the outer self-play iterations is more promising than higher numbers of the inner training epochs, search simulations, and game episodes. At first this is a surprising result, since conventional wisdom tells us that deep learning networks should be trained well, and MCTS needs many play-out simulations to find good training targets. 

In AlphaZero-like self-play, the outer-iterations subsume the inner training and search. Performing more outer iterations automatically implies that more inner training and search is performed. The training and search improvements carry over from one self-play iteration to the next, and long self-play sessions with many iterations can get by with surprisingly little inner training epochs and MCTS simulations. The sample-efficiency of self-play is higher than simple composition of the constituent elements would predict. Also, the implied high number of training epochs may cause overfitting, to be reduced by small values for epochs.

 Much research in self-play  uses the default loss function (sum of value and policy loss). More research is needed into the relative importance of value function and policy function. We evaluate 4 alternative loss functions for 3 games and 2 board sizes, and find that 
the best setting depends on the game and is usually not the sum of policy and value loss, but simply the value loss. However, the sum may be a good compromise.

In our experiments we also noted that care must be taken in computing Elo ratings. Computing Elo based on game-play results during training typically gives biased results that differ greatly from tournaments between multiple opponents. Final best models tournament Elo calculation should be used.

For future work, more insight into training bias is needed. Also, automatic optimization frameworks can be explored, such as~\cite{Birattari2002,Hutter2011}. Also, reproducibility studies should be performed to see how our results carry over to larger games, computational load permitting. %In addition, relationship among parameters may be much more complicate. This study introduces a further evaluation among $I$, $E$, $m$ and $ep$.  What's more, the paper shows that the choice of the optimal combined loss function can have a huge impact on Elo performance. Unfortunately, our computational resources did not allow us to test the approach on large board sizes, but the results should encourage research of loss functions and alternative Elo computation also for large scale games.

\section*{Acknowledgments.} Hui Wang acknowledges financial support from the China Scholarship Council (CSC), CSC No.201706990015.


\begin{thebibliography}{1}

\bibitem{Silver2016} Silver D, Huang A, Maddison C J, et al: Mastering the game of Go with deep neural networks and tree search. Nature \textbf{529}(7587), 484--489 (2016)

\bibitem{Silver2017a} Silver D, Schrittwieser J, Simonyan K, et al: Mastering the game of go without human knowledge. Nature \textbf{550}(7676), 354--359 (2017)

\bibitem{Silver2018} Silver D, Hubert T, Schrittwieser J, et al. A general reinforcement learning algorithm that masters chess, shogi, and Go through self-play. Science, 2018, 362(6419): 1140-1144.

%\bibitem{Granter2017} Granter S R, Beck A H, Papke Jr D J: AlphaGo, deep learning, and the future of the human microscopist. Archives of pathology $\&$ laboratory medicine \textbf{141}(5), 619--621 (2017)

%\bibitem{Wang2016} Wang F Y, Zhang J J, Zheng X, et al: Where does AlphaGo go: From church-turing thesis to AlphaGo thesis and beyon. IEEE/CAA Journal of Automatica Sinica \textbf{3}(2), 113--120 (2016)

%\bibitem{Fu2016} Fu M C: AlphaGo and Monte Carlo tree search: the simulation optimization perspective. Proceedings of the 2016 Winter Simulation Conference. IEEE Press pp. 659--670 (2016)

\bibitem{Tao2016} Tao J, Wu L, Hu X: Principle Analysis on AlphaGo and Perspective in Military Application of Artificial Intelligence. Journal of Command and Control \textbf{2}(2), 114--120 (2016)

\bibitem{Zhang2016} Zhang Z: When doctors meet with AlphaGo: potential application of machine learning to clinical medicine. Annals of translational medicine \textbf{4}(6), (2016)

\bibitem{Surag2018} N. Surag, https://github.com/suragnair/alpha-zero-general, 2018.

\bibitem{Wang2018} Wang H, Emmerich M, Plaat A. Monte Carlo Q-learning for General Game Playing. arXiv preprint arXiv:1802.05944 (2018)

\bibitem{Browne2012} Browne C B, Powley E, Whitehouse D, et al: A survey of monte carlo tree search methods. IEEE Transactions on Computational Intelligence and AI in games \textbf{4}(1), 1--43 (2012)

\bibitem{Ruijl2014} B Ruijl, J Vermaseren, A Plaat, J Herik: Combining Simulated Annealing and Monte Carlo Tree Search for Expression Simplification. In: B\'eatrice Duval, H. Jaap van den Herik, St\'ephane Loiseau, Joaquim Filipe. Proceedings of the 6th International Conference on Agents and Artificial Intelligence 2014, vol. 1, pp. 724--731. SciTePress, Set\'ubal, Portugal (2014)

\bibitem{Schmidhuber2015} Schmidhuber J: Deep learning in neural networks: An overview. Neural networks \textbf{61} 85--117 (2015)

\bibitem{Clark2015} Clark C, Storkey A. Training deep convolutional neural networks to play go. International Conference on Machine Learning. pp. 1766--1774 (2015)

%\bibitem{Mnih2015} Mnih V, Kavukcuoglu K, Silver D, et al: Human-level control through deep reinforcement learning. Nature \textbf{518}(7540), 529--533 (2015)

\bibitem{Heinz2000} Heinz E A: New self-play results in computer chess. International Conference on Computers and Games. Springer, Berlin, Heidelberg. pp. 262--276 (2000)

\bibitem{Wiering2010} Wiering M A: Self-Play and Using an Expert to Learn to Play Backgammon with Temporal Difference Learning. Journal of Intelligent Learning Systems and Applications \textbf{2}(2), 57--68 (2010)

\bibitem{Van2013} Van Der Ree M, Wiering M: Reinforcement learning in the game of Othello: Learning against a fixed opponent and learning from self-play. In Adaptive Dynamic Programming And Reinforcement Learning. pp. 108--115 (2013)

\bibitem{Plaat2020} Aske Plaat, Learning to Play: Reinforcement Learning and Games, Leiden, 2020, forthcoming.

\bibitem{mandai2018alternative} Mandai Y, Kaneko T. Alternative Multitask Training for Evaluation Functions in Game of Go. 2018 Conference on Technologies and Applications of Artificial Intelligence (TAAI). IEEE, 2018: 132-135.

\bibitem{caruana1997multitask} Caruana R. Multitask learning. Machine learning, 1997, 28(1): 41-75.

\bibitem{MatsuzakiK2018} Matsuzaki K, Kitamura N. Do evaluation functions really improve Monte-Carlo tree search?[J]. ICGA Journal, 2018 (Preprint): 1-11

\bibitem{Matsuzaki18} Matsuzaki K. Empirical Analysis of PUCT Algorithm with Evaluation Functions of Different Quality. 2018 Conference on Technologies and Applications of Artificial Intelligence (TAAI). IEEE, 2018: 142-147.

\bibitem{Iwata1994} Iwata S, Kasai T. The Othello game on an n$\times$n board is PSPACE-complete. Theoretical Computer Science. \textbf{123}(2), 329--340 (1994)

\bibitem{Allis1988} Allis V. A knowledge-based approach of Connect-Four-the game is solved: White wins. 1988.

\bibitem{Reisch1980} Reisch, S. Gobang ist PSPACE-vollst\"andig. Acta Informatica 13, 59¨C66 (1980). 

\bibitem{Wang2019} Wang H, Emmerich M, Preuss M and Plaat A. Alternative Loss Functions in AlphaZero-like Self-play. 2019 IEEE Symposium Series on Computational Intelligence (SSCI), Xiamen, China, 155--162 (2019).

\bibitem{Buro97} Buro M. The Othello match of the year: Takeshi Murakami vs. Logistello. ICGA Journal, 1997, 20(3): 189-193.

\bibitem{ChongTW05} Chong S Y, Tan M K, White J D. Observing the evolution of neural networks learning to play the game of Othello. IEEE Transactions on Evolutionary Computation, 2005, 9(3): 240-251.

\bibitem{ThillBKK14} Thill M, Bagheri S, Koch P, et al. Temporal difference learning with eligibility traces for the game connect four. 2014 IEEE Conference on Computational Intelligence and Games. IEEE, 2014: 1-8.

\bibitem{ZhangWu2012} Zhang M L, Wu J, Li F Z. Design of evaluation-function for computer Gobang game system [J][J]. Journal of Computer Applications, 2012, 7: 051.

\bibitem{BanerjeeS07} Banerjee B, Stone P. General Game Learning Using Knowledge Transfer. IJCAI. 2007: 672-677.

\bibitem{Wang18} Wang H., Emmerich M., Plaat A. (2019) Assessing the Potential of Classical Q-learning in General Game Playing. In: Atzmueller M., Duivesteijn W. (eds) Artificial Intelligence. BNAIC 2018. Communications in Computer and Information Science, vol 1021. Springer, Cham.

\bibitem{Ioffe2015} Ioffe S, Szegedy C: Batch normalization: accelerating deep network training by reducing internal covariate shift. Proceedings of the 32nd International Conference on International Conference on Machine Learning-Volume 37. pp. 448--456 (2015)

\bibitem{Kingma2014} Kingma D P, Ba J: Adam: A method for stochastic optimization. arXiv preprint arXiv:1412.6980 (2014)

\bibitem{Srivastava2014} Srivastava N, Hinton G, Krizhevsky A, et al: Dropout: a simple way to prevent neural networks from overfitting. The Journal of Machine Learning Research. \textbf{15}(1), 1929--1958 (2014)

\bibitem{Coulom08} Coulom R. Whole-history rating: A Bayesian rating system for players of time-varying strength. International Conference on Computers and Games. Springer, Berlin, Heidelberg, 113--124, 2008

\bibitem{MichaelA2006} Emmerich M T M, Deutz A H. A tutorial on multiobjective optimization: fundamentals and evolutionary methods. Natural computing, 2018, 17(3): 585-609.

%\bibitem{Schneider2002} Schneider M O, Rosa J L G: Neural connect 4-A connectionist approach to the game. Neural Networks SBRN 2002. Proceedings. VII Brazilian Symposium on. IEEE pp. 236--241 (2002)

\bibitem{Birattari2002} Birattari M, St\"utzle T, Paquete L, et al. A racing algorithm for configuring metaheuristics. Proceedings of the 4th Annual Conference on Genetic and Evolutionary Computation. Morgan Kaufmann Publishers Inc. 11-18 (2002)

\bibitem{Hutter2011} Hutter F, Hoos H H, Leyton-Brown K: Sequential model-based optimization for general algorithm configuration. International Conference on Learning and Intelligent Optimization. Springer, Berlin, Heidelberg, pp. 507--523 (2011)

\end{thebibliography}
\end{document}